\documentclass[sn-mathphys-num]{sn-jnl}

\usepackage{graphicx}%
\usepackage{multirow}%
\usepackage{amsmath,amssymb,amsfonts}%
\usepackage{amsthm}%
\usepackage{mathrsfs}%
\usepackage[title]{appendix}%
\usepackage{xcolor}%
\usepackage{textcomp}%
\usepackage{manyfoot}%
\usepackage{booktabs}%
\usepackage{algorithm}%
\usepackage{algorithmicx}%
\usepackage{algpseudocode}%
\usepackage{listings}%

\usepackage{anyfontsize} 
\usepackage{arydshln} 
\usepackage{subcaption}

\usepackage[markup=bfit, deletedmarkup=sout, authormarkup=brackets]{changes}

\definechangesauthor[name={Filipe}, color=brown]{FG}
\definechangesauthor[name={Matej_Hoffmann}, color=orange]{MH}
\definechangesauthor[name={Sergiu}, color=green]{SP}

\usepackage[firstpageonly=true]{draftwatermark}

\begin{document}

\SetWatermarkAngle{0}
\SetWatermarkColor{black}
\SetWatermarkLightness{0.5}
\SetWatermarkFontSize{13pt}
\SetWatermarkVerCenter{30pt}
\SetWatermarkText{\parbox{30cm}{%
\centering This is the final version of the manuscript that is published in \\
\centering Behavior Research Methods, 57, 280 (2025). \url{https://doi.org/10.3758/s13428-025-02816-x} \\
\centering (read-only open access link: \url{https://rdcu.be/eFwac})}}

\title[Automatic infant 2D pose estimation from videos: comparing seven deep neural network methods]{Automatic infant 2D pose estimation from videos: comparing seven deep neural network methods}

\author{\fnm{Filipe} \sur{Gama}}
\author{\fnm{Matěj} \sur{Mísař}}
\author{\fnm{Lukáš} \sur{Navara}}
\author{\fnm{Sergiu} \sur{T. Popescu}}
\author*{\fnm{Matej} \sur{Hoffmann}*}\email{matej.hoffmann@fel.cvut.cz}


\affil{Czech Technical University in Prague, Faculty of Electrical Engineering, Department of Cybernetics, Prague, Czech Republic}








\abstract{
Automatic markerless estimation of infant posture and motion from ordinary videos carries great potential for movement studies "in the wild", facilitating understanding of motor development and massively increasing the chances of early diagnosis of disorders. There is rapid development of human pose estimation methods in computer vision thanks to advances in deep learning and machine learning. However, these methods are trained on datasets that feature adults in different contexts. This work tests and compares seven popular methods (AlphaPose, DeepLabCut/DeeperCut, Detectron2, HRNet, MediaPipe/BlazePose, OpenPose, and ViTPose) on videos of infants in supine position and in more complex settings. Surprisingly, all methods except DeepLabCut and MediaPipe have competitive performance without additional finetuning, with ViTPose performing best. Next to standard performance metrics (average precision and recall), we introduce errors expressed in the neck-mid-hip (torso length) ratio and additionally study missing and redundant detections, and the reliability of the internal confidence ratings of the different methods, which are relevant for downstream tasks. Among the networks with competitive performance, only AlphaPose could run close to real time (27 fps) on our machine. We provide documented Docker containers or instructions for all the methods we used, our analysis scripts, and the processed data at \url{https://hub.docker.com/u/humanoidsctu} and \url{https://osf.io/x465b/}.
}




\maketitle

\section*{Introduction}
\label{sec:intro}

Movement analysis of infants is key to understanding their development. Insights from spontaneous infant behavior often draw on small datasets and simplified but laborious manual scoring of video recordings (e.g., \cite{DiMercurio2018}). Quantitative data such as kinematics from motion capture are an exception \cite{sloan2023meaning, kanazawa2023open}, and the length of the recordings is limited. Daily and longer spontaneous recordings of infants will be needed \cite{Adolph2008} to uncover developmental trajectories on multiple and nested time scales and to employ nonlinear or dynamical systems analysis tools \cite{sloan2023meaning}. The sharing of datasets, including raw video footage, is critical to making progress in psychological science \cite{adolph2017video, gilmore2017video}. Some efforts in this direction are under way, but manually scoring hours, days, and weeks of recordings becomes a bottleneck. Automatic and accurate extraction of motion data from this material is a critical prerequisite for progress. One example is the analysis of kinematics during reaching development. In order to obtain the relevant features (e.g., average speed, movement duration, maximum speed, jerk, path length, distance, straightness ratio) \cite{berthier2006development, konczak1997development} infants need to be tested in the laboratory, and motion capture markers or similar equipment is typically used, which limits the amount of data that can be collected and may even affect the infants' performance on the task. Another example are infants' self-directed spontaneous movements like self-touch. In \cite{stupperich2024quantity, DiMercurio2018, thomas2015independent}, they were scored manually, which is extremely laborious. Both manual and automatic scoring has been used in \cite{Khoury2022}. The key benefits of automatic pose estimation are thus the following: (i) recordings can be performed in the wild---at homes and without additional instrumentation; (ii) very long recordings can be automatically analyzed and reliability can be established by manually scoring only a subset of the data; (iii) complete 2D postures and movement trajectories are available. 

Another key area is clinical practice---identifying deviations from normal development.
Trained experts can identify the risks of developmental disorders, such as cerebral palsy, from spontaneous movements (using the General Movement Assessment (GMA) \cite{einspieler2005prechtl}) or through a neurological examination based on movement, posture, and reflexes (Hammersmith Infant Neurological Examination (HINE) \cite{romeo2016use}). However, expert evaluation constitutes a critical bottleneck in the early detection of developmental disorders, especially in less developed countries. 
Automated extraction and evaluation of movement patterns constitutes a key enabling technology to make screening available to a much larger population. 

In general, there are two classes of methods for collecting motion data \cite{chen2012sensor}: (i) \textit{direct sensing} where movements are captured using hardware attached to the bodies (e.g., inertial sensors and magnetic tracking systems) and (ii) \textit{indirect sensing} (e.g., 3D motion capture, RGB cameras, RGB-D cameras). From the first group (see \cite{chen2016review} for a review of wearable sensor systems to monitor body movements of neonates), the most popular are inertial sensors. These can be accelerometers \cite{heinze2010movement}, or, more frequently, inertial measurement units (IMUs -- 3-axis accelerometer, 3-axis gyroscope, 3 magnetometers) integrated into a wearable suit \cite{airaksinen2020automatic}. Wearable inertial sensors may be relatively inexpensive (compared to 3D motion capture), but the fact that they need to be physically attached to infants prevents their widespread use and may also affect spontaneous movement production. Moreover, inertial sensors do not provide absolute position information but only angular velocities and linear accelerations, which limits the analyses that can be run on the data (position information can be obtained from numerical integration but is prone to noise). Within indirect sensing, we want to draw a line between marker-based 3D motion capture systems and standard video cameras (RGB). On one end of the spectrum, 3D motion capture systems typically use infrared retroreflective markers placed on infant bodies and record their positions with multiple cameras. This can yield sub-millimeter accuracy but requires very expensive equipment, and the markers may affect spontaneous movement production. Furthermore, markers cannot be placed on the back of infants in supine position which prevents the use of standard skeleton solvers. Markers may also be used with standard video cameras \cite{kanemaru2013specific}. Multi-camera setups (e.g., 9 high-frequency cameras with studio lighting in \cite{needham2021accuracy}) provide an alternative to marker-based motion capture systems. On the other end of the spectrum are standard video cameras \cite{Shin2022}, such as those in consumer cell phones, which constitute the only device that is truly accessible to almost everyone, without any barriers. Recordings from two cameras or RGB-D cameras containing depth information (e.g., Microsoft Kinect, Intel Realsense) could still be relatively easily applied ``in the wild''. 

The next step after motion data acquisition is their analysis. One of the key applications is automated clinical movement assessment. The methods can be classified based on the type of input data (e.g., accurate absolute 3D positions of body parts from a motion capture on one end of the spectrum or a single video stream from an ordinary camera on the other end) and on the assessment being performed---see \cite{marcroft2015movement} for a review. Most often, the desired output is automated GMA classification (see \cite{Silva2021, irshad2020ai} for reviews), with estimated risks of cerebral palsy prediction as the typical clinical outcome. Direct sensing methods (e.g., accelerometers in \cite{heinze2010movement}) or 3D data from marker-based motion capture have been used \cite{meinecke2006movement} but our focus is on indirect sensing and video-based approaches in particular (see Silva et al.\cite{Silva2021} for a review). The images of infants can be used to directly extract features that are used for subsequent analyzes and assessment (e.g., \cite{Tsuji2020} for GMA or \cite{Kinoshita2020} to analyze U-shaped developmental changes). With the rapid advent of pose estimation methods in computer vision and machine learning, methods that use 2D keypoint extraction in infant recordings are becoming increasingly popular. McCay et al.~\cite{McCay2020} extracted the positions of infant body keypoints using OpenPose~\cite{Cao2019_Openpose}, computed histograms of joint orientation and displacement in 2D and then used deep learning to classify abnormal movements (GMA labels). Chambers et al.~\cite{Chambers2020} used the keypoints obtained from OpenPose as input to extract movement features and then applied a Na{\"\i}ve Bayesian Surprise Metric to predict neuromotor development risk. Reich et al.~\cite{reich2021} used the positions of the 25 keypoints obtained from OpenPose directly for classification and evaluated the agreement with GMA. Shin et al.~\cite{Shin2022} obtained the positions of 2D keypoints using AlphaPose \cite{Fang2022_Alphapose}, estimated joint angle values from the 2D skeletons, analyzed complexity and then applied deep learning for classification (reporting correlations with HINE). Similarly, the risk of autism spectrum disorder (ASD) can be predicted from infant video recordings using input images directly \cite{Doi2022} or using 2D pose estimation (OpenPose) first \cite{Kojovic2021}.

\begin{figure}[!htb]
    \centering \includegraphics[scale=1.0, width=1\linewidth]{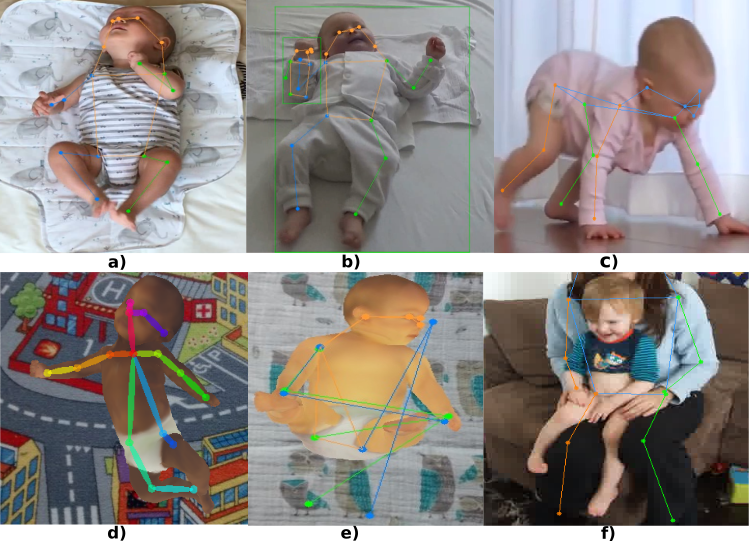}
    \caption{Examples of successful and erroneous keypoint estimates obtained from the different human pose estimation methods: a) HRnet Bottom-Up (``Supine'' dataset, infant AA, 17 weeks) successful; b) HRnet Top-Down (``Supine'' dataset, TH, 10 weeks) with a successful estimate but a second hallucinated person; c) ViTPose (SyRIP image 293) successful; d) OpenPose missing the left ankle keypoint (synthetic infant 1); e) HRNet Bottom-Up failing to estimate a complex leg position (synthetic infant 10); f) ViTPose detecting the parent but not the infant (``Lap'' dataset).}
    \label{fig:pem_failings}
\end{figure}

Methods for automatic human keypoint extraction (for example, eyes, wrists, hips, feet, etc.), also called human pose estimation, from images and videos are rapidly evolving, with performances increasing every year (see \cite{elshami2024comparative, zheng2023deep, munea2020progress} for recent surveys). These provide key enabling technology to make automatic motion extraction and analysis from ``in the wild'' recordings possible. 
These methods were developed primarily to estimate posture and movement of adult bodies (see \cite{koul2025accurately} for a recent comparison of OpenPose and marker-based motion capture). So far, these methods have been applied to infant videos ``as is''. However, the morphology of the infant body is different from the body proportions of the adult, especially in early infancy \cite{Boniol2008, Huelke1998}. Additionally, small infants are typically in supine position (on their back) and move differently than standing adults, which constitutes a dataset which is out of the distribution the adult pose estimation models were trained on.
Although automated pose estimation seems to be generally good (see Fig.~\ref{fig:pem_failings} a-c), in a previous study \cite{Khoury2022}, some limitations of current pose estimation methods in infants were highlighted. For example, OpenPose \cite{Cao2019_Openpose} was found to struggle with some camera angles, body postures, and keypoints. In particular, complex leg positions, such as when legs are crossed in supine position, are not often estimated with high accuracy. This can be seen in Fig.~\ref{fig:pem_failings} (d): OpenPose misses the left ankle keypoint and places the left knee keypoint on the thigh, very close to the left hip. In the current study, comparing different pose estimation methods, we found similar examples. Fig.~\ref{fig:pem_failings} (e) shows that HRNet Bottom-Up fails on a complex leg position and fails to place the left shoulder and elbow keypoints, placing them on the right; while Fig.~\ref{fig:pem_failings} (b) shows that HRNet Top-Down hallucinates a second person in the image (see the additional bounding box at the infant's right forearm), and Fig.~\ref{fig:pem_failings} (f) shows ViTPose not detecting the infant while sitting on the lap of a parent (the image was cropped to not show the parent's face in the figure). 

Thus, our first goal and contribution is to assess the performance of 2D human pose estimation methods when applied specifically to infants in supine position. In order to provide a more general outlook, we additionally present results on images of infants in a wider variety of postures, including standing or turning away from the camera, as well as with infants on the lap, or sitting or standing in front, of their parent.
Only few studies have partially addressed this problem. However, if present, ``in the wild'' recordings did not constitute the majority of the datasets. Needham et al.~\cite{needham2021accuracy} used a multi-camera setup in the lab and compared OpenPose, AlphaPose and DeepLabCut.
Groos et al.~\cite{Groos2022} created a large proprietary dataset of human-annotated infant images from both controlled and standardized environments and ``in the wild'' recordings, and used them to retrain several models, including OpenPose. Their best model achieves on average 81.11\% of its estimates being located within an error of 10\% of the length of the infant's face with an average error at 6.8\% of the face length, while the ``vanilla'' OpenPose had 49.66\% of its estimates within this margin of error with an average error of 14.3\% of the face length. Sermpon et al.~\cite{Sermpon2024} have found weak correlations between pose estimation features derived from MediaPipe's estimates and manually-annotated movements towards the midline. Jahn et al.~\cite{Jahn2025} compared OpenPose, MediaPipe, HRNet, AggPose and VitPose, as well as fine-tuned versions of ViTPose and HRNet (AGMA-HRNet48) on infant recordings in a laboratory context made to specifically target GMA settings (infants in supine position, seen from above and at a diagonal angle). Their results show that ViTPose-huge has the second best results (higher than AGMA-HRNet48, the HRNet model fine-tuned with infant images from a similar GMA context), locating 84.6\% of its estimates within 2.5cm of the ground truth (estimated as 10\% of the torso length), while their re-trained ViTPose locates 93.89\% of estimates within this distance but seems to overfit and to have lower performance than the non-trained ViTPose on other datasets according to their supplementary materials. 

In this article, we compare the performance of seven methods: Alphapose \cite{Fang2022_Alphapose}, DeepLabCut / DeeperCut (v2.2.1.1) \cite{Insafutdino2016_DeeperCut}, Detectron2 (v0.6) \cite{Wu2019_detectron2}, MediaPipe / BlazePose (v0.10.14) \cite{Lugaresi2019_mediapipe, bazarevsky2020_blazepose}, HRNet (\textit{MMPose} v0.28.0) \cite{xiao2018_hrnet, sun2019deep_hrnet} \textit{Bottom-Up} (HRnet BU) and \textit{Top-Down} (HRnet TD), OpenPose \cite{Cao2019_Openpose}, and ViTPose (\textit{MMPose} v1.1.0) \cite{xu2022_vitpose}.

As a second contribution of this work, we put together and evaluated a number of measures to assess the performance of the pose estimation methods. Besides the ones typically used by the machine learning community (Average Precision and Recall), we consider alternative measures that express the errors scaled to the infant's body dimensions (Neck-MidHip error), and we additionally analyze the percentage of missing and redundant detections and the reliability of the confidence estimates that the pose estimation methods output. We also report the processing time of the methods. All of these together provide a more complete picture and allow the user to make an optimal choice of the method. 

The third contribution of this work constitutes in making all the pose estimation methods and tools for the results analysis available to the community. Although the pose estimation algorithms themselves are publicly accessible in their corresponding repositories, we make available the versions we used including the complete environment in public Docker containers (at \url{https://hub.docker.com/u/humanoidsctu}) with instructions and additional abstractions to simplify their use. In addition, we also share the evaluation scripts and the detailed results (at \url{https://osf.io/x465b/}). 

\section*{Methods}

\subsection*{2D Pose estimation methods}

To ensure that comparisons are as fair as possible, we used versions of the pose estimation methods that were trained with the same dataset: COCO \cite{Lin_2014_COCO}, although some methods only have available weights that used other datasets in their training, instead of or in addition to COCO (see each method's respective subsection below or the summary Tab.~\ref{tab:summary_methods}). The methods provide 17 estimated keypoints, with the exception of DeepLabCut, MediaPipe and OpenPose that provide 14, 33 and 18 keypoints respectively. We will call a single set of estimated keypoints a \textit{detection}. The details of the parameters with which the methods were run are provided in the Supplementary Materials.
All methods except MediaPipe provide a form of confidence values as their own internal evaluation of the estimated result, with higher values indicating higher confidence in the quality of the estimates. This value can be provided for the whole set of keypoints or for individual keypoints (or both). Following the trend in the documentation and methods' outputs, we will use \textit{score} when referring to the whole detection value and \textit{confidence} when referring to the individual keypoints values.
The methods follow either one of the two approaches: \textit{Top-Down}, which first uses a detector to delimit areas of interest in the image that encompass a person to detect and then estimates a set of keypoints within this space; and \textit{Bottom-Up}, which finds keypoints or sets of keypoints separately on the whole image before joining them into full keypoint sets and separating them into several individuals if necessary.

\textbf{AlphaPose} \cite{Fang2022_Alphapose} is based on a Residual Convolutional Neural Network (R-CNN) architecture. It is a Top-Down approach. It provides both scores and confidences.

\textbf{DeepLabCut/DeeperCut} We use the DeepLabCut environment \cite{Mathis2018_DeepLabCut}, with its human pose estimation \textit{DeeperCut} \cite{Insafutdino2016_DeeperCut} based on an R-CNN architecture. It follows a Bottom-Up approach. It has been trained on the MPII dataset \cite{andriluka2014_mpii} and provides 14 keypoints. It provides confidences, but no scores.

\textbf{Detectron2} \cite{Wu2019_detectron2} is based on an R-CNN architecture and follows a Top-Down approach. It provides both scores and confidences.

\textbf{MediaPipe/BlazePose} While commonly known as MediaPipe \cite{Lugaresi2019_mediapipe}, the human pose estimation of this environment uses BlazePose \cite{bazarevsky2020_blazepose} and is based on a CNN architecture, following a Top-Down approach. It has been trained on a custom dataset with additional face and hand annotations on top of COCO's regular 17 keypoints. It provides neither scores nor confidences.

\textbf{HRnet \textit{Bottom-Up} and \textit{Top-Down}} \cite{xiao2018_hrnet, sun2019deep_hrnet} are named after the name of the neural architecture used, a High Resolution Network, itself based on CNNs. It has been trained using both COCO and MPII datasets. We use its implementation through the \textit{MMPose} environment \cite{mmpose2020}, which proposes bottom-up (BU) and top-down (TD) versions of HRNet. HRNet BU provides scores, while HRNet TD provides confidences.

\textbf{OpenPose} \cite{Cao2019_Openpose} is based on a CNN architecture. It is a Bottom-Up approach. It has been trained using both COCO and MPII datasets and provides confidences.

\textbf{ViTPose} \cite{xu2022_vitpose} uses a non-hierarchical, no-CNN backbone, vision Transformer Network. It is a Top-Down approach. It has been trained on multiple datasets, including COCO, AI Challenger, and MPII, in the version we use: VitPose-H (huge). It provides confidences. We use its implementation in the \textit{MMPose} environment \cite{mmpose2020}.

\begin{table}[!ht]%
    \addtolength{\tabcolsep}{-0.6em}
    \centering
    \footnotesize
    \begin{tabular}{l|c|c|c|c|c|c|c|c|}
    Properties & AlphaPose & DeepLabCut & Detectron 2 & MediaPipe & HRNet BU & HRNet TD & OpenPose & ViTpose \\
    \hline
    TD/BU  & TD & BU & TD & TD & BU & TD & BU & TD \\
    Architecture & R-CNN & R-CNN & R-CNN & CNN & HRNet & HRNet & CNN & Transformer \\
    Scores (Y/N) & Y & N & Y & N & Y & N & N & N \\
    Confidences (Y/N) & Y & Y & Y & N & N & Y & Y & Y \\
    Training dataset & COCO & MPII & COCO & COCO & COCO+MPII & COCO+MPII & COCO+MPII & COCO+MPII+AIC \\
    \end{tabular}
    \caption{Summary comparison overview of the methods used. TD: Top-Down, BU: Bottom-Up.}
    \label{tab:summary_methods}
\end{table}

\subsection*{Datasets}
Unlike for adult pose estimation, datasets of infant recordings with annotated keypoints are scarce and usually not publicly shared. For this study, we used the only two available public datasets that we know of, as well as two datasets from our own video recordings, which we manually annotated (occluded keypoints were not annotated).

Our main comparison datasets concern infants in supine position, with a single infant in every image.

\begin{itemize}
 \item \textit{Real infants ``\textbf{Supine}'' dataset}. 720 annotated images (90 per video) from 8 videos (59200 images total) of two infants, followed longitudinally between 2 and 6 months of age. The annotations were made using the DeepLabCut labeling tool. 480 images (60 per video) were chosen by taking the first ten images every 100 images, from 0 to 9, then 100 to 109, until 509. The remaining 240 images (30 per video) were selected through DeepLabCut's integrated selection tool using k-means clustering to pick the least similar images in each video. Fourteen keypoints were annotated, corresponding to the COCO keypoints minus the ears and the nose. 
 Our dataset is entirely comprised of infants recorded at home by their parents, each video being different from the others in terms of camera angles, lighting conditions, as well as background and clothes colors and type.
 
 \item \textit{Synthetic infants -- \textbf{MINI-RGBD}}. Based on recordings of real infants, Hesse and colleagues have trained a model capturing infant shape and posture distributions which they subsequently used to produce and render synthetic sequences of artificial infants under 7 months in supine position---the MINI-RGBD dataset \cite{Hesse2018_minirgbd}. There are 12 \textit{synthetic infants} with 1000 images each. Ground truth for 25 keypoints is available. Of these 25 keypoints, 13 are in common with COCO's 17 keypoints and were used to evaluate the methods. Each of the twelve synthetic sequences is unique, including backgrounds with large variations between sequences (see Fig.~\ref{fig:pem_failings} (d, e) for two examples, or see the original paper \cite{Hesse2018_minirgbd}).
\end{itemize}

In addition, we used two datasets of annotated images of infants in different and more complex contexts:
\begin{itemize}
 \item \textit{Real infants on the ``\textbf{Lap} (sitting)'' dataset}. 359 annotated images from 6 videos of 6 different infants aged from 17 weeks to 18 months. Two of these videos (infants) were recorded at home by the parents, with the infant sitting on the lap of a parent, themselves sitting on a couch. The remaining four videos (infants) were recorded in a laboratory context, with the camera placed in front of the participants, who were sitting (or standing) on a mat placed on the ground. The parent, and in particular their face, was not always fully visible in the images. The number of 359 images was reached after filtering down the images that were originally annotated but where the infant was missing (moved out of the field of the camera), or because an experimenter was in front of the camera and occluded most of the infant.
 \item \textit{Real infants -- \textbf{SyRIP}}. Huang and colleagues created a hybrid dataset consisting of 1700 infant images: 1000 synthetic and 700 real infants, which they manually annotated and made public: the SyRIP dataset \cite{Huang2021_syrip}. The subset dataset dedicated to benchmarking performance rather than training, which is what we used in our comparisons, contains 500 images of 50 infants up to 1 year old, in a large variety of contexts (at home, outside, in a laboratory), postures (supine, prone, standing, facing towards or away from the camera) and camera angles (see the original paper, which shows numerous examples). The images were taken from YouTube videos and Google Images results. 
\end{itemize}

For real infants, it is difficult to estimate the difficulty of each sequence or image, though it is clear that our ``Lap'' and the SyRIP datasets would be more difficult to estimate than our ``Supine'' dataset. The SyRIP dataset contains a wider variety of infant postures and camera angles, including many that partially hide the infants (e.g., from the side), making the estimation more difficult. As for the ``Lap'' dataset, since the methods were all originally trained on datasets of adult images, they are biased toward detecting adults, and the presence of the adult overlapping right behind the infant is a strong source of disturbance. For synthetic infants, the MINI-RGBD dataset has a definition of the sequence difficulty: \textit{ easy} (IDs 1 to 4), \textit{medium} (5 to 9) and \textit{difficult} (10 to 12), with this difficulty judged primarily with regard to the posture of the infant. Although the main manuscript focuses on the average results, it might be interesting to look at the results at the level of individual videos or their estimated difficulty. Such results are available in the Supplementary Materials.

A second coder annotated 20\% of our manually-annotated image dataset. Intraclass Correlation Coefficients (ICC) were calculated for each keypoint and for each axis independently to measure the reliability of the human annotations using the Python module \textit{pingouin}. The lowest ICC was 0.91 for the y-axis of the Left Hip keypoint, with the lower bound of its 95\% Confidence Interval at 0.89. The mean ICC was $0.97 \pm 0.03$, with all p-values $\ll 0.001$.

Our supplementary materials include the results for an extended version of our ``Supine'' dataset: 720 supplementary images from 8 additional videos; as well as a continuous sequence of 900 images (30 seconds) of a single infant at 17 weeks old. A summary table of each dataset's characteristics is available in Sup. Tab.~1.

\subsection*{Comparison metrics}
\label{subsec:comp_metrics}

The metrics used for comparison are described below.
Because we observed differences in the results when the methods were given input in the form of videos or images, we present the results for both input types independently, except for DeepLabCut and ViTPose (within the \textit{MMPose} environment) that only accepted video input.

The image-reference error that we compute as a building block for some of the metrics is the Euclidean distance between an estimated keypoint and its ground truth. Note that this ignores any gaps in the data when a method is not able to provide a keypoint's estimated coordinates (see the \textit{Missing Data} metric), or when ground truth is missing because the body part is occluded, as no distance can be computed in such cases.

Out of all these metrics, the ones that can be used even without ground truth, and for which we have the results on the full videos on our ``Supine'' dataset are: percentages of missing data and redundant detections, and the processing speed.

DeepLabCut's errors were evaluated on the 12 out of 14 keypoints that were in common with the ground truths.

\subsubsection*{Object Keypoint Similarity (OKS)}
OKS \cite{Ronchi2017} is a standard metric computing a single value per detected object, encompassing all its keypoints. OKS values lie between 0 and 1, higher values indicating a higher \textit{similarity}.

Given $K$ keypoints that are both annotated for ground truth and detected by the pose estimation method, with $k \in [1, K]$: $d_k$ is the Euclidean distance between an estimated keypoint $p_k = {(x_{kp}, y_{kp})}$ and its ground truth $gt_k = {(x_{gt}, y_{gt})}$; $s$ is the area of the \textit{bounding-box} around the target: a rectangle that encompasses all visible parts of the target within the image. As we do not have manually annotated bounding-boxes, we computed this area by taking the minimum and maximum X and Y values from all ground-truth keypoint coordinates in the image as an approximation. The coefficient $c_k$ is specific to each keypoint location and gives more weight to keypoints of which the ground-truth position varies less between human coders. We used the values recommended by the COCO challenge \cite{Ronchi2017}, available on its website, which are based on the annotations of the COCO dataset and give the most weight to the eyes and nose, and the least weight to the hips. From these, \textit{keypoint similarity} $ks$ is computed for each estimated keypoint. OKS is the average $ks$ for a detection.

\begin{equation}
    d_k = \sqrt{(x_{gt} - x_{kp})^2 + (y_{gt} - y_{kp})^2}
    \qquad
    ks(k) = e^{-\frac{d_k^2}{2s^2c_k^2}}
    \qquad
    OKS = \frac{\sum_{k=1}^K ks(k)}{K}
    \label{eq:oks}
\end{equation}

\subsubsection*{Average Precision (AP) and Average Recall (AR)}
To evaluate the performance of pose estimation, the gold standard metrics in the literature are Average Precision and Average Recall \cite{Ronchi2017}, ranging between 0 and 100, higher values indicating better performance. We follow the benchmark evaluation set by the COCO challenge to calculate AP and AR, averaging the values computed over ten thresholds from 0.5 to 0.95, with steps of 0.05. Roughly, if the OKS for a detection is above the threshold and there is any ground truth that has not been matched with a detection yet on a given image, then the detection counts as True Positive. Other cases count as False Positives. If there is no detection but there is a ground truth, then it counts as a False Negative. Precision and Recall are computed for each threshold, then averaged.
Note that despite the name \textit{Average Precision}, it is actually computed as the area under the Precision-Recall Curve~\cite{Everingham_2012_pascal-voc, Ronchi2017}.
In our case, the computation is simplified, as all our annotated images contain one single infant (one ground truth). For our ``Lap'' dataset, the ground truth annotations of the parents were ignored, so that each method was only evaluated against the infant annotations.

\subsubsection*{Neck-MidHip error}
\label{subsubsec:meth_nmhe}
As an alternative to OKS, we additionally use a reference distance, the \textit{Neck-MidHip} length, corresponding roughly to the length of the infants' torso. With this reference distance, we normalize the errors between recordings to compensate for different camera settings and infants of different ages and with different body sizes. Unlike OKS, such errors are easier to grasp as they directly relate to a measurable part of the body of the infant. Lower is better, and the minimum is 0.
Others and ourselves have previously used a similar strategy in a similar context \cite{Jahn2025, Khoury2022, Chambers2020, McCay2020}.

The head length used in \cite{Groos2022} can be an alternative. However, the torso is a more stable body part that does not rotate as much as the head, especially for infants in supine position, making it a better choice as a normalization reference. Additionally, neither the ground truth following the COCO format \cite{Lin_2014_COCO} nor the MINI-RGBD ground truth \cite{Hesse2018_minirgbd} contain keypoints that can be used to compute the length of the head without larger approximations than for the torso.

The equations are shown in Eq.~\ref{eq:nmhe}. Given $j$ a recording with $I_j$ annotated images for which Neck and MidHip keypoints ground truth are available or can be approximated. Given the images from the recording $i \in [0, I_j]$, then the Neck-MidHip length for this image $nmh\_len_{ji}$ is defined in pixels as the Euclidean distance between the Neck and the MidHip ground-truth coordinates in this image.
For our ``Supine'' and the MINI-RGBD dataset, we take the median of all $nmh\_len_{ji}$ as the NeckMidHip length to use to normalize the recording: $nmh\_len_j$.
Given a keypoint $k$ that is visible in the image and annotated, and detected by the pose estimation method, then its Neck-MidHip error $nmh\_error_{kij}$ is its error (Euclidean distance) relative to its ground truth, $d_{kij}$, divided by the median Neck-MidHip length for the recording, $nmh\_len_j$.
For our ``Lap'' and the SyRIP dataset, which contain images from different sources mixed together, we normalize using each images' neck-MidHip distance $nmh\_len_{ji}$ instead of a median from a whole recording. This has the downside that only images where this length can be estimated are used to calculate the error.
Then, we obtain per-keypoint-location mean Neck-MidHip errors, and an overall mean full-body error, from all images and recordings.

\begin{equation}
    \begin{split}
        & nmh\_len_{ji} = \sqrt{(x_{neck} - x_{midhip})^2 + (y_{neck} - y_{midhip})^2}
        \qquad
        nmh\_len_j = med(nmh_{ji}) \\ 
        & nmh\_error_{kij} = \frac{d_{kij}}{nmh\_len_j}
    \label{eq:nmhe}
    \end{split}
\end{equation}

For the synthetic infants in the MINI-RGBD dataset, the Neck and MidHip positions are provided directly in the ground truth. In the real infants datasets, the ground truths do not provide Neck or MidHip keypoints. Instead, the centre of the right and left Shoulder and Hip keypoints are used, turning into a ``MidShoulder-MidHip'' length. We will simply write Neck-MidHip for simplification in the rest of the paper.
On our ``Supine'' dataset and on the MINI-RGBD dataset, the Neck-MidHip length is computed using the median of the Neck-MidHip lengths for a whole video. This enables the use of all annotated images and keypoints even if one of the keypoints needed to compute the length is missing from the ground truth in a particular image (occlusions). It is also less subject to noise in the annotations, with the drawback that a single length is used for a whole video and does not fit exactly to each individual image. Using the median can only be done when there are no changes in camera settings, in particular zoom or camera position, which can happen, for example, when recordings are performed at home by a parent holding a smartphone with their hand; otherwise, it requires to separate the recording into subsets according to the different settings, or use per-image normalization instead, which is what we opted to do for our ``Lap'' and the SyRIP datasets.

\subsubsection*{Percentage of missing data}
\label{subsubsec:meth_mdat}
Missing data is especially undesirable for motion analysis, which requires accurate and consistent sampling over time. Missing data can happen in two ways.
First, \textit{missing detections}, when a method is not able to detect an infant in the image and does not provide any keypoint at all.
Second, \textit{missing keypoints}, when a method provides at least one detection, but fails to provide coordinates for one or more individual keypoints. An example is given in Fig.~\ref{fig:pem_failings} (d), where the left foot keypoint is missing.
This can happen either because of an erroneous estimation from the method or when body parts are occluded.
Average Recall takes into account missing detections, but no other metric takes into account missing keypoints.

The total percentage of missing data computation is described in Eq.~\ref{eq:mdet}. We define the maximum amount of data $maxdata_m$ that can be estimated by a method $m$ as the number of images to estimate $N$ times the number of keypoints provided by the method $m_{kp}$. Missing detections $mdet$ are converted to this unit by multiplying with $m_{kp}$ as well, since any missing detection is equivalent to missing all keypoints in this one image. This number is added to the count of individual missing keypoints $mkp$, and divided by $maxdata_m$ to obtain the total percentage of missing data for this method: $mdata_m$.
\begin{equation}
    maxdata_m = N \times m_{kp}
    \qquad
    mdata_m = ((mdet_m \times m_{kp}) + mkp) / maxdata_m
    \label{eq:mdet}
\end{equation}

\subsubsection*{Percentage of redundant detections}
\label{subsec:meth_rdet}
Some methods can erroneously detect more people than there are in the image, as shown in Fig.~\ref{fig:pem_failings} (b), where an additional "person" is detected. These redundant detections might indicate that the method \textit{hallucinates} other people on the image where there are none, or provides several estimates for the same person with different positions. Such behavior is undesirable, in particular for applications with more complex environments that, for example, involve interaction with an experimenter or the parents, such as in our ``Lap'' dataset, as it complexifies tracking each separate individual when ground truth is not available.

The percentage of redundant detections is computed as the ratio between i) the difference in the number of detections between the expected and provided detection numbers (number of estimated detections minus the number of expected detections) divided by ii) the number of images the method provided detections for (total number of images minus the number of missing detections). For example, if there are 150 detections for 100 input images with 100 expected detections (1 infant per image), then 50 detections would be considered redundant, resulting in a percentage of 50\% of redundant detections, translating that, on average, the method provides one additional detection every two images where it detected something. A percentage of 100\% would mean that, on average, the method always provides two detections for each image, when one is expected.
DeepLabCut, MediaPipe, and OpenPose did not have redundant detections (MediaPipe and OpenPose, have a built-in parameter to limit the detections to 1 person per image).

In the case of the ``Lap'' dataset, where an adult is always present behind the infant, there are also cases where only the adult and not the infant is detected (Fig.~\ref{fig:pem_failings} (f)). We empirically observed that some pose estimation methods have a tendency to often detect only the adult and not the infant in such context. This would show as a negative ratio of redundant detections. Taking the example from above, if a method had provided 150 detections for 100 images where 200 detections were expected due to the presence of both an infant and an adult, then we would get -50\% ``redundant'' detections: on average, that method provided the two expected detections for half of the images, and only provided one detection for the other half. For this particular dataset, this is not a foolproof measure, as it is possible that a method detected the adult and then provided a second, redundant, detection of the adult or hallucinated another person somewhere else in the image instead of detecting the infant.

Percentages of redundant detections closer to 0 are preferred.
Average Precision partially accounts for redundant detections: redundant detections with high scores modify AP values, but redundant detections with low scores do not.

\subsubsection*{Correlations between scores and OKS}
In a real application scenario, without available ground truth, confidence or score values can be useful to provide cues about detections or keypoints that should not be trusted. For example, confidences are used as coefficients in the optimization process of \textit{smplify-x} \cite{Pavlakos2019_SMPL-X} when optimizing the parameters of the 3D full-body pose estimation from the 2D pose estimation.
These values should be positively correlated with the OKS (the higher the OKS, the higher the score). When available, we use the score provided by the methods for each image. When not available, we used the median of the confidences as an alternative score for the whole detection. MediaPipe could not be evaluated as it does not provide scores or confidences.
Using its implementation within the Python package \textit{scipy.stats} \cite{Virtanen2020_Scipy}, we measured correlations with Spearman Rank Order Coefficients.

\subsubsection*{Processing speed}
Measured in frames per second, it represents how fast the methods can process the data. This can make the difference for real-time applications.

\subsubsection*{Combined Performance Evaluation (CPE)}
It is desirable to have a metric that combines the metrics above. However, several metrics (OKS, AP and AR, Neck-MidHip error) are concerned with the accuracy of the keypoint detections in the image and hence are not independent. Additional metrics we consider are supplementary and their relevance depends on the context (redundant detections, confidence-OKS correlations, processing speed). It would still be useful to combine one of the accuracy metrics with the the percentage of missing data.

AP and AR could be combined into a single metric like the F1 score, but cannot be easily combined with the percentage of missing data as AR already considers missing detections. OKS uses weighted coefficients for each keypoint type and has a non-linear relationship with the absolute errors. As such, we preferred to opt for the metric that had the least overlap with others and was the most neutral: Neck-MidHip errors.

Therefore, we propose the Combined Performance Evaluation (CPE) as a metric combining the accuracy of detections based on the Neck-MidHip errors with the percentage of missing data through Equation~\ref{eq:single_unified}.

\begin{equation}
    f\colon x \rightarrow (1 - MIN(1, (x/(c \times 100)))) \in [0, 1]
    \qquad
    \text{CPE}_m = (f(nmh\_error_m) + f(mdata_m)) / 2
    \label{eq:single_unified}
\end{equation}
where $f(x)$ is a function that turns the positive percentage-based values of the Neck-MidHip errors and missing data into a normalized scale between 0 and 1. Here, 1 corresponds to perfect estimations with no missing data and no estimation error. The scaling coefficient $c$ determines the limit (threshold) value that allows non-zero output for any $x$ value below the threshold. Here, we set $c = 0.5$ so that the threshold (divisor) is $50$ by considering that estimation errors above 50\% of the torso length and missing data on more than 50\% of a recording are not acceptable for the use of pose estimation methods for further analysis. This coefficient penalizes larger errors while providing finer granularity of the final values to separate methods that have similar performance. The final score is computed as the mean between the scaled and normalized values obtained for the Neck-MidHip errors and the missing data.

\subsection*{Hardware and methods' code modifications}
\label{subsec:hardw}
The methods were run on a computer with the following specs: CPU: Intel(R) Xeon(R) W-2295 (18C / 36T, 3.0 / 4.8GHz, 24.75MB), GPU: Two NVIDIA TU104GL Quadro RTX 5000, RAM: 251 GB DDR4, OS: Ubuntu 20.04.5 LTS

All methods were run while turning off the display of visualizations on the screen and without saving the visualizations to the disk. To do so, some of the code of Detectron2 and the \textit{MMPose} environment was modified, as no parameter was provided to turn them off. Further modifications were made to the Detectron2 code to output the keypoint coordinates as a file, as it did not provide a way to get such output by default.

The ``Lap'' and SyRIP datasets contain non-sequential images extracted from different videos of different infants at different ages. The video input option was thus not appropriate.
On ViTPose, which in the version of \textit{MMPose} we used could only process videos, we modified the code of \textit{MMPose}'s ViTPose to accept images as inputs, without modifying any of the underlying processing done by ViTPose. While that enabled us to present ViTPose results on these datasets, an important comparison due to ViTPose being the contender state-of-the-art method in adult pose estimation, ViTPose might internally only expect videos and have processing steps that enhance estimations assuming a time coherence between two processed frames. As there was no parameter available to indicate to ViTPose a lack of sequential coherence, we avoided using the image processing steps for the other two datasets that had videos that could be processed (``Supine'', and MINI-RGBD). The performance of ViTPose on the ``Lap'' and SyRIP dataset could be lower than in a real application scenario with full video recordings to process.
All modifications are effective in the Dockers that we share (at \url{https://hub.docker.com/u/humanoidsctu}) and are indicated in the documentation.

\section*{Results}

The results for both real and synthetic infants are shown for each metric described above. In each subsection, we provide the results for the manually-annotated dataset of real infants first: in order, our ``Supine'' and ``Lap'' datasets, then the SyRIP dataset, and then the results for the MINI-RGBD dataset of synthetic infants are presented last.
Due to the varying complexity of infant postures across the datasets, we provide detailed results for individual sequences in the Supplementary Materials when relevant.

\subsection*{Object Keypoint Similarity}
\label{subsec:oks}

The average OKS values for each dataset (average across individual infant images) are shown in Tab.~\ref{tab:oks} (see Supplementary Tables ST. 6 and ST. 7 to see the full details per video on the ``Supine'' and MINI-RGBD datasets).

For real infant images in supine position, ViTPose and HRNet TD have the highest OKS, followed by HRNet BU. The results between frame-by-frame and video inputs are almost identical, except for OpenPose, for which image input leads to better results. On the ``Lap'' and SyRIP datasets, we observe a significant drop of performance for all methods, with particularly high variance, and wider gaps of performance between the methods. HRNet BU seems to perform the best on the ``Lap'' dataset, while ViTPose leads on the SyRIP dataset. 
For synthetic infants, ViTPose and HRNet TD also show the best results. The differences between image and video inputs are narrow but consistent between the different methods, with a slight edge for image inputs. Synthetic infants in supine position are more difficult to process for all methods, in particular for HRNet BU, compared to the real infants in supine position, with the exceptions being DeepLabCut and MediaPipe, which perform better on synthetic infants, but the performance is overall higher than for the ``Lap'' or the SyRIP datasets.

\begin{table}[!ht]%
    \addtolength{\tabcolsep}{-0.4em}
    \centering
    \begin{tabular}{c|c|cccccccc|}
    & Input & AlphaPose & DeepLabCut & Detectron 2 & MediaPipe & HRnet BU & HRnet TD & OpenPose & ViTPose\\
    \hline
    \textbf{Real} & Images & 0.87$\pm$0.07 & N/A & 0.87$\pm$0.10 & 0.39$\pm$0.19 & 0.90$\pm$0.07 & \textbf{0.92$\pm$0.05} & 0.87$\pm$0.11 & N/A \\
    ours -- Sup. & Videos & 0.87$\pm$0.08 & 0.12$\pm$0.12 & 0.86$\pm$0.11 & 0.40$\pm$0.18 & 0.90$\pm$0.07 & 0.92$\pm$0.05 & 0.79$\pm$0.19 & \textbf{0.92$\pm$0.04} \\
    \hdashline
    ours -- Lap & Images & 0.58$\pm$0.33 & N/A & 0.54$\pm$0.28 & 0.35$\pm$0.28 & \textbf{0.68$\pm$0.26} & 0.59$\pm$0.37 & 0.63$\pm$0.30 & 0.62$\pm$0.39 \\
    \hdashline
    SyRIP & Images & 0.72$\pm$0.17 & N/A & 0.75$\pm$0.19 & 0.01$\pm$0.02 & 0.74$\pm$0.16 & 0.78$\pm$0.16 & 0.74$\pm$0.16 & \textbf{0.79$\pm$0.16} \\
    \hline
    \hline
    \textbf{Synth.} & Images & 0.84$\pm$0.11 & N/A & 0.84$\pm$0.10 & 0.50$\pm$0.21 & 0.83$\pm$0.16 & \textbf{0.88$\pm$0.7} & 0.83$\pm$0.12 & N/A\\
    MINI-RGBD & Videos & 0.81$\pm$0.15 & 0.43$\pm$0.20 & 0.81$\pm$0.12 & 0.47$\pm$0.22 & 0.81$\pm$0.17 & 0.86$\pm$0.10 & 0.81$\pm$0.12 & \textbf{0.87$\pm$0.07}\\
    \hline
    \end{tabular}
    \caption{Average OKS values across individual images for each method, dataset and input type. N/A values correspond to methods not evaluated on image inputs as they only accept video input (see Sec. ``Comparison metrics'' first paragraph and Sec. ``Hardware and methods' code modifications'' for more details)}.
    \label{tab:oks}
\end{table}

\subsection*{Average Precision and Average Recall}
\label{subsec:apar}

The details of the AP and AR values are shown in Tab.~\ref{tab:arap}.

For real infants, ViTPose has the highest AP and AR. Considering only AP, ViTPose is followed by HRNet BU and then by OpenPose. Lower AP values for HRNet TD and Detectron2 in our ``Supine'' dataset are influenced by a large number of redundant high-confidence detections (see Tab.~\ref{tab:totalrdet}). Considering only AR, ViTPose is followed by HRNet TD and HRNet BU. Whether the inputs are images (frames) or videos, the results are roughly similar except for OpenPose, though it seems that, generally, image inputs lead to better results (except for HRNet TD, for which video input seems preferred).
There is a large drop of AP and AR in the ``Lap'' and SyRIP datasets compared to the supine infants.

For synthetic infants, both the AP and AR values are lower than those in the ``Supine'' dataset of real infants; it seems that the methods have more difficulty processing synthetic infants, except for DeepLabCut and MediaPipe. Considering only AP, ViTPose is the best method, followed by OpenPose. Considering only AR, HRNet TD is the best method, followed by ViTPose. We observe again that despite the lower scores, the synthetic infants of the MINI-RGBD datasets, which are also in supine position and seen from above, are better estimated by all methods than the real infants in the ``Lap'' and in the SyRIP datasets.

In conclusion, ViTPose displays the best overall results regardless of the benchmarking dataset or the input method. Depending on whether AP or AR needs to be prioritized and on the specific dataset used for benchmarking, the second place is shared by HRNet BU, HRNet TD or OpenPose.

\begin{table}[!ht]%
    \addtolength{\tabcolsep}{-0.4em}
    \centering
    \begin{tabular}{l|cccccccc|}
    & AlphaPose & DeepLabCut & Detectron 2 & MediaPipe & HRnet BU & HRnet TD & OpenPose & ViTPose \\
    \hline
    \textbf{ours--Supine images} & & & & & & &\\
    AP & 67.7 & N/A & 52.6 & 1.8 & \textbf{75.3} & 59.7 & 74.4 & N/A \\
    AR & 74.7 & N/A & 79.4 & 6.1 & 85.1 & \textbf{89.3} & 79.0 & N/A \\
    
    \textbf{ours--Supine videos} & & & & & & &\\
    AP & 67.5 & 0.0 & 24.1 & 1.8 & 74.6 & 60.5 & 55.2 & \textbf{88.5} \\
    AR & 74.3 & 0.0 & 78.7 & 5.9 & 84.8 & 89.2 & 62.4 & \textbf{90.9} \\
    \hdashline
    \textbf{ours--Lap images} & & & & & & &\\
    AP & 25.0 & N/A & 11.0 & 7.8 & 10.9 & 20.1 & 27.0 & \textbf{30.4} \\
    AR & 57.6 & N/A & 49.4 & 20.3 & 60.1 & \textbf{69.5} & 67.4 & 60.7 \\
    \hdashline
    \textbf{SyRIP images} & & & & & & &\\
    AP & 28.0 & N/A & 25.6 & 0.0 & 26.2 & 30.9 & 38.4 & \textbf{41.9} \\
    AR & 50.3 & N/A & 57.1 & 0.0 & 53.8 & 63.8 & 54.2 & \textbf{65.4} \\
    \hline
    \hline
    \textbf{MINI-RGBD images} & & & & & & &\\
    AP & 60.5 & N/A & 48.1 & 6.1 & 62.2 & 59.9 & \textbf{66.5} & N/A\\
    AR & 69.7 & N/A & 73.9 & 17.8 & 76.7 & \textbf{81.7} & 72.5 & N/A\\
    \hline
    \textbf{MINI-RGBD videos} & & & & & & &\\
    AP & 52.1 & 3.2 & 29.5 & 5.6 & 53.8 & 56.8 & 61.2 & \textbf{73.7}\\
    AR & 62.7 & 12.3 & 68.0 & 17.4 & 71.6 & 77.9 & 67.5 & \textbf{79.1}\\
    \hline
    \end{tabular}
    \caption{Average Precision (AP) and Average Recall (AR) for each method, dataset and input type.}
    \label{tab:arap}
\end{table}

\subsection*{Neck-MidHip error}
\label{subsec:n-mhip}

\begin{figure}[!ht]
    \centering
    \includegraphics[scale=1, width=1\linewidth]{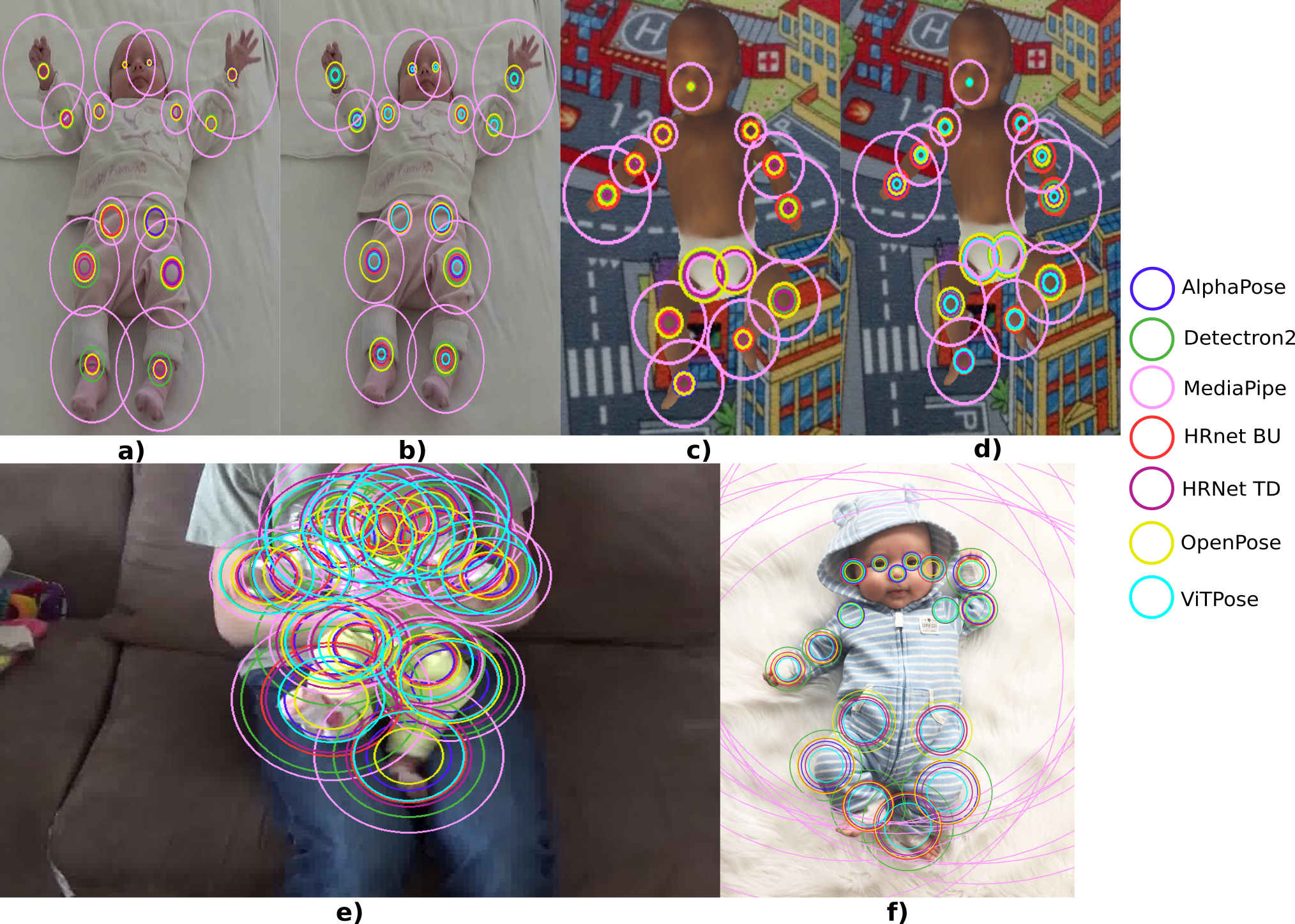}

    \caption{Neck-midhip errors for each keypoint with available ground truth. The centre of the circles is the ground-truth position for that keypoint. The radius of each circle shows the average error as a proportion of the Neck-MidHip length (see Methods). The colors represent separately each pose estimation method. a) Real Infants ``Supine'', image input; b) Real infants ``Supine'', video input; c) MINI-RGBD synthetic infants, image input; d) MINI-RGBD synthetic infants, video input; e) Real infants ``Lap''; and f) Real infants, SyRIP. DeepLabCut estimates leading to much higher errors not shown (see Supplementary Fig. SF. 2).}
    \label{fig:n-mh_circles}
\end{figure}

Neck-MidHip errors are shown in Fig.~\ref{fig:n-mh_circles}. DeepLabCut errors are not shown, as they create large radius circles that impede readability of the figures. The version of the figures including DeepLabCut is available in the Supplementary Materials Figure SF. 2.
Across all methods, the estimates of the eyes and nose are the most accurate, leading to the smallest errors, followed by shoulders and wrists. The positions of the hips and knees are detected with the highest errors. We can observe the increase in errors between the real infants in the ``Supine'' dataset, the MINI-RGBD dataset, the SyRIP dataset, and finally the ``Lap'' dataset where the errors make the image difficult to read.
The details of the overall average Neck-MidHip errors across all keypoints with their standard deviations for each method are shown in Tab. ~\ref{tab:n-mh_median}. The variability of the errors is high, often above one third, and sometimes even half, of the average error. This might be explained by the large variability between each keypoint type as can be seen on Fig.~\ref{fig:n-mh_circles}, or Supplementary Materials Tables ST 8-11.

Overall, for real infants, the best method is ViTPose, followed by HRNet TD. However, in the ``Lap'' dataset, while the performance of all the methods drops significantly, ViTPose and the other top-down methods seem to struggle particularly compared to their performance on the other datasets. HRNet BU has the lowest errors among all methods, though the errors are high in absolute terms: an average error of 24\% of the Neck-MidHip length. A less dramatic but general drop of performance is observed in the SyRIP dataset as well, where the best performing method is ViTPose with relatively high average errors at 12\% of the Neck-MidHip length.
For synthetic infants, the best method is HRnet TD, followed by ViTPose, although HRNet TD has slightly higher standard deviations.

As an additional comparison, we tested to what extent a ``mixture of experts'' approach---combining several of the estimation methods---could yield better performance. We tested  averaging the estimates from the best-performing methods: ViTPose, HRNet TD and HRNet BU. The details can be found in the Supplementary Materials, sections ``Mixture of experts estimates'' and ``Results summary for the `mixture of experts' estimates''. Overall, the averaged estimates are worse than the best individual method. The mixture could make the estimates more robust in face of missing data though.

With the final goal of completely automating the pose estimation and replacing human annotation, we were interested in how the pose estimation methods directly compare against human coders (note that ground truth for the datasets is obtained by humans clicking on individual keypoints in every image). 20\% of images from our ``Supine'' dataset was annotated by a second coder and a good inter-rater reliability was achieved (0.97 ICC on average). To compare the accuracy of the human annotators with the performance metrics used here, we computed the Neck-MidHip errors for the second human annotator with respect to the first annotator (ground truth). The average Neck-MidHip error is 7.6\%$\pm$4.2, which is on average higher than for the best performing automated methods (ViTPose scored 6.0\%$\pm$2.7), demonstrating the potential of automated pose estimation. The details of errors per keypoint is available in the Supplementary Materials Tab. 8.

\begin{table}[!ht]%
    \addtolength{\tabcolsep}{-0.4em}
    \centering
    \begin{tabular}{c|c|cccccccc|}
    Dataset & Input & AlphaPose & DeepLabCut & Detectron 2 & MediaPipe & HRNet BU & HRNet TD & OpenPose & ViTPose \\
    \hline
    \textbf{Real} 
    & Images& $8.3\pm3.1$ & N/A & $10.2\pm5.2$ & 38.0$\pm$13.8 & $7.8\pm3.8$ & \textbf{6.7}$\pm$\textbf{3.6} & $8.8\pm4.1$ & N/A \\
    Supine & Videos& $8.3\pm2.9$ & $99.6\pm25.9$ & $10.5\pm5.4$ & 35.1$\pm$13.7 & $7.9\pm3.9$ & $6.7\pm3.6$ & $12.3\pm4.3$ & \textbf{6.0}$\pm$\textbf{2.7} \\ 
    \hdashline
    Lap & Images & 38.0$\pm$6.1 & N/A & 41.6$\pm$10.4 & 55.2$\pm$12.2 & \textbf{24.3$\pm$9.0} & 42.9$\pm$9.3 & 25.5$\pm$6.0 & 40.0$\pm$8.9 \\
    \hdashline
    SyRIP & Images & 15.7$\pm$6.3 & N/A & 19.3$\pm$11.0 &  243.5$\pm$22.4 & 15.8$\pm$7.2 & 13.5$\pm$5.8 & 14.9$\pm$8.6 & \textbf{12.0$\pm$4.3} \\
    \hline
    \hline
    \textbf{Synth.}
    & Images & 9.8$\pm$4.6 & N/A & 9.6$\pm$5.1 & 28.2$\pm$10.9 & 11.7$\pm$3.8 & \textbf{7.6}$\pm$\textbf{5.0} & 10.7$\pm$5.4 & N/A \\
    MINI-RGBD & Videos & 11.6$\pm$4.3 & 43.5$\pm$13.5 & 11.3$\pm$5.0 & 31.2$\pm$12.1 & 13.5$\pm$4.5 & 9.1$\pm$4.7 & 11.4$\pm$5.4 & \textbf{8.2}$\pm$\textbf{4.7} \\
    \hline
    \end{tabular}
    \caption{Average errors across all keypoints as a percentage of the Neck-MidHip segment, with standard deviation, for each method.}
    \label{tab:n-mh_median}
\end{table}
\normalsize

\subsection*{Missing Data}
\label{subsec:mdata}

The percentages of missing data are shown in Tab.~\ref{tab:total_mdata}.

For real infants, the methods that never miss detections or keypoints are DeepLabCut and Detectron2, while HRNet BU is very close behind, with only a few misses that are lost when rounded at less than 2 decimal points. ViTPose and HRNet TD miss an amount of data that could be considered acceptable, below 0.5\%. On the other hand, OpenPose, MediaPipe, and AlphaPose display high amounts of missing data. We observe a tendency to miss smaller amounts of data with video input compared to image input. OpenPose reaches more than 50\%, the highest number of missing data overall, on the ``Lap'' dataset.

For synthetic infants, the methods that produce missing data are OpenPose, MediaPipe, and AlphaPose, though to a lesser degree than for real infants. Contrary to the real infants, we observe a tendency to miss smaller amounts of data with image input compared to video input.

\begin{table}[!ht]%
    \addtolength{\tabcolsep}{-0.4em}
    \centering
    \begin{tabular}{c|c|c|c|c|c|c|c|c|c|}
    & Input & AlphaPose & DeepLabCut & Detectron 2 & MediaPipe & HRnet BU & HRnet TD & OpenPose & ViTPose\\
    \hline
    \textbf{Real} & Images & 7.4 & N/A & \textbf{0} & 9.3 & \textbf{0} & 0.1 & 10.1 & N/A \\
    Supine & Videos & 6.9 & \textbf{0} & \textbf{0} & 9.5 & \textbf{0} & 2.9 & 8.3 & 0.3 \\
    \hdashline
    Lap & Images & 0.8 & N/A & \textbf{0} & 6.4 & \textbf{0} & \textbf{0} & 54.3 & \textbf{0} \\
    \hdashline
    SyRIP & Images & 2.8 & N/A & \textbf{0} & 20.4 & 0.4 & 0.2 & 13.8 & \textbf{0} \\
    \hline
    \hline
    \textbf{Synth.} & Images & 5.7 & N/A & \textbf{0} & 8.5 & \textbf{0} & \textbf{0} & 3.6 & N/A \\
     MINI-RGBD & Videos & 9.9 & \textbf{0} & \textbf{0} & 8.5 & \textbf{0} & \textbf{0} & 3.7 & \textbf{0} \\
     \hline
    \end{tabular}
    \caption{Percentage of missing data for each method, dataset and input type.}
    \label{tab:total_mdata}
\end{table}

\subsection*{Redundant detections}
\label{subsec:rdet}

The percentages of redundant detections for each method are shown in Tab.~\ref{tab:totalrdet}. The details for each individual video can be found in the Supplementary Tables ST. 12 and 13.

DeepLabCut, MediaPipe and OpenPose cannot have redundant detections by design, though for the case of the ``Lap'' dataset, they could have minus detections.
For the other methods, on the ``Supine'' real infants dataset, the methods with the least amount of redundant data are AlphaPose and HRNet BU. Except for HRNet TD, we observe a tendency, especially for Detectron2, to produce more redundant detections with video input.

On the ``Lap'' dataset, HRNet BU, Detectron2, and HRNet TD display higher rates of redundant detections, while on the contrary, MediaPipe, ViTPose, and AlphaPose seem to not always detect both individuals and only detect one of them (often the adult) for high percentages of the images. OpenPose is the method that overall has the least difference in detections compared to the expected amount. 
On SyRIP, all methods except MediaPipe and OpenPose display high amounts of redundant detections.

For synthetic infants, AlphaPose has the least amount of redundant detections, outside of the methods that could not have any.

Overall, Detectron2 and HRNet TD overall provide disproportionate amounts of redundant detections, especially on real infants, though HRNet BU has the most redundant detections in the ``Lap'' context, where, on the contrary, MediaPipe struggles the most at providing at least 2 detections (one of each individual on the image).

In the ``Supine'' and MINI-RGBD datasets, it is possible to look at the details per video (see Supplementary Materials Tables ST~11 and 12.) In doing so, we see that the distribution of redundant detections is not uniform. Detectron2, HRNet TD and ViTPose show higher redundant detections for synthetic infants that are considered more difficult in the MINI-RGBD dataset, while HRNet BU shows higher redundant detections for the easier synthetic infants.

\begin{table}[!ht]%
    \addtolength{\tabcolsep}{-0.4em}
    \centering
    \begin{tabular}{c|c|c|c|c|c|c|c|c|c|}
    & Input & AlphaPose & DeepLabCut & Detectron 2 & MediaPipe & HRnet BU & HRnet TD & OpenPose & ViTPose\\
    \hline
    \textbf{Real} & Images & 2.5 & N/A & 32.4 & \textbf{0} & 6.2 & 51.4 & \textbf{0} & N/A \\
    Supine & Videos & 2.7 & \textbf{0} & 277.1 & \textbf{0} & 6.7 & 48.8 & \textbf{0} & 23.1 \\
    \hdashline
    Lap & Images & -34.0 & N/A & 105.8 & -84.2 & 220.3 & 78.0 & \textbf{-5.0} & -45.1 \\
    \hdashline
    SyRIP & Images & 23.5 & N/A & 53.9 & \textbf{0} & 41.2 & 58.3 & \textbf{0} & 32.0 \\
    \hline
    \hline
    \textbf{Synth.} & Images & \textbf{0} & N/A & 34.4 & \textbf{0} & 8.3 & 26.6 & \textbf{0} & N/A \\
     MINI-RGBD & Videos & 0.1 & \textbf{0} & 92.5 & \textbf{0} & 11.8 & 23.8 & \textbf{0} & 14.7 \\
     \hline
    \end{tabular}
    \caption{Percentage of redundant detections for each method, dataset and input type.}
    \label{tab:totalrdet}
\end{table}

\subsection*{Correlations between scores and OKS}
\label{subsec:corr}
The different methods provide internal estimates of the quality of their detections in each frame (score for the whole keypoint set; confidence for each keypoint). This can be useful for downstream tasks---detections with a low score can be ignored, for example. We investigated the correlation between scores (method estimates) and actual accuracy (OKS values). Table ~\ref{tab:corr} reports the Spearman Rank Correlation Coefficients.

For real infants, in the ``Supine'' context, the method with the highest Correlation Coefficients is HRNet BU for both input types. With image input, it is closely followed by OpenPose. The correlations are in the correct direction for all methods.
It is interesting to note that on the ``Lap'' dataset, some methods maintain correlation scores in line with the other datasets, while others have lower correlations (AlphaPose, OpenPose, and ViTPose). In contrast, on the SyRIP dataset, most methods display higher correlations, especially AlphaPose. Still, HRNet BU is the method with the highest correlations in all the real infants datasets.

For synthetic infants, the methods with the highest Correlation Coefficients is OpenPose for image inputs. For video inputs, HRNet BU has the highest correlations, closely followed by OpenPose.

Overall, the correlations between the scores and the real accuracy are moderate, as the highest reach values of 0.68, while most of the correlations are low, between 0.2 and 0.5, and should therefore be used with caution. 

For completeness, the scatterplots of the score and OKS values for each method on each dataset are in the Supplementary Materials, Figs. SF.~3-6.

\begin{table}[!ht]%
    \addtolength{\tabcolsep}{-0.4em}
    \centering
    \begin{tabular}{c|c|c|c|c|c|c|c|c|c|}
    & Input & AlphaPose & DeepLabCut & Detectron 2 & MediaPipe & HRnet BU & HRnet TD & OpenPose & ViTPose\\
    \hline
    \textbf{Real} & Images & 0.28 & N/A & 0.19 & N/A & \textbf{0.68} & 0.47 & 0.63 & N/A \\
    Supine & Videos & 0.31 & 0.01 ($p=0.79$) & 0.19 & N/A & \textbf{0.67} & 0.49 & 0.28 & 0.52 \\
    \hdashline
    Lap & Images & 0.09 ($p=0.08$) & N/A & 0.54 ($p=0.31$) & N/A & \textbf{0.70} & 0.45 & 0.41 & 0.22  \\
    \hdashline
    SyRIP & Images & 0.58 & N/A & 0.42 & N/A & \textbf{0.80} & 0.67 & 0.57 & 0.56 \\
    \hline
    \hline
    \textbf{Synth.} & Images & 0.41 & N/A & 0.51 & N/A & 0.29 & 0.43 & \textbf{0.58} & N/A \\
     MINI-RGBD & Videos & 0.39 & 0.46 & 0.53 & N/A & \textbf{0.66} & 0.51 & 0.61 & 0.48 \\
     \hline
    \end{tabular}
    \caption{Spearman correlations between individual frames OKS values and score values for each method, dataset and input type. All p-values $\protect<$ 0.005 except for DeepLabCut on real infant videos with video input.}
    \label{tab:corr}
\end{table} 

\subsection*{Processing speed}
\label{subsec:fps_mem}

Table~\ref{tab:totalruntime} shows the processing speeds for each method in the synthetic infant dataset.
DeepLabCut and AlphaPose are the fastest methods by a large margin, running at close to 30 fps, while MediaPipe and OpenPose are close to running at 15 fps.
\begin{table}[!ht]%
    \centering
    \begin{tabular}{c|c|c|c|c|c|c|c|c|}
    \textbf{Synth.} & AlphaPose & DeepLabCut & Detectron 2 & MediaPipe & HRnet BU & HRnet TD & OpenPose & ViTPose\\
    \hline
    Images & \textbf{27.5} & N/A & 7.1 & 14.1 & 2.4 & 6.5 & 14.0 & N/A \\
    Videos & 27.0 & \textbf{28.8} & 5.9 & 15.4 & 2.3 & 7.1 & 13.0 & 4.8 \\
    \hline
    \end{tabular}
    \caption{Processing speed (fps), for each method and input type across the synthetic infants dataset.}
    \label{tab:totalruntime}
\end{table}

\subsection*{Combined Performance Evaluation (CPE)}
\label{subsec:uscore}
The CPE values, combining Neck-MidHip errors (error-based metric) and the percentage of missing data, are shown in Tab.~\ref{tab:total_uscore}.

On real infants, ViTPose is the best performing method overall due to its lower estimation errors and a low amount of missing data, except on the ``Lap'' dataset where it has higher errors and HRNet BU leads with the highest CPE value. The differences appear for other methods that are closer on their estimation errors but vary more largely on their amounts of missing data. The performance gap between the methods shows more clearly when looking at the more challenging SyRIP and ``Lap" datasets, with the extreme example of OpenPose that had the second lowest Neck-MidHip errors on the ``Lap'' dataset, right after HRNet BU, but ends up with the lowest CPE value among all methods on this dataset, as it misses 54\% of the data, while HRNet BU does not miss any data and has the highest CPE value.

For synthetic infants, the CPE values are overall very close to the real infants in supine position, with ViTPose and HRNet TD coming out on top.

\begin{table}[!ht]%
    \addtolength{\tabcolsep}{-0.4em}
    \centering
    \begin{tabular}{c|c|c|c|c|c|c|c|c|c|}
    & Input & AlphaPose & DeepLabCut & Detectron 2 & MediaPipe & HRnet BU & HRnet TD & OpenPose & ViTPose\\
    \hline
    \textbf{Real} & Images & 0.84 & N/A & 0.90 & 0.53 & 0.92 &\textbf{0.93} & 0.81 & N/A \\
    Supine & Videos & 0.85 & 0.50 & 0.90 & 0.55 & 0.92 & 0.90 & 0.79 & \textbf{0.94} \\
    \hdashline
    Lap & Images & 0.61 & N/A & 0.58 & 0.44 & \textbf{0.76} & 0.57 & 0.25 & 0.6 \\
    \hdashline
    SyRIP & Images & 0.82 & N/A & 0.81 & 0.30 & 0.84 & 0.86 & 0.71 & \textbf{0.88} \\
    \hline
    \hline
    \textbf{Synth.} & Images & 0.85 & N/A & 0.90 & 0.63 & 0.88 & \textbf{0.92} & 0.86 & N/A \\
     MINI-RGBD & Videos & 0.79 & 0.57 & 0.89 & 0.60 & 0.87 & 0.91 & 0.85 & \textbf{0.92} \\
     \hline
    \end{tabular}
    \caption{Combined Performance Evaluation values based on Neck-MidHip errors and Missing Data percentages for each method, dataset and input type.}
    \label{tab:total_uscore}
\end{table}

\section*{Conclusion}
\label{sec:conclusion}
With extremely rapid progress in human pose estimation methods from images and videos, this technology lends itself to deployment in infant motion analysis, providing a tool that can be truly applied ``in the wild'' with inputs, say, from a cell phone camera and without additional requirements or restrictions. If the pose estimation accuracy is satisfactory, this will have far-reaching implications for our understanding of normal motor development as well as the early diagnosis of infant developmental disorders. 

This article provided an empirical comparison of seven state-of-the-art human pose estimation methods---trained on images of adults during various activities, typically in upright positions---on several datasets of infant images, including infants under 7 months of age in supine position, infants between 17 weeks to 18 months sitting and in close proximity with a parent, and infants below 1 year old in a variety of contexts and postures. Our results are summarized below.

First, we conclude that state-of-the-art human pose estimation methods work well to estimate infant poses even without additional training or fine-tuning for infants in supine position. An overview of the estimation accuracy for individual keypoints on the body is provided in Fig.~\ref{fig:n-mh_circles}, and in Supplementary Materials Tables ST~6-9. Overall accuracy evaluations are available in Tables~\ref{tab:oks} and~\ref{tab:arap} using standard performance metrics (OKS, AP/AR) and in Tab.~\ref{tab:n-mh_median} using an infant-size centred metric. ViTPose has the best accuracy \cite{xu2022_vitpose}, followed by HRNet \cite{xiao2018_hrnet, sun2019deep_hrnet} (top-down variant). The other networks tested (HRNet bottom-up, AlphaPose~\cite{Fang2022_Alphapose}, Detectron2~\cite{Wu2019_detectron2}, and OpenPose~\cite{Cao2019_Openpose}) have higher errors. The DeepLabCut environment \cite{Mathis2018_DeepLabCut}, with human pose estimation using \textit{DeeperCut}~\cite{Insafutdino2016_DeeperCut}, as well as MediaPipe \cite{Lugaresi2019_mediapipe} with BlazePose \cite{bazarevsky2020_blazepose} do not provide competitive results at all. It is coherent that ViTPose, being based on one of the newest state-of-the-art neural network architecture, Transformers, which introduces an "attention" mechanism, as well as being trained on more datasets than the other methods, ended up with the best performance.

These results are in line with \cite{Jahn2025} who used Percentage of Correct Keypoints (PCK), an accuracy-based evaluation that can be compared to our Neck-MidHip errors to some extent, as they also base it on the infant's approximate torso length. The performance rank order of the methods we commonly compare are the same, with the identical conclusion that ViTPose performs the best. The error levels also converge towards similar levels: they find that 84.6\% of the ViTPose estimates are equal or below 10\% of the infant torso length, while we find that 68\% of the ViTPose estimates are within 3.3\% to 8.7\% of the infant torso length. However, compared to \cite{Jahn2025}, who only used this single accuracy-based evaluation, PCK, we complemented our results using several accuracy-based metrics, including the standard AP and AR, with other analyses such as the amount of missing data or of produced redundant detection, the relation between scores and confidence values with accuracy-based performance, as well as the processing speed of each method. We also included the methods that we observed to be often used by infant development researchers (AlphaPose and DeepLabCut in particular, in addition to OpenPose).

On the more complex ``Lap'' and SyRIP datasets, ViTPose and either OpenPose or HRNet depending on the metric and dataset have the highest performance. However, for all methods, the drop in performance is drastic, for example losing between 32 to 65 points in Average Precision and no method reaching errors lower than an average of 12\% of the infants' torso length.

Second, we complemented accuracy-based performance analysis with additional criteria that are important for the practical applications of the methods. The accuracy (OKS, Neck-midhip errors) cannot be computed if keypoints are not detected and does not penalize the case where multiple people are erroneously detected (though AP partially consider these cases). Moreover, both missing and redundant detections negatively impact any downstream motion analysis. Detectron2 and HRNet (top-down) are in particular susceptible to redundant detections. OpenPose, MediaPipe and AlphaPose have high rates of missing data (missing detections of the whole infant or of individual keypoints).
In the case of infant-parent (dyadic) interaction settings, such as in our ``Lap'' dataset, disentangling individual detections is not trivial, even when using dual RGB-D camera setup in a laboratory setting \cite{Diaz-Rojas2024}. One reason for this is that the methods are biased towards adults instead of infants, and will sometimes only detect the parent. This is particularly noticeable when looking at methods that are unable to provide as many detections as expected in this context; see Tab.~\ref{tab:totalrdet}.

It is also important to consider both accuracy-based metrics (AP/AR, Neck-MidHip, OSK) together with higher level metrics (missing data, redundant detections): for example, OpenPose, MediaPipe and AlphaPose have high amounts of missing data. Missing keypoints, in particular, which happen for bottom-up methods either because the methods cannot detect them in the image, or because their tentative placement was too far from the model's priors and were ignored, can not be evaluated by accuracy-based errors. The most striking example might be OpenPose, a Bottom-Up method, which misses up to 54\% of the data, mainly as missing keypoints, in the ``Lap'' dataset, yet manages to reach the second highest AP values and the second lowest average Neck-MidHip errors for this dataset.
In contrast, top-down methods would have a tendency to provide all keypoints even when misplaced, giving a lower confidence or score value to flag this. While confidence thresholds can be arbitrarily chosen to remove those keypoints, which some methods do internally, we decided to not set such an arbitrary threshold and to not remove them in our results for the methods that did not do so by default internally, making these keypoints impact the accuracy-based metrics instead of showing as missing data in our results. 

For this purpose, we introduced a Combined Performance Evaluation (CPE) that merges accuracy-based Neck-MidHip errors with the percentage of missing data into a single value. OpenPose drops to being the worst performing method on the ``Lap'' dataset in this Combined Performance Evaluation. Using this evaluation, we find again that ViTPose displays the best performance except for the ``Lap'' dataset, most likely due to a strong bias towards the detection of the adult parent.

Third, we evaluated the computation time of all the methods on the same computer. On this machine, which can be considered a powerful workstation in 2024, most of the methods could process less than 15 frames per second. AlphaPose was an exception; with approximately 27 fps, it could be deployed if a real-time pose estimation is necessary. On the other hand, any application that does not need real-time processing would not care about this, and rather focus on the most accurate method, ViTPose, even if it is much slower.

Finally, we make available most pose estimation methods including complete environments in public Docker containers (at \url{https://hub.docker.com/u/humanoidsctu}). In addition, we also share the results evaluation scripts and the detailed results (at \url{https://osf.io/x465b/}). 

\section*{Discussion and Future Work}
\label{sec:disc_and_future}

Researchers in behavioral science are not experts in machine learning and computer vision and cannot keep up with the rapid progress in human pose estimation methods. Hence, they choose their tools based on methods that have been used in the past (in the context of infants, OpenPose was used in \cite{Kojovic2021, reich2021, Chambers2020}; AlphaPose in \cite{Shin2022}). The DeepLabCut environment \cite{Mathis2018_DeepLabCut} is a frequent tool of choice in behavioral science. Here we show that for the estimation of the pose of infants in supine position, DeepLabCut with human pose estimation using \textit{DeeperCut}~\cite{Insafutdino2016_DeeperCut} currently does not provide satisfactory performance at all. OpenPose and AlphaPose may be employed but fall behind ViTPose and HRNet (top-down) in accuracy, as well as percentage of missing detections of infants or body keypoints.

While fine-tuning these methods with infant data can improve their performance \cite{Jahn2025, Groos2022, Chambers2020}, due to the limited amount of annotated infant data, it seems to easily and quickly overfit the models to the specificities of the training subset and to reduce the performance when used on other image sets \cite{Jahn2025}. The amount of ``difference'' that leads to under performance after fine-tuning is difficult to quantify due to the nature of these models (``black-box'' deep learning networks), and depends on the quantity of available data and how the training dataset is separated from the rest of the data on which the model will be applied to. We recommend the community to use the currently best-performing tools and profit from the Docker containers we released. 

We would like to emphasize that the human pose estimation tools are unfortunately not standard computer programs that can be easily deployed and used ``as is''. Their installation may be more or less involved. Moreover, performance is affected by preprocessing steps and settings. For example, cropping images so that only the infant is in the image helps to improve performances. Images with multiple persons in the scene (e.g., caregiver or experimenter), especially in close proximity, are currently challenging, and body keypoints of adults and the infant can even get mixed up. Also, as the pose estimation networks were predominantly trained on people in upright positions, the orientation of the infant image importantly affects the performance---the camera should be positioned, or every frame rotated afterwards, so that the infant is positioned approximately with the head at the top.
Some pose estimation networks may have additional settings (e.g., OpenPose and MediaPipe allow to set the maximum number of persons to detect in the image).

There are also differences between keypoint estimations within a single method. Some of these differences come from postural differences between infants and the datasets on which the methods were trained (featuring almost exclusively adults)---for example when legs are brought upwards, or are bent or crossed, as in Fig.~\ref{fig:pem_failings}, e). Other differences could reflect the accuracy of the human annotators (see the article and website for the COCO dataset \cite{Lin_2014_COCO} and how the $\sigma_k$ specific to each keypoint used for OKS were computed based on the variability between human annotators, as well as the differences we find between our two annotators). Errors can also occur naturally as an interaction between body posture and camera angle: in supine position and at younger ages, the head will have a tendency to be tilted in one direction rather than the other, which can lead to the occlusion of one ear that results either in a missing keypoint or in a less accurate estimation. Bottom-Up methods seem more prone to provide misestimations over the whole body when a subset of keypoints are suddenly misestimated or missing (e.g., the infant bringing its legs upwards and crossing the feet can lead to misestimation of the upper body).
Our impression is that pose estimation methods seem to be able to estimate locations that are occluded but that could be guessed by a human annotator: for example, if the right arm is flexed and the right shoulder is hidden by the forearm or the hand, the shoulder keypoint will still be estimated with good enough accuracy.

We have observed issues when estimating images with higher luminance, on recordings where the infant was near windows during a sunny day. We have observed difficulties when the infant and the background are less contrasted, for example with white clothes on the infant with a white sheet under it, but we have also observed cases where the estimations looked accurate with infants having black clothes with a black sheet under (most likely because the cloth did not cover the lower legs and arms, and the infant's skin color provided enough contrast). In other cases, we observed higher estimation errors when the camera angle was too low or too much on the side compared to a view from above. The distance between the camera and the infant and the zoom level also play a role \cite{koul2025accurately}, as well as the resolution (in particular if the infant has a pacifier).

For subsequent motion analysis or any downstream tasks, one may take advantage of the internal confidence estimates about the detected keypoints in every frame. However, we found that the correlation between the score for the complete keypoint set and the actual accuracy is moderate (or even low for Detectron2 and AlphaPose) and should therefore be used with caution. If specific keypoints are needed for subsequent analysis, the correlation between the confidence values and the actual errors in their estimation could be analyzed. 

Pose estimation methods can take either videos or individual images as input (or both). Although this may seem equivalent at first sight---a video is a sequence of consecutive images---, we realized that we sometimes obtained quite different results for these two input modes. The differences might partially come from how each method cuts the videos into frames, which can lead to slight offsets compared to the ground-truth sequences we used for evaluation. However, this alone cannot explain the larger differences in performance for OpenPose or the difference in the number of redundant detections for Detectron 2. Some of the pose estimation methods have divergent processing paths for image and video processing. In particular, \textit{tracking}, a step that exploits the temporal consistency of estimate over the sequence of detections, when implemented, is often enabled only for video input and not for images. With the exception of MediaPipe, for which we enabled it for both input modes in the ``Supine'' and MINI-RGBD datasets, none of the methods we tested had a clear tracking option in its parameters. Tracking can be currently achieved in postprocessing, as not all methods contain such a step, but in the future it is likely to find its way into the pose estimation methods themselves.   

The top-down methods, which identify areas (bounding boxes) where persons are likely to be in the images first, and then look for keypoints on the body within these areas, require a (bounding-box) \textit{detector} that provides the coordinates of this bounding box, in theory encompassing all the visible body parts of detected individuals. They are the main source of redundant and missing detections for these methods. Detectors are separate from the methods but often embedded within them, and some work might be needed to compare their performance in detecting infants.
For example, we hypothesize that the poor performance of MediaPipe is influenced by its detector, which was made to detect (adult) faces instead of bodies and assumes that the body is directly below, as a compromise to increase processing speed.
We can also observe that on our ``Lap'' dataset, it is the bottom-up methods, HRNet BU and OpenPose that have the lowest Neck-MidHip errors, while all top-down methods show much higher errors.
Another example is by looking at ViTPose's reported results on the OCHuman dataset: a dataset comprised of multi-body scenes with adults. The AP for ViTPose-b is reported to be 87.3 when using the ground truth bounding boxes (\cite{xu2022_vitpose} Table 11), while another work comparing the performance of several detectors on this dataset showed that the AP of ViTPose-b using the RTMDet detector drops to 45.0; and even the best detector only reached 48.3 AP on this dataset (\cite{Purkrabek2024}, Table 1).

In this work, the ``vanilla'' versions of pose estimation methods, the models trained on datasets featuring mostly adults, were directly used to process videos with infants in supine position. Contrary to expectations, the drop in performance was not dramatic. In fact, comparing the average precision and recall (AP, AR) results with those reported for these methods on the COCO validation or the test datasets, or on the ``body'' part of the Halpe and COCO-WholeBody datasets \cite{xu2022_vitpose, Fang2022_Alphapose, xiao2018_hrnet, sun2019deep_hrnet}, it seems that VitPose performs better on real infants in supine position than reported on adults by around 9 and 7 points respectively. OpenPose also performs better than reported by around 10 points in both AP and AR when processing images, but worse when processing videos. For AlphaPose and HRNet BU, the results are roughly similar. For HRNet TD, its AP is at least 10 points lower, most likely due to higher amounts of high-scoring redundant detections, although its AR is around 8 points higher.

Such results could be explained by the dataset containing infants only in supine position with a view from above, which, as long as the infants do not perform complex leg movements, is rather simple and could be considered similar enough to an adult standing in front of a camera: the situation that these methods encounter most commonly during their training. This is supported by looking at the average OKS values from individual videos (Supp. Materials ST. 6), where the performance of all methods decreases noticeably and is more variable for ``hard'' synthetic infants (IDs 10-12) compared to their performance on ``easy'' infants (IDs 1-4).
Furthermore, using more complex datasets, such as SyRIP or our ``Lap'' dataset, shows large drops of performance at all levels. The ``Lap'' condition, which would fall under a multi-body detection scenario, is challenging even when only adults are involved, partially due to the poor performance of detectors for top-down methods \cite{Purkrabek2024}. In our results, ViTPose-h performs 15 AP points lower in our ``Lap'' dataset compared to ViTPose-b on the OCHuman dataset, which might highlight a bias toward detecting adults rather than infants when both are present in a scene together due to the training datasets only containing adults. It also scores about 40 points lower in AP on the SyRIP dataset compared to the common benchmark dataset for adults, COCO.

There are several directions for future work. First, although small infants (under seven months) differ significantly in their body proportions from adults and are thus strongly ``out of distribution'' data for the models in terms of shape estimation, pose estimation may be relatively easier because infants are in supine position all the time and their motor repertoire is constrained. Older infants and children in prone, sitting, standing, and other postures which are partially covered in the SyRIP and our ``Lap'' datasets should be considered as a future challenge for pose estimation, and be included in the training of future methods considering the difficulty for current methods to handle such postures. The situation is similar for scenes with multiple people, like infants on a parent's lap. Second, in addition to 2D pose estimation in the form of keypoints, there are methods that estimate 3D pose and shape (a complete body mesh) from an image or video (e.g., 
Smplify-x \cite{Pavlakos2019_SMPL-X}, with SMIL for an infant shape~\cite{Hesse2018_SMIL}, or 4D Humans currently with adult shape only \cite{goel2023_humans4D}). Third, while pose estimation is a key prerequisite for additional analyzes, the quality of motion extracted from the sequences of estimated keypoints should be studied in detail (see also \cite{seethapathi2019movement}).

Pose estimation methods have been used in clinical movement assessment of infants (OpenPose outputs used for automated GMA scoring in~\cite{reich2021} and for ASD prediction in \cite{Kojovic2021}; AlphaPose used in~\cite{Shin2022} as input for subsequent classification using HINE). Here, we show that with ViTPose, the errors are considerably smaller (about 25\% reduced magnitude of Neck-MidHip errors compared to OpenPose and AlphaPose for infants in supine position, and little to no missing data). Therefore, using state-of-the-art methods, together with the code and additional recommendations on the processing we provide here, holds promise for improving automated infant clinical assessment in the very near future.

Unlike for adults, datasets of infant video recordings with additional reference (e.g., from motion capture, or manually-annotated) are scarce and often kept private. We are open to collaborate with research groups that are willing to share such datasets. We would also be happy to share our experience on how to best record infants such that the pose estimation result is optimal.

\backmatter

\bmhead{Supplementary Information}
This work has supplementary materials that include additional details about the pose estimation methods and datasets as well as additional results and analysis.

\bmhead{Acknowledgements}
We thank Miroslav Purkrábek and Pramod Murthy for their advice, and Lukáš Rustler, Valentin Marcel, Jason Khoury, Vojtěch Ježek, and Vojtěch Volprecht for their assistance.

\section*{Declarations}

\subsection*{Funding}
This work was supported by the Czech Science Foundation (GA CR), project no. 20-24186X and project no. 25-18113S. F.G. was additionally supported by the Grant Agency of the Czech Technical University in Prague, grant No. SGS24/096/OHK3/2T/13. S.T.P was additionally supported by the project Mobility ČVUT MSCA-F-CZ-I  CZ.02.01.01/00/22\_010/0003405.

\subsection*{Conflicts of interest/Competing interests}
The authors declare no competing interests.

\subsection*{Ethics approval and consent to participate and for publication}
The current study used video recordings of two full-term healthy infants (1f, 1m), observed between 8 weeks and and 25 weeks of age in supine position, and two full-term healthy infants (2f) observed at 4 months and 1 year and 3 months of age sitting on the lap of a parent. Informed consent was obtained from the participants' legal guardians, including consent for publication of images in an online open-access publication. The study was approved by the Committee for Research Ethics at the Czech Technical University in Prague under reference number 00000-07/21/51903/EKCVUT and was carried out in accordance with all relevant guidelines and regulations. We additionally included video recordings of four full-term infants (2f, 2m), observed between 6 and 18 months of age sitting or standing in front of their parents. Informed consent was obtained from the participants' legal guardians. The study was approved by the Pavia Ethics Committee (p–20190004177). 

\subsection*{Availability of data, code, and materials}

Two of the datasets used (MINI-RGBD \cite{Hesse2018_minirgbd} and SyRIP \cite{Huang2021_syrip}) are publicly available in their respective public repositories. Video recordings and images from our infant datasets (``Supine'' and ``Lap'') cannot be publicly shared due to data protection and sharing rules.

We publicly share the Docker images and Dockerfiles of the pose estimation methods that we created and used, all the program scripts used to compute the analyses, as well as the anonymized sections of our datasets (estimated keypoints) and the analysis output files at the following locations: \url{https://osf.io/x465b/} and \url{https://hub.docker.com/u/humanoidsctu}.  DeepLabCut’s official installation instructions include creating its own conda environment, while MediaPipe is already its own easy-to-install Python library, and as such did not need to be further containerized with Docker, although we provide Python scripts and instructions for utility. Due to a technical error during upload, the OpenPose container could not be uploaded to DockerHub at the time of writing.
We do not share the direct outputs of the methods due to the sheer amount of files, but their content is available in the provided resources in a compact and easy-to-use format.

\subsection*{Author contribution}
The work was conceived by F.G. and M.H. The data was collected by M.H., L.N. and F.G. The data was processed by F.G, M.M. and L.N. The data was analyzed by F.G. The manuscript was prepared by F.G. and revised and reviewed by M.H. and S.T.P. Funding was acquired by M.H.

\bibliography{bibliography}


\begin{thebibliography}{64}
\ifx \bisbn   \undefined \def \bisbn  #1{ISBN #1}\fi
\ifx \binits  \undefined \def \binits#1{#1}\fi
\ifx \bauthor  \undefined \def \bauthor#1{#1}\fi
\ifx \batitle  \undefined \def \batitle#1{#1}\fi
\ifx \bjtitle  \undefined \def \bjtitle#1{#1}\fi
\ifx \bvolume  \undefined \def \bvolume#1{\textbf{#1}}\fi
\ifx \byear  \undefined \def \byear#1{#1}\fi
\ifx \bissue  \undefined \def \bissue#1{#1}\fi
\ifx \bfpage  \undefined \def \bfpage#1{#1}\fi
\ifx \blpage  \undefined \def \blpage #1{#1}\fi
\ifx \burl  \undefined \def \burl#1{\textsf{#1}}\fi
\ifx \doiurl  \undefined \def \doiurl#1{\url{https://doi.org/#1}}\fi
\ifx \betal  \undefined \def \betal{\textit{et al.}}\fi
\ifx \binstitute  \undefined \def \binstitute#1{#1}\fi
\ifx \binstitutionaled  \undefined \def \binstitutionaled#1{#1}\fi
\ifx \bctitle  \undefined \def \bctitle#1{#1}\fi
\ifx \beditor  \undefined \def \beditor#1{#1}\fi
\ifx \bpublisher  \undefined \def \bpublisher#1{#1}\fi
\ifx \bbtitle  \undefined \def \bbtitle#1{#1}\fi
\ifx \bedition  \undefined \def \bedition#1{#1}\fi
\ifx \bseriesno  \undefined \def \bseriesno#1{#1}\fi
\ifx \blocation  \undefined \def \blocation#1{#1}\fi
\ifx \bsertitle  \undefined \def \bsertitle#1{#1}\fi
\ifx \bsnm \undefined \def \bsnm#1{#1}\fi
\ifx \bsuffix \undefined \def \bsuffix#1{#1}\fi
\ifx \bparticle \undefined \def \bparticle#1{#1}\fi
\ifx \barticle \undefined \def \barticle#1{#1}\fi
\bibcommenthead
\ifx \bconfdate \undefined \def \bconfdate #1{#1}\fi
\ifx \botherref \undefined \def \botherref #1{#1}\fi
\ifx \url \undefined \def \url#1{\textsf{#1}}\fi
\ifx \bchapter \undefined \def \bchapter#1{#1}\fi
\ifx \bbook \undefined \def \bbook#1{#1}\fi
\ifx \bcomment \undefined \def \bcomment#1{#1}\fi
\ifx \oauthor \undefined \def \oauthor#1{#1}\fi
\ifx \citeauthoryear \undefined \def \citeauthoryear#1{#1}\fi
\ifx \endbibitem  \undefined \def \endbibitem {}\fi
\ifx \bconflocation  \undefined \def \bconflocation#1{#1}\fi
\ifx \arxivurl  \undefined \def \arxivurl#1{\textsf{#1}}\fi
\csname PreBibitemsHook\endcsname

\bibitem[\protect\citeauthoryear{DiMercurio et~al.}{2018}]{DiMercurio2018}
\begin{barticle}
\bauthor{\bsnm{DiMercurio}, \binits{A.}},
\bauthor{\bsnm{Connell}, \binits{J.P.}},
\bauthor{\bsnm{Clark}, \binits{M.}},
\bauthor{\bsnm{Corbetta}, \binits{D.}}:
\batitle{A naturalistic observation of spontaneous touches to the body and
  environment in the first 2 months of life}.
\bjtitle{Frontiers in psychology}
\bvolume{9},
\bfpage{2613}
(\byear{2018})
\end{barticle}
\endbibitem

\bibitem[\protect\citeauthoryear{Sloan et~al.}{2023}]{sloan2023meaning}
\begin{barticle}
\bauthor{\bsnm{Sloan}, \binits{A.T.}},
\bauthor{\bsnm{Jones}, \binits{N.A.}},
\bauthor{\bsnm{Kelso}, \binits{J.S.}}:
\batitle{Meaning from movement and stillness: Signatures of coordination
  dynamics reveal infant agency}.
\bjtitle{Proceedings of the National Academy of Sciences}
\bvolume{120}(\bissue{39}),
\bfpage{2306732120}
(\byear{2023})
\end{barticle}
\endbibitem

\bibitem[\protect\citeauthoryear{Kanazawa et~al.}{2023}]{kanazawa2023open}
\begin{barticle}
\bauthor{\bsnm{Kanazawa}, \binits{H.}},
\bauthor{\bsnm{Yamada}, \binits{Y.}},
\bauthor{\bsnm{Tanaka}, \binits{K.}},
\bauthor{\bsnm{Kawai}, \binits{M.}},
\bauthor{\bsnm{Niwa}, \binits{F.}},
\bauthor{\bsnm{Iwanaga}, \binits{K.}},
\bauthor{\bsnm{Kuniyoshi}, \binits{Y.}}:
\batitle{Open-ended movements structure sensorimotor information in early human
  development}.
\bjtitle{Proceedings of the National Academy of Sciences}
\bvolume{120}(\bissue{1}),
\bfpage{2209953120}
(\byear{2023})
\end{barticle}
\endbibitem

\bibitem[\protect\citeauthoryear{Adolph et~al.}{2008}]{Adolph2008}
\begin{barticle}
\bauthor{\bsnm{Adolph}, \binits{K.E.}},
\bauthor{\bsnm{Robinson}, \binits{S.R.}},
\bauthor{\bsnm{Young}, \binits{J.W.}},
\bauthor{\bsnm{Gill-Alvarez}, \binits{F.}}:
\batitle{What is the shape of developmental change?}
\bjtitle{Psychological Review}
\bvolume{115}(\bissue{3}),
\bfpage{527}--\blpage{543}
(\byear{2008})
\doiurl{10.1037/0033-295X.115.3.527}
\end{barticle}
\endbibitem

\bibitem[\protect\citeauthoryear{Adolph et~al.}{2017}]{adolph2017video}
\begin{botherref}
\oauthor{\bsnm{Adolph}, \binits{K.}},
\oauthor{\bsnm{Gilmore}, \binits{R.}},
\oauthor{\bsnm{Kennedy}, \binits{J.}}:
Video data and documentation will improve psychological science.
Psychological Science Agenda
\textbf{31}(10)
(2017)
\end{botherref}
\endbibitem

\bibitem[\protect\citeauthoryear{Gilmore and Adolph}{2017}]{gilmore2017video}
\begin{barticle}
\bauthor{\bsnm{Gilmore}, \binits{R.O.}},
\bauthor{\bsnm{Adolph}, \binits{K.E.}}:
\batitle{Video can make behavioural science more reproducible}.
\bjtitle{Nature Human Behaviour}
\bvolume{1}(\bissue{7}),
\bfpage{0128}
(\byear{2017})
\end{barticle}
\endbibitem

\bibitem[\protect\citeauthoryear{Berthier and
  Keen}{2006}]{berthier2006development}
\begin{barticle}
\bauthor{\bsnm{Berthier}, \binits{N.E.}},
\bauthor{\bsnm{Keen}, \binits{R.}}:
\batitle{Development of reaching in infancy}.
\bjtitle{Experimental Brain Research}
\bvolume{169},
\bfpage{507}--\blpage{518}
(\byear{2006})
\end{barticle}
\endbibitem

\bibitem[\protect\citeauthoryear{Konczak and
  Dichgans}{1997}]{konczak1997development}
\begin{barticle}
\bauthor{\bsnm{Konczak}, \binits{J.}},
\bauthor{\bsnm{Dichgans}, \binits{J.}}:
\batitle{The development toward stereotypic arm kinematics during reaching in
  the first 3 years of life}.
\bjtitle{Experimental Brain Research}
\bvolume{117},
\bfpage{346}--\blpage{354}
(\byear{1997})
\end{barticle}
\endbibitem

\bibitem[\protect\citeauthoryear{Stupperich
  et~al.}{2024}]{stupperich2024quantity}
\begin{barticle}
\bauthor{\bsnm{Stupperich}, \binits{S.}},
\bauthor{\bsnm{Dathe}, \binits{A.-K.}},
\bauthor{\bsnm{DiMercurio}, \binits{A.}},
\bauthor{\bsnm{Connell}, \binits{J.P.}},
\bauthor{\bsnm{Baumann}, \binits{N.}},
\bauthor{\bsnm{Jover}, \binits{M.}},
\bauthor{\bsnm{Corbetta}, \binits{D.}},
\bauthor{\bsnm{Jaekel}, \binits{J.}},
\bauthor{\bsnm{Felderhoff-Mueser}, \binits{U.}},
\bauthor{\bsnm{Huening}, \binits{B.}}:
\batitle{Quantity of spontaneous touches to body and surface in very preterm
  and healthy term infants}.
\bjtitle{Frontiers in Psychology}
\bvolume{15},
\bfpage{1459009}
(\byear{2024})
\end{barticle}
\endbibitem

\bibitem[\protect\citeauthoryear{Thomas et~al.}{2015}]{thomas2015independent}
\begin{barticle}
\bauthor{\bsnm{Thomas}, \binits{B.L.}},
\bauthor{\bsnm{Karl}, \binits{J.M.}},
\bauthor{\bsnm{Whishaw}, \binits{I.Q.}}:
\batitle{Independent development of the reach and the grasp in spontaneous
  self-touching by human infants in the first 6 months}.
\bjtitle{Frontiers in psychology}
\bvolume{5},
\bfpage{1526}
(\byear{2015})
\end{barticle}
\endbibitem

\bibitem[\protect\citeauthoryear{Khoury et~al.}{2022}]{Khoury2022}
\begin{bchapter}
\bauthor{\bsnm{Khoury}, \binits{J.}},
\bauthor{\bsnm{Popescu}, \binits{S.T.}},
\bauthor{\bsnm{Gama}, \binits{F.}},
\bauthor{\bsnm{Marcel}, \binits{V.}},
\bauthor{\bsnm{Hoffmann}, \binits{M.}}:
\bctitle{Self-touch and other spontaneous behavior patterns in early infancy}.
In: \bbtitle{2022 IEEE International Conference on Development and Learning
  (ICDL)},
pp. \bfpage{148}--\blpage{155}
(\byear{2022}).
\bcomment{IEEE}
\end{bchapter}
\endbibitem

\bibitem[\protect\citeauthoryear{Einspieler and
  Prechtl}{2005}]{einspieler2005prechtl}
\begin{barticle}
\bauthor{\bsnm{Einspieler}, \binits{C.}},
\bauthor{\bsnm{Prechtl}, \binits{H.F.}}:
\batitle{Prechtl's assessment of general movements: a diagnostic tool for the
  functional assessment of the young nervous system}.
\bjtitle{Mental retardation and developmental disabilities research reviews}
\bvolume{11}(\bissue{1}),
\bfpage{61}--\blpage{67}
(\byear{2005})
\end{barticle}
\endbibitem

\bibitem[\protect\citeauthoryear{Romeo et~al.}{2016}]{romeo2016use}
\begin{barticle}
\bauthor{\bsnm{Romeo}, \binits{D.M.}},
\bauthor{\bsnm{Ricci}, \binits{D.}},
\bauthor{\bsnm{Brogna}, \binits{C.}},
\bauthor{\bsnm{Mercuri}, \binits{E.}}:
\batitle{Use of the {Hammersmith Infant Neurological Examination} in infants
  with cerebral palsy: a critical review of the literature}.
\bjtitle{Developmental Medicine \& Child Neurology}
\bvolume{58}(\bissue{3}),
\bfpage{240}--\blpage{245}
(\byear{2016})
\end{barticle}
\endbibitem

\bibitem[\protect\citeauthoryear{Chen et~al.}{2012}]{chen2012sensor}
\begin{barticle}
\bauthor{\bsnm{Chen}, \binits{L.}},
\bauthor{\bsnm{Hoey}, \binits{J.}},
\bauthor{\bsnm{Nugent}, \binits{C.D.}},
\bauthor{\bsnm{Cook}, \binits{D.J.}},
\bauthor{\bsnm{Yu}, \binits{Z.}}:
\batitle{Sensor-based activity recognition}.
\bjtitle{IEEE Transactions on Systems, Man, and Cybernetics, Part C
  (Applications and Reviews)}
\bvolume{42}(\bissue{6}),
\bfpage{790}--\blpage{808}
(\byear{2012})
\end{barticle}
\endbibitem

\bibitem[\protect\citeauthoryear{Chen et~al.}{2016}]{chen2016review}
\begin{barticle}
\bauthor{\bsnm{Chen}, \binits{H.}},
\bauthor{\bsnm{Xue}, \binits{M.}},
\bauthor{\bsnm{Mei}, \binits{Z.}},
\bauthor{\bsnm{Bambang~Oetomo}, \binits{S.}},
\bauthor{\bsnm{Chen}, \binits{W.}}:
\batitle{A review of wearable sensor systems for monitoring body movements of
  neonates}.
\bjtitle{Sensors}
\bvolume{16}(\bissue{12}),
\bfpage{2134}
(\byear{2016})
\end{barticle}
\endbibitem

\bibitem[\protect\citeauthoryear{Heinze et~al.}{2010}]{heinze2010movement}
\begin{barticle}
\bauthor{\bsnm{Heinze}, \binits{F.}},
\bauthor{\bsnm{Hesels}, \binits{K.}},
\bauthor{\bsnm{Breitbach-Faller}, \binits{N.}},
\bauthor{\bsnm{Schmitz-Rode}, \binits{T.}},
\bauthor{\bsnm{Disselhorst-Klug}, \binits{C.}}:
\batitle{Movement analysis by accelerometry of newborns and infants for the
  early detection of movement disorders due to infantile cerebral palsy}.
\bjtitle{Medical \& biological engineering \& computing}
\bvolume{48},
\bfpage{765}--\blpage{772}
(\byear{2010})
\end{barticle}
\endbibitem

\bibitem[\protect\citeauthoryear{Airaksinen
  et~al.}{2020}]{airaksinen2020automatic}
\begin{barticle}
\bauthor{\bsnm{Airaksinen}, \binits{M.}},
\bauthor{\bsnm{R{\"a}s{\"a}nen}, \binits{O.}},
\bauthor{\bsnm{Il{\'e}n}, \binits{E.}},
\bauthor{\bsnm{H{\"a}yrinen}, \binits{T.}},
\bauthor{\bsnm{Kivi}, \binits{A.}},
\bauthor{\bsnm{Marchi}, \binits{V.}},
\bauthor{\bsnm{Gallen}, \binits{A.}},
\bauthor{\bsnm{Blom}, \binits{S.}},
\bauthor{\bsnm{Varhe}, \binits{A.}},
\bauthor{\bsnm{Kaartinen}, \binits{N.}}, \betal:
\batitle{Automatic posture and movement tracking of infants with wearable
  movement sensors}.
\bjtitle{Scientific Reports}
\bvolume{10}(\bissue{1}),
\bfpage{169}
(\byear{2020})
\end{barticle}
\endbibitem

\bibitem[\protect\citeauthoryear{Kanemaru et~al.}{2013}]{kanemaru2013specific}
\begin{barticle}
\bauthor{\bsnm{Kanemaru}, \binits{N.}},
\bauthor{\bsnm{Watanabe}, \binits{H.}},
\bauthor{\bsnm{Kihara}, \binits{H.}},
\bauthor{\bsnm{Nakano}, \binits{H.}},
\bauthor{\bsnm{Takaya}, \binits{R.}},
\bauthor{\bsnm{Nakamura}, \binits{T.}},
\bauthor{\bsnm{Nakano}, \binits{J.}},
\bauthor{\bsnm{Taga}, \binits{G.}},
\bauthor{\bsnm{Konishi}, \binits{Y.}}:
\batitle{Specific characteristics of spontaneous movements in preterm infants
  at term age are associated with developmental delays at age 3 years}.
\bjtitle{Developmental Medicine \& Child Neurology}
\bvolume{55}(\bissue{8}),
\bfpage{713}--\blpage{721}
(\byear{2013})
\end{barticle}
\endbibitem

\bibitem[\protect\citeauthoryear{Needham et~al.}{2021}]{needham2021accuracy}
\begin{barticle}
\bauthor{\bsnm{Needham}, \binits{L.}},
\bauthor{\bsnm{Evans}, \binits{M.}},
\bauthor{\bsnm{Cosker}, \binits{D.P.}},
\bauthor{\bsnm{Wade}, \binits{L.}},
\bauthor{\bsnm{McGuigan}, \binits{P.M.}},
\bauthor{\bsnm{Bilzon}, \binits{J.L.}},
\bauthor{\bsnm{Colyer}, \binits{S.L.}}:
\batitle{The accuracy of several pose estimation methods for {3D} joint centre
  localisation}.
\bjtitle{Scientific Reports}
\bvolume{11}(\bissue{1}),
\bfpage{20673}
(\byear{2021})
\end{barticle}
\endbibitem

\bibitem[\protect\citeauthoryear{Shin et~al.}{2022}]{Shin2022}
\begin{botherref}
\oauthor{\bsnm{Shin}, \binits{H.I.}},
\oauthor{\bsnm{Shin}, \binits{H.-I.}},
\oauthor{\bsnm{Bang}, \binits{M.S.}},
\oauthor{\bsnm{Kim}, \binits{D.-K.}},
\oauthor{\bsnm{Shin}, \binits{S.H.}},
\oauthor{\bsnm{Kim}, \binits{E.-K.}},
\oauthor{\bsnm{Kim}, \binits{Y.-J.}},
\oauthor{\bsnm{Lee}, \binits{E.S.}},
\oauthor{\bsnm{Park}, \binits{S.G.}},
\oauthor{\bsnm{Ji}, \binits{H.M.}},
\oauthor{\bsnm{Lee}, \binits{W.H.}}:
Deep learning-based quantitative analyses of spontaneous movements and their
  association with early neurological development in preterm infants.
Scientific Reports
\textbf{12}
(2022)
\end{botherref}
\endbibitem

\bibitem[\protect\citeauthoryear{Marcroft et~al.}{2015}]{marcroft2015movement}
\begin{barticle}
\bauthor{\bsnm{Marcroft}, \binits{C.}},
\bauthor{\bsnm{Khan}, \binits{A.}},
\bauthor{\bsnm{Embleton}, \binits{N.D.}},
\bauthor{\bsnm{Trenell}, \binits{M.}},
\bauthor{\bsnm{Pl{\"o}tz}, \binits{T.}}:
\batitle{Movement recognition technology as a method of assessing spontaneous
  general movements in high risk infants}.
\bjtitle{Frontiers in neurology}
\bvolume{5},
\bfpage{284}
(\byear{2015})
\end{barticle}
\endbibitem

\bibitem[\protect\citeauthoryear{Silva et~al.}{2021}]{Silva2021}
\begin{barticle}
\bauthor{\bsnm{Silva}, \binits{N.}},
\bauthor{\bsnm{Zhang}, \binits{D.}},
\bauthor{\bsnm{Kulvicius}, \binits{T.}},
\bauthor{\bsnm{Gail}, \binits{A.}},
\bauthor{\bsnm{Barreiros}, \binits{C.}},
\bauthor{\bsnm{Lindstaedt}, \binits{S.}},
\bauthor{\bsnm{Kraft}, \binits{M.}},
\bauthor{\bsnm{Bölte}, \binits{S.}},
\bauthor{\bsnm{Poustka}, \binits{L.}},
\bauthor{\bsnm{Nielsen-Saines}, \binits{K.}},
\bauthor{\bsnm{Wörgötter}, \binits{F.}},
\bauthor{\bsnm{Einspieler}, \binits{C.}},
\bauthor{\bsnm{Marschik}, \binits{P.B.}}:
\batitle{The future of general movement assessment: The role of computer vision
  and machine learning – a scoping review}.
\bjtitle{Research in Developmental Disabilities}
\bvolume{110},
\bfpage{103854}
(\byear{2021})
\end{barticle}
\endbibitem

\bibitem[\protect\citeauthoryear{Irshad et~al.}{2020}]{irshad2020ai}
\begin{barticle}
\bauthor{\bsnm{Irshad}, \binits{M.T.}},
\bauthor{\bsnm{Nisar}, \binits{M.A.}},
\bauthor{\bsnm{Gouverneur}, \binits{P.}},
\bauthor{\bsnm{Rapp}, \binits{M.}},
\bauthor{\bsnm{Grzegorzek}, \binits{M.}}:
\batitle{{AI} approaches towards {P}rechtl’s assessment of general movements:
  A systematic literature review}.
\bjtitle{Sensors}
\bvolume{20}(\bissue{18}),
\bfpage{5321}
(\byear{2020})
\end{barticle}
\endbibitem

\bibitem[\protect\citeauthoryear{Meinecke et~al.}{2006}]{meinecke2006movement}
\begin{barticle}
\bauthor{\bsnm{Meinecke}, \binits{L.}},
\bauthor{\bsnm{Breitbach-Faller}, \binits{N.}},
\bauthor{\bsnm{Bartz}, \binits{C.}},
\bauthor{\bsnm{Damen}, \binits{R.}},
\bauthor{\bsnm{Rau}, \binits{G.}},
\bauthor{\bsnm{Disselhorst-Klug}, \binits{C.}}:
\batitle{Movement analysis in the early detection of newborns at risk for
  developing spasticity due to infantile cerebral palsy}.
\bjtitle{Human Movement Science}
\bvolume{25}(\bissue{2}),
\bfpage{125}--\blpage{144}
(\byear{2006})
\end{barticle}
\endbibitem

\bibitem[\protect\citeauthoryear{Tsuji et~al.}{2020}]{Tsuji2020}
\begin{barticle}
\bauthor{\bsnm{Tsuji}, \binits{T.}},
\bauthor{\bsnm{Nakashima}, \binits{S.}},
\bauthor{\bsnm{Hayashi}, \binits{H.}},
\bauthor{\bsnm{Soh}, \binits{Z.}},
\bauthor{\bsnm{Furui}, \binits{A.}},
\bauthor{\bsnm{Shibanoki}, \binits{T.}},
\bauthor{\bsnm{Shima}, \binits{K.}},
\bauthor{\bsnm{Shimatani}, \binits{K.}}:
\batitle{Markerless measurement and evaluation of general movements in
  infants}.
\bjtitle{Scientific Reports}
\bvolume{10}(\bissue{1}),
\bfpage{1422}
(\byear{2020})
\end{barticle}
\endbibitem

\bibitem[\protect\citeauthoryear{Kinoshita et~al.}{2020}]{Kinoshita2020}
\begin{barticle}
\bauthor{\bsnm{Kinoshita}, \binits{N.}},
\bauthor{\bsnm{Furui}, \binits{A.}},
\bauthor{\bsnm{Soh}, \binits{Z.}},
\bauthor{\bsnm{Hayashi}, \binits{H.}},
\bauthor{\bsnm{Shibanoki}, \binits{T.}},
\bauthor{\bsnm{Mori}, \binits{H.}},
\bauthor{\bsnm{Shimatani}, \binits{K.}},
\bauthor{\bsnm{Funabiki}, \binits{Y.}},
\bauthor{\bsnm{Tsuji}, \binits{T.}}:
\batitle{Longitudinal assessment of u-shaped and inverted u-shaped
  developmental changes in the spontaneous movements of infants via markerless
  video analysis}.
\bjtitle{Scientific Reports}
\bvolume{10}(\bissue{1}),
\bfpage{16827}
(\byear{2020})
\end{barticle}
\endbibitem

\bibitem[\protect\citeauthoryear{McCay et~al.}{2020}]{McCay2020}
\begin{barticle}
\bauthor{\bsnm{McCay}, \binits{K.D.}},
\bauthor{\bsnm{Ho}, \binits{E.S.L.}},
\bauthor{\bsnm{Shum}, \binits{H.P.H.}},
\bauthor{\bsnm{Fehringer}, \binits{G.}},
\bauthor{\bsnm{Marcroft}, \binits{C.}},
\bauthor{\bsnm{Embleton}, \binits{N.D.}}:
\batitle{Abnormal infant movements classification with deep learning on
  pose-based features}.
\bjtitle{IEEE Access}
\bvolume{8},
\bfpage{51582}--\blpage{51592}
(\byear{2020})
\end{barticle}
\endbibitem

\bibitem[\protect\citeauthoryear{{Cao} et~al.}{2019}]{Cao2019_Openpose}
\begin{botherref}
\oauthor{\bsnm{{Cao}}, \binits{Z.}},
\oauthor{\bsnm{{Hidalgo Martinez}}, \binits{G.}},
\oauthor{\bsnm{{Simon}}, \binits{T.}},
\oauthor{\bsnm{{Wei}}, \binits{S.}},
\oauthor{\bsnm{{Sheikh}}, \binits{Y.A.}}:
Openpose: Realtime multi-person {2D} pose estimation using part affinity
  fields.
IEEE Transactions on Pattern Analysis and Machine Intelligence
(2019)
\end{botherref}
\endbibitem

\bibitem[\protect\citeauthoryear{Chambers et~al.}{2020}]{Chambers2020}
\begin{barticle}
\bauthor{\bsnm{Chambers}, \binits{C.}},
\bauthor{\bsnm{Seethapathi}, \binits{N.}},
\bauthor{\bsnm{Saluja}, \binits{R.}},
\bauthor{\bsnm{Loeb}, \binits{H.}},
\bauthor{\bsnm{Pierce}, \binits{S.R.}},
\bauthor{\bsnm{Bogen}, \binits{D.K.}},
\bauthor{\bsnm{Prosser}, \binits{L.}},
\bauthor{\bsnm{Johnson}, \binits{M.J.}},
\bauthor{\bsnm{Kording}, \binits{K.P.}}:
\batitle{Computer vision to automatically assess infant neuromotor risk}.
\bjtitle{{IEEE} Transactions on Neural Systems and Rehabilitation Engineering}
\bvolume{28}(\bissue{11}),
\bfpage{2431}--\blpage{2442}
(\byear{2020})
\end{barticle}
\endbibitem

\bibitem[\protect\citeauthoryear{Reich et~al.}{2021}]{reich2021}
\begin{barticle}
\bauthor{\bsnm{Reich}, \binits{S.}},
\bauthor{\bsnm{Zhang}, \binits{D.}},
\bauthor{\bsnm{Kulvicius}, \binits{T.}},
\bauthor{\bsnm{Bölte}, \binits{S.}},
\bauthor{\bsnm{Nielsen-Saines}, \binits{K.}},
\bauthor{\bsnm{Pokorny}, \binits{F.B.}},
\bauthor{\bsnm{Peharz}, \binits{R.}},
\bauthor{\bsnm{Poustka}, \binits{L.}},
\bauthor{\bsnm{Wörgötter}, \binits{F.}},
\bauthor{\bsnm{Einspieler}, \binits{C.}},
\bauthor{\bsnm{Marschik}, \binits{P.B.}}:
\batitle{Novel {AI} driven approach to classify infant motor functions}.
\bjtitle{Scientific Reports}
\bvolume{11}(\bissue{1}),
\bfpage{9888}
(\byear{2021})
\end{barticle}
\endbibitem

\bibitem[\protect\citeauthoryear{Fang et~al.}{2022}]{Fang2022_Alphapose}
\begin{botherref}
\oauthor{\bsnm{Fang}, \binits{H.-S.}},
\oauthor{\bsnm{Li}, \binits{J.}},
\oauthor{\bsnm{Tang}, \binits{H.}},
\oauthor{\bsnm{Xu}, \binits{C.}},
\oauthor{\bsnm{Zhu}, \binits{H.}},
\oauthor{\bsnm{Xiu}, \binits{Y.}},
\oauthor{\bsnm{Li}, \binits{Y.-L.}},
\oauthor{\bsnm{Lu}, \binits{C.}}:
AlphaPose: Whole-Body Regional Multi-Person Pose Estimation and Tracking in
  Real-Time.
arXiv
(2022)
\end{botherref}
\endbibitem

\bibitem[\protect\citeauthoryear{Doi et~al.}{2022}]{Doi2022}
\begin{barticle}
\bauthor{\bsnm{Doi}, \binits{H.}},
\bauthor{\bsnm{Iijima}, \binits{N.}},
\bauthor{\bsnm{Furui}, \binits{A.}},
\bauthor{\bsnm{Soh}, \binits{Z.}},
\bauthor{\bsnm{Yonei}, \binits{R.}},
\bauthor{\bsnm{Shinohara}, \binits{K.}},
\bauthor{\bsnm{Iriguchi}, \binits{M.}},
\bauthor{\bsnm{Shimatani}, \binits{K.}},
\bauthor{\bsnm{Tsuji}, \binits{T.}}:
\batitle{Prediction of autistic tendencies at 18 months of age via markerless
  video analysis of spontaneous body movements in 4-month-old infants}.
\bjtitle{Scientific Reports}
\bvolume{12}(\bissue{1}),
\bfpage{18045}
(\byear{2022})
\end{barticle}
\endbibitem

\bibitem[\protect\citeauthoryear{Kojovic et~al.}{2021}]{Kojovic2021}
\begin{botherref}
\oauthor{\bsnm{Kojovic}, \binits{N.}},
\oauthor{\bsnm{Shreyasvi}, \binits{N.}},
\oauthor{\bsnm{Mohanty}, \binits{S.P.}},
\oauthor{\bsnm{Maillart}, \binits{T.}},
\oauthor{\bsnm{Schaer}, \binits{M.}}:
Using {2D} video-based pose estimation for automated prediction of autism
  spectrum disorders in young children.
Scientific Reports
\textbf{11}
(2021)
\end{botherref}
\endbibitem

\bibitem[\protect\citeauthoryear{Elshami et~al.}{2024}]{elshami2024comparative}
\begin{bchapter}
\bauthor{\bsnm{Elshami}, \binits{N.E.}},
\bauthor{\bsnm{Salah}, \binits{A.}},
\bauthor{\bsnm{Mohsen}, \binits{H.}}:
\bctitle{A comparative study of recent {2D} human pose estimation methods}.
In: \bbtitle{2024 6th International Conference on Computing and Informatics
  (ICCI)},
pp. \bfpage{528}--\blpage{537}
(\byear{2024})
\end{bchapter}
\endbibitem

\bibitem[\protect\citeauthoryear{Zheng et~al.}{2023}]{zheng2023deep}
\begin{barticle}
\bauthor{\bsnm{Zheng}, \binits{C.}},
\bauthor{\bsnm{Wu}, \binits{W.}},
\bauthor{\bsnm{Chen}, \binits{C.}},
\bauthor{\bsnm{Yang}, \binits{T.}},
\bauthor{\bsnm{Zhu}, \binits{S.}},
\bauthor{\bsnm{Shen}, \binits{J.}},
\bauthor{\bsnm{Kehtarnavaz}, \binits{N.}},
\bauthor{\bsnm{Shah}, \binits{M.}}:
\batitle{Deep learning-based human pose estimation: A survey}.
\bjtitle{ACM Computing Surveys}
\bvolume{56}(\bissue{1}),
\bfpage{1}--\blpage{37}
(\byear{2023})
\end{barticle}
\endbibitem

\bibitem[\protect\citeauthoryear{Munea et~al.}{2020}]{munea2020progress}
\begin{barticle}
\bauthor{\bsnm{Munea}, \binits{T.L.}},
\bauthor{\bsnm{Jembre}, \binits{Y.Z.}},
\bauthor{\bsnm{Weldegebriel}, \binits{H.T.}},
\bauthor{\bsnm{Chen}, \binits{L.}},
\bauthor{\bsnm{Huang}, \binits{C.}},
\bauthor{\bsnm{Yang}, \binits{C.}}:
\batitle{The progress of human pose estimation: A survey and taxonomy of models
  applied in {2D} human pose estimation}.
\bjtitle{IEEE Access}
\bvolume{8},
\bfpage{133330}--\blpage{133348}
(\byear{2020})
\end{barticle}
\endbibitem

\bibitem[\protect\citeauthoryear{Koul and Novembre}{2025}]{koul2025accurately}
\begin{barticle}
\bauthor{\bsnm{Koul}, \binits{A.}},
\bauthor{\bsnm{Novembre}, \binits{G.}}:
\batitle{How accurately can we estimate spontaneous body kinematics from video
  recordings? effect of movement amplitude on openpose accuracy}.
\bjtitle{Behavior Research Methods}
\bvolume{57}(\bissue{1}),
\bfpage{1}--\blpage{13}
(\byear{2025})
\end{barticle}
\endbibitem

\bibitem[\protect\citeauthoryear{Boniol et~al.}{2008}]{Boniol2008}
\begin{barticle}
\bauthor{\bsnm{Boniol}, \binits{M.}},
\bauthor{\bsnm{Verriest}, \binits{J.-P.}},
\bauthor{\bsnm{Pedeux}, \binits{R.}},
\bauthor{\bsnm{Doré}, \binits{J.-F.}}:
\batitle{Proportion of skin surface area of children and young adults from 2 to
  18 years old}.
\bjtitle{Journal of Investigative Dermatology}
\bvolume{128}(\bissue{2}),
\bfpage{461}--\blpage{464}
(\byear{2008})
\end{barticle}
\endbibitem

\bibitem[\protect\citeauthoryear{Huelke}{1998}]{Huelke1998}
\begin{bchapter}
\bauthor{\bsnm{Huelke}, \binits{D.F.}}:
\bctitle{An overview of anatomical considerations of infants and children in
  the adult world of automobile safety design}.
In: \bbtitle{Annu. Proc. Assoc. Adv. Automot. Med.},
vol. \bseriesno{42},
pp. \bfpage{93}--\blpage{113}
(\byear{1998})
\end{bchapter}
\endbibitem

\bibitem[\protect\citeauthoryear{Groos et~al.}{2022}]{Groos2022}
\begin{barticle}
\bauthor{\bsnm{Groos}, \binits{D.}},
\bauthor{\bsnm{Adde}, \binits{L.}},
\bauthor{\bsnm{Støen}, \binits{R.}},
\bauthor{\bsnm{Ramampiaro}, \binits{H.}},
\bauthor{\bsnm{Ihlen}, \binits{E.A.F.}}:
\batitle{Towards human-level performance on automatic pose estimation of infant
  spontaneous movements}.
\bjtitle{Computerized Medical Imaging and Graphics}
\bvolume{95},
\bfpage{102012}
(\byear{2022})
\doiurl{10.1016/j.compmedimag.2021.102012}
\end{barticle}
\endbibitem

\bibitem[\protect\citeauthoryear{Sermpon and Gima}{2024}]{Sermpon2024}
\begin{barticle}
\bauthor{\bsnm{Sermpon}, \binits{N.}},
\bauthor{\bsnm{Gima}, \binits{H.}}:
\batitle{Correlation between pose estimation features regarding movements
  towards the midline in early infancy}.
\bjtitle{Plos One}
\bvolume{19}(\bissue{2}),
\bfpage{1}--\blpage{15}
(\byear{2024})
\end{barticle}
\endbibitem

\bibitem[\protect\citeauthoryear{Jahn et~al.}{2025}]{Jahn2025}
\begin{barticle}
\bauthor{\bsnm{Jahn}, \binits{L.}},
\bauthor{\bsnm{Flügge}, \binits{S.}},
\bauthor{\bsnm{Zhang}, \binits{D.}},
\bauthor{\bsnm{Poustka}, \binits{L.}},
\bauthor{\bsnm{Bölte}, \binits{S.}},
\bauthor{\bsnm{Wörgötter}, \binits{F.}},
\bauthor{\bsnm{Marschik}, \binits{P.B.}},
\bauthor{\bsnm{Kulvicius}, \binits{T.}}:
\batitle{Comparison of marker-less 2d image-based methods for infant pose
  estimation}.
\bjtitle{Scientific Reports}
\bvolume{15},
\bfpage{12148}
(\byear{2025})
\end{barticle}
\endbibitem

\bibitem[\protect\citeauthoryear{Insafutdinov
  et~al.}{2016}]{Insafutdino2016_DeeperCut}
\begin{botherref}
\oauthor{\bsnm{Insafutdinov}, \binits{E.}},
\oauthor{\bsnm{Pishchulin}, \binits{L.}},
\oauthor{\bsnm{Andres}, \binits{B.}},
\oauthor{\bsnm{Andriluka}, \binits{M.}},
\oauthor{\bsnm{Schiele}, \binits{B.}}:
Deepercut: {A} deeper, stronger, and faster multi-person pose estimation model.
CoRR
\textbf{abs/1605.03170}
(2016)
\end{botherref}
\endbibitem

\bibitem[\protect\citeauthoryear{Wu et~al.}{2019}]{Wu2019_detectron2}
\begin{botherref}
\oauthor{\bsnm{Wu}, \binits{Y.}},
\oauthor{\bsnm{Kirillov}, \binits{A.}},
\oauthor{\bsnm{Massa}, \binits{F.}},
\oauthor{\bsnm{Lo}, \binits{W.-Y.}},
\oauthor{\bsnm{Girshick}, \binits{R.}}:
Detectron2.
\url{https://github.com/facebookresearch/detectron2}
(2019)
\end{botherref}
\endbibitem

\bibitem[\protect\citeauthoryear{Lugaresi
  et~al.}{2019}]{Lugaresi2019_mediapipe}
\begin{botherref}
\oauthor{\bsnm{Lugaresi}, \binits{C.}},
\oauthor{\bsnm{Tang}, \binits{J.}},
\oauthor{\bsnm{Nash}, \binits{H.}},
\oauthor{\bsnm{McClanahan}, \binits{C.}},
\oauthor{\bsnm{Uboweja}, \binits{E.}},
\oauthor{\bsnm{Hays}, \binits{M.}},
\oauthor{\bsnm{Zhang}, \binits{F.}},
\oauthor{\bsnm{Chang}, \binits{C.}},
\oauthor{\bsnm{Yong}, \binits{M.G.}},
\oauthor{\bsnm{Lee}, \binits{J.}},
\oauthor{\bsnm{Chang}, \binits{W.}},
\oauthor{\bsnm{Hua}, \binits{W.}},
\oauthor{\bsnm{Georg}, \binits{M.}},
\oauthor{\bsnm{Grundmann}, \binits{M.}}:
Mediapipe: {A} framework for building perception pipelines.
CoRR
(2019)
\end{botherref}
\endbibitem

\bibitem[\protect\citeauthoryear{Bazarevsky
  et~al.}{2020}]{bazarevsky2020_blazepose}
\begin{botherref}
\oauthor{\bsnm{Bazarevsky}, \binits{V.}},
\oauthor{\bsnm{Grishchenko}, \binits{I.}},
\oauthor{\bsnm{Raveendran}, \binits{K.}},
\oauthor{\bsnm{Zhu}, \binits{T.}},
\oauthor{\bsnm{Zhang}, \binits{F.}},
\oauthor{\bsnm{Grundmann}, \binits{M.}}:
{BlazePose}: On-device real-time body pose tracking.
arXiv preprint arXiv:1907.10226
(2020)
\end{botherref}
\endbibitem

\bibitem[\protect\citeauthoryear{Xiao et~al.}{2018}]{xiao2018_hrnet}
\begin{bchapter}
\bauthor{\bsnm{Xiao}, \binits{B.}},
\bauthor{\bsnm{Wu}, \binits{H.}},
\bauthor{\bsnm{Wei}, \binits{Y.}}:
\bctitle{Simple baselines for human pose estimation and tracking}.
In: \bbtitle{European Conference on Computer Vision (ECCV)}
(\byear{2018})
\end{bchapter}
\endbibitem

\bibitem[\protect\citeauthoryear{Sun et~al.}{2019}]{sun2019deep_hrnet}
\begin{bchapter}
\bauthor{\bsnm{Sun}, \binits{K.}},
\bauthor{\bsnm{Xiao}, \binits{B.}},
\bauthor{\bsnm{Liu}, \binits{D.}},
\bauthor{\bsnm{Wang}, \binits{J.}}:
\bctitle{Deep high-resolution representation learning for human pose
  estimation}.
In: \bbtitle{CVPR}
(\byear{2019})
\end{bchapter}
\endbibitem

\bibitem[\protect\citeauthoryear{Xu et~al.}{2022}]{xu2022_vitpose}
\begin{bchapter}
\bauthor{\bsnm{Xu}, \binits{Y.}},
\bauthor{\bsnm{Zhang}, \binits{J.}},
\bauthor{\bsnm{Zhang}, \binits{Q.}},
\bauthor{\bsnm{Tao}, \binits{D.}}:
\bctitle{Vi{TP}ose: Simple vision transformer baselines for human pose
  estimation}.
In: \bbtitle{Advances in Neural Information Processing Systems}
(\byear{2022})
\end{bchapter}
\endbibitem

\bibitem[\protect\citeauthoryear{Lin et~al.}{2014}]{Lin_2014_COCO}
\begin{botherref}
\oauthor{\bsnm{Lin}, \binits{T.}},
\oauthor{\bsnm{Maire}, \binits{M.}},
\oauthor{\bsnm{Belongie}, \binits{S.J.}},
\oauthor{\bsnm{Bourdev}, \binits{L.D.}},
\oauthor{\bsnm{Girshick}, \binits{R.B.}},
\oauthor{\bsnm{Hays}, \binits{J.}},
\oauthor{\bsnm{Perona}, \binits{P.}},
\oauthor{\bsnm{Ramanan}, \binits{D.}},
\oauthor{\bsnm{Doll{\'{a}}r}, \binits{P.}},
\oauthor{\bsnm{Zitnick}, \binits{C.L.}}:
Microsoft {COCO:} common objects in context.
CoRR
\textbf{abs/1405.0312}
(2014)
\end{botherref}
\endbibitem

\bibitem[\protect\citeauthoryear{Mathis et~al.}{2018}]{Mathis2018_DeepLabCut}
\begin{botherref}
\oauthor{\bsnm{Mathis}, \binits{A.}},
\oauthor{\bsnm{Mamidanna}, \binits{P.}},
\oauthor{\bsnm{Cury}, \binits{K.M.}},
\oauthor{\bsnm{Abe}, \binits{T.}},
\oauthor{\bsnm{Murthy}, \binits{V.N.}},
\oauthor{\bsnm{Mathis}, \binits{M.W.}},
\oauthor{\bsnm{Bethge}, \binits{M.}}:
Deeplabcut: markerless pose estimation of user-defined body parts with deep
  learning.
Nature Neuroscience
(2018)
\end{botherref}
\endbibitem

\bibitem[\protect\citeauthoryear{Andriluka et~al.}{2014}]{andriluka2014_mpii}
\begin{bchapter}
\bauthor{\bsnm{Andriluka}, \binits{M.}},
\bauthor{\bsnm{Pishchulin}, \binits{L.}},
\bauthor{\bsnm{Gehler}, \binits{P.}},
\bauthor{\bsnm{Schiele}, \binits{B.}}:
\bctitle{{2D} human pose estimation: New benchmark and state of the art
  analysis}.
In: \bbtitle{IEEE Conference on Computer Vision and Pattern Recognition (CVPR)}
(\byear{2014})
\end{bchapter}
\endbibitem

\bibitem[\protect\citeauthoryear{Contributors}{2020}]{mmpose2020}
\begin{botherref}
\oauthor{\bsnm{Contributors}, \binits{M.}}:
OpenMMLab Pose Estimation Toolbox and Benchmark.
\url{https://github.com/open-mmlab/mmpose}
(2020)
\end{botherref}
\endbibitem

\bibitem[\protect\citeauthoryear{Hesse et~al.}{2018}]{Hesse2018_minirgbd}
\begin{bchapter}
\bauthor{\bsnm{Hesse}, \binits{N.}},
\bauthor{\bsnm{Bodensteiner}, \binits{C.}},
\bauthor{\bsnm{Arens}, \binits{M.}},
\bauthor{\bsnm{Hofmann}, \binits{U.G.}},
\bauthor{\bsnm{Weinberger}, \binits{R.}},
\bauthor{\bsnm{Sebastian~Schroeder}, \binits{A.}}:
\bctitle{Computer vision for medical infant motion analysis: State of the art
  and {RGB-D} data set}.
In: \bbtitle{Proceedings of the European Conference on Computer Vision (ECCV)
  Workshops}
(\byear{2018})
\end{bchapter}
\endbibitem

\bibitem[\protect\citeauthoryear{Huang et~al.}{2021}]{Huang2021_syrip}
\begin{bchapter}
\bauthor{\bsnm{Huang}, \binits{X.}},
\bauthor{\bsnm{Fu}, \binits{N.}},
\bauthor{\bsnm{Liu}, \binits{S.}},
\bauthor{\bsnm{Ostadabbas}, \binits{S.}}:
\bctitle{Invariant representation learning for infant pose estimation with
  small data}.
In: \bbtitle{IEEE International Conference on Automatic Face and Gesture
  Recognition (FG), 2021}
(\byear{2021})
\end{bchapter}
\endbibitem

\bibitem[\protect\citeauthoryear{Ruggero~Ronchi and Perona}{2017}]{Ronchi2017}
\begin{bchapter}
\bauthor{\bsnm{Ruggero~Ronchi}, \binits{M.}},
\bauthor{\bsnm{Perona}, \binits{P.}}:
\bctitle{Benchmarking and error diagnosis in multi-instance pose estimation}.
In: \bbtitle{Proceedings of the IEEE International Conference on Computer
  Vision (ICCV)}
(\byear{2017})
\end{bchapter}
\endbibitem

\bibitem[\protect\citeauthoryear{Everingham
  et~al.}{2010}]{Everingham_2012_pascal-voc}
\begin{barticle}
\bauthor{\bsnm{Everingham}, \binits{M.}},
\bauthor{\bsnm{Van~Gool}, \binits{L.}},
\bauthor{\bsnm{Williams}, \binits{C.}},
\bauthor{\bsnm{Winn}, \binits{J.}},
\bauthor{\bsnm{Zisserman}, \binits{A.}}:
\batitle{The pascal visual object classes ({VOC}) challenge}.
\bjtitle{International Journal of Computer Vision}
\bvolume{88},
\bfpage{303}--\blpage{338}
(\byear{2010})
\end{barticle}
\endbibitem

\bibitem[\protect\citeauthoryear{Pavlakos et~al.}{2019}]{Pavlakos2019_SMPL-X}
\begin{bchapter}
\bauthor{\bsnm{Pavlakos}, \binits{G.}},
\bauthor{\bsnm{Choutas}, \binits{V.}},
\bauthor{\bsnm{Ghorbani}, \binits{N.}},
\bauthor{\bsnm{Bolkart}, \binits{T.}},
\bauthor{\bsnm{Osman}, \binits{A.A.A.}},
\bauthor{\bsnm{Tzionas}, \binits{D.}},
\bauthor{\bsnm{Black}, \binits{M.J.}}:
\bctitle{Expressive body capture: {3D} hands, face, and body from a single
  image}.
In: \bbtitle{Proceedings IEEE Conf. on Computer Vision and Pattern Recognition
  (CVPR)}
(\byear{2019})
\end{bchapter}
\endbibitem

\bibitem[\protect\citeauthoryear{Virtanen et~al.}{2020}]{Virtanen2020_Scipy}
\begin{barticle}
\bauthor{\bsnm{Virtanen}, \binits{P.}},
\bauthor{\bsnm{Gommers}, \binits{R.}},
\bauthor{\bsnm{Oliphant}, \binits{T.E.}},
\bauthor{\bsnm{Haberland}, \binits{M.}},
\bauthor{\bsnm{Reddy}, \binits{T.}},
\bauthor{\bsnm{Cournapeau}, \binits{D.}},
\bauthor{\bsnm{Burovski}, \binits{E.}},
\bauthor{\bsnm{Peterson}, \binits{P.}},
\bauthor{\bsnm{Weckesser}, \binits{W.}},
\bauthor{\bsnm{Bright}, \binits{J.}},
\bauthor{\bsnm{{van der Walt}}, \binits{t.J.}},
\bauthor{\bsnm{Brett}, \binits{M.}},
\bauthor{\bsnm{Wilson}, \binits{J.}},
\bauthor{\bsnm{Millman}, \binits{K.J.}},
\bauthor{\bsnm{Mayorov}, \binits{N.}},
\bauthor{\bsnm{Nelson}, \binits{A.R.J.}},
\bauthor{\bsnm{Jones}, \binits{E.}},
\bauthor{\bsnm{Kern}, \binits{R.}},
\bauthor{\bsnm{Larson}, \binits{E.}},
\bauthor{\bsnm{Carey}, \binits{C.J.}},
\bauthor{\bsnm{Polat}, \binits{{\. I}.}},
\bauthor{\bsnm{Feng}, \binits{Y.}},
\bauthor{\bsnm{Moore}, \binits{E.W.}},
\bauthor{\bsnm{{VanderPlas}}, \binits{J.}},
\bauthor{\bsnm{Laxalde}, \binits{D.}},
\bauthor{\bsnm{Perktold}, \binits{J.}},
\bauthor{\bsnm{Cimrman}, \binits{R.}},
\bauthor{\bsnm{Henriksen}, \binits{I.}},
\bauthor{\bsnm{Quintero}, \binits{E.A.}},
\bauthor{\bsnm{Harris}, \binits{C.R.}},
\bauthor{\bsnm{Archibald}, \binits{A.M.}},
\bauthor{\bsnm{Ribeiro}, \binits{A.H.}},
\bauthor{\bsnm{Pedregosa}, \binits{F.}},
\bauthor{\bsnm{{van Mulbregt}}, \binits{P.}},
\bauthor{\bsnm{{SciPy 1.0 Contributors}}}:
\batitle{{{SciPy} 1.0: Fundamental Algorithms for Scientific Computing in
  Python}}.
\bjtitle{Nature Methods}
\bvolume{17},
\bfpage{261}--\blpage{272}
(\byear{2020})
\doiurl{10.1038/s41592-019-0686-2}
\end{barticle}
\endbibitem

\bibitem[\protect\citeauthoryear{Diaz-Rojas and Myowa}{2024}]{Diaz-Rojas2024}
\begin{barticle}
\bauthor{\bsnm{Diaz-Rojas}, \binits{F.}},
\bauthor{\bsnm{Myowa}, \binits{M.}}:
\batitle{Estimation of human body 3d pose for parent-infant interaction
  settings using azure kinect and openpose}.
\bjtitle{MethodsX}
\bvolume{13},
\bfpage{102861}
(\byear{2024})
\doiurl{10.1016/j.mex.2024.102861}
\end{barticle}
\endbibitem

\bibitem[\protect\citeauthoryear{Purkrabek and Matas}{2024}]{Purkrabek2024}
\begin{botherref}
\oauthor{\bsnm{Purkrabek}, \binits{M.}},
\oauthor{\bsnm{Matas}, \binits{J.}}:
Detection, pose estimation and segmentation for multiple bodies: Closing the
  virtuous circle.
arXiv preprint 2412.01562
(2024)
{\href{https://arxiv.org/abs/2412.01562}{{2412.01562}}}
\end{botherref}
\endbibitem

\bibitem[\protect\citeauthoryear{Hesse et~al.}{2018}]{Hesse2018_SMIL}
\begin{bchapter}
\bauthor{\bsnm{Hesse}, \binits{N.}},
\bauthor{\bsnm{Pujades}, \binits{S.}},
\bauthor{\bsnm{Romero}, \binits{J.}},
\bauthor{\bsnm{Black}, \binits{M.J.}},
\bauthor{\bsnm{Bodensteiner}, \binits{C.}},
\bauthor{\bsnm{Arens}, \binits{M.}},
\bauthor{\bsnm{Hofmann}, \binits{U.G.}},
\bauthor{\bsnm{Tacke}, \binits{U.}},
\bauthor{\bsnm{Hadders-Algra}, \binits{M.}},
\bauthor{\bsnm{Weinberger}, \binits{R.}},
\bauthor{\bsnm{M{\"u}ller-Felber}, \binits{W.}},
\bauthor{\bsnm{Sebastian~Schroeder}, \binits{A.}}:
\bctitle{Learning an infant body model from {RGB-D} data for accurate full body
  motion analysis}.
In: \bbtitle{Medical Image Computing and Computer Assisted Intervention --
  MICCAI 2018},
pp. \bfpage{792}--\blpage{800}
(\byear{2018})
\end{bchapter}
\endbibitem

\bibitem[\protect\citeauthoryear{Goel et~al.}{2023}]{goel2023_humans4D}
\begin{bchapter}
\bauthor{\bsnm{Goel}, \binits{S.}},
\bauthor{\bsnm{Pavlakos}, \binits{G.}},
\bauthor{\bsnm{Rajasegaran}, \binits{J.}},
\bauthor{\bsnm{Kanazawa}, \binits{A.}},
\bauthor{\bsnm{Malik}, \binits{J.}}:
\bctitle{Humans in 4{D}: Reconstructing and tracking humans with transformers}.
In: \bbtitle{International Conference on Computer Vision (ICCV)}
(\byear{2023})
\end{bchapter}
\endbibitem

\bibitem[\protect\citeauthoryear{Seethapathi
  et~al.}{2019}]{seethapathi2019movement}
\begin{botherref}
\oauthor{\bsnm{Seethapathi}, \binits{N.}},
\oauthor{\bsnm{Wang}, \binits{S.}},
\oauthor{\bsnm{Saluja}, \binits{R.}},
\oauthor{\bsnm{Blohm}, \binits{G.}},
\oauthor{\bsnm{Kording}, \binits{K.P.}}:
Movement science needs different pose tracking algorithms.
arXiv preprint arXiv:1907.10226
(2019)
\end{botherref}
\endbibitem

\end{thebibliography}

\flushbottom
\newpage

\thispagestyle{empty}
\part*{Automatic infant 2D pose estimation from videos: comparing seven deep neural network methods (Supplementary Materials)}
\setcounter{table}{0}
\setcounter{figure}{0}
\setcounter{equation}{0}

\section*{Materials and Methods}
\label{sec:methods}

\subsection*{2D Pose estimation techniques}

To make the comparisons between the methods fair, we used the versions of the models trained on the COCO dataset, but some of the methods have models trained on other datasets, which sometimes provide a different amount of keypoints and different training dataset sizes and examples, such as BODY 25 (25 keypoints), MediaPipe/BlazePose (33 keypoints), COCO-WholeBody (133 keypoints), or Halpe (136 keypoints). These other variants may or may not perform better than the ones we used and might be more suited to study some tasks (e.g., finger estimation would be needed to study grasping, and more facial keypoints could help to study head orientation).

The details of the parameters and weights version used are described below, for each method.

\subsubsection*{AlphaPose}
Alphapose is publicly available at \url{(https://github.com/MVIG-SJTU/AlphaPose)}, was run using the \textit{Fast Pose (DUC) - ResNet50 unshuffled} version with following parameters:
\begin{verbatim}
    --cfg 256x192_res50_lr1e-3_1x-duc.yaml \
    --checkpoint fast_421_res50-shuffle_256x192.pth \
    --detbatch 2 --posebatch 40
\end{verbatim}

\subsubsection*{DeepLabCut}
DeepLabCut, publicly available at \url{https://github.com/DeepLabCut/DeepLabCut}, was run on version 2.2.1.1 with the \textit{full\_human} pre-trained model from DLC's Model Zoo trained on the MPII dataset, with the following parameters.
\begin{verbatim}
    shuffle=1, trainingsetindex=0
\end{verbatim}

\subsubsection*{Detectron2}
Detectron2 is publicly available at \url{https://github.com/facebookresearch/detectron2}. Detectron2 is a library with models trained to solve several computer vision tasks, some of them being keypoint detection and pose estimation. Those have been trained on the COCO dataset. We used the model \textit{R50-FPN} with model ID: \textit{137849621} (see Detectron2/ModelZoo on github) on version 0.6. It has been run using the following parameters:
\begin{verbatim}
    --config-file keypoint_rcnn_R_50_FPN_3x.yaml \
    --opts MODEL.WEIGHTS model_final_a6e10b.pkl
\end{verbatim}

\subsubsection*{MediaPipe}
MediaPipe is publicly available at \url{https://github.com/google-ai-edge/mediapipe}. It is a library with models trained to solve several tasks, including human pose estimation with \textit{BlazePose}. It has been trained with BlazeFace and BlazePalm on top of COCO, for extra face and hand keypoints, which we did not use. We used its Python library on version 0.10.14 with the \textit{heavy} model, with default parameters and tracking enabled for both input types:
\begin{verbatim}
    model_asset_path="pose_landmarker_heavy.task"
    num_poses=1
    running_mode=vision_running_mode.VIDEO
\end{verbatim}

\subsubsection*{HRNet}
We used the HRNet implementation of the \textit{MMPose} environment on version 0.28.0, which is publicly available at \url{https://github.com/open-mmlab/mmpose}.
This environment contains many pre-trained models, trained on different datasets including COCO, MPII, COCO-WholeBody and Halpe. Some, including HRNet, have been implemented using the two general approaches to pose estimation methods, \textit{Bottom-Up} and \textit{Top-Down}.
The Bottom-Up version has been run using the following parameters:
\begin{verbatim}
    HRNet_w32_coco_512x512-bcb8c247_20200816.pth
\end{verbatim}
The Top-Down version has been run using the following parameters, with the default detector:
\begin{verbatim}
    faster_rcnn_r50_caffe_fpn_mstrain_1x_coco-5324cff8.pth \
    HRNet_w48_coco_384x288_dark-e881a4b6_20210203.pth
\end{verbatim}

\subsubsection*{OpenPose}
OpenPose, publicly available at \url{https://github.com/CMU-Perceptual-Computing-Lab/openpose}, has been run with the following parameters:
\begin{verbatim}
    --display 0 --render_pose 0 --model_pose COCO \
    --number_people_max 1 --net resolution "512x400" \
    --scale_number 2 --scale_gap 0.25
\end{verbatim}

\subsubsection*{ViTPose}
We used the implementation of ViTPose available through the \textit{MMPose} environment on version 1.1.0, which is publicly available at \url{https://github.com/open-mmlab/mmpose} .
There are several pre-trained models, trained on different datasets, we used VitPose-H (Huge) trained on COCO, AIC and MPII, with the default detector. It has been run with the following parameters:
\begin{verbatim}
    faster_rcnn_r50_fpn_1x_coco_20200130-047c8118.pth \
    td-hm_ViTPose-huge_8xb64-210e_coco-256x192-e32adcd4_20230314.pth
\end{verbatim}

\subsubsection*{``Mixture of experts'' estimates}
In some cases, averaging estimates from different sources can lead to higher accuracy than any single source. We took the three methods that perform the best overall, ViTPose, HRNet TD, and HRNet BU, and averaged their highest-confidence estimates into a new ``mixture of experts'' method. 

However, note that this way of averaging renders some of the evaluation metrics difficult to interpret. Averaging confidences between different methods is difficult considering that all methods have their own, partially arbitrary and proprietary, way of using the confidences (see Sup. Mat. Figs.~\ref{fig:scatter_sco_oks_suprea}-\ref{fig:scatter_sco_oks_synth}.); as such, OKS, AP and AR, as well as correlations between scores and OKS cannot be computed for this ``mixture of experts''. In addition, redundant detections are intrinsic to each method, and selection of the best detection for each method has to be done before the averaging is computed. Missing data would only happen if none of the three methods, ViTPose, HRNet TD and HRNet BU provide an estimate. Processing time would be either the sum of each individual method, if processed serially, or have a lower boundary equal to the slowest method as it would bottleneck the process if the methods are run in parallel. 
As such, we only directly evaluated these estimates for the Neck-MidHip errors.

\subsection*{Datasets}
In addition to the annotated ``Supine'' dataset mentioned in the main manuscript, additional images were annotated.

``Extended supine'' dataset: we manually annotated 720 additional images from eight additional videos of the same two infants. The images were selected in a similar manner as described in the main manuscript, but were processed only with video input by the methods, hence why they are not included in the main manuscript where the results between videos and images input are compared. Thus, we reach a total of 1440 annotated images of infants in supine position from 16 videos (191 862 images) that were processed as video inputs by all methods.

``Sequential supine'' dataset: we additionally annotated 900 consecutive images from a single video, identified as "AA\_17w" in several tables in the Supplementary Materials. This sequence was selected because of the presence of typical hand movements and self-touch occurrences that we are particularly interested in processing in the future to study the development of infant body know-how. In retrospective, we observed that this video is the one where all the methods performed the best (except for DeepLabCut and MediaPipe, see Tab. ST.~\ref{tab:real_oks}), most likely because the camera is properly positioned and angled above the infant (see Figure~1., b) in the main manuscript), and the infant does not perform many complex leg movements. As such, it could be considered to have optimal conditions and show the upper bound performance of the methods on infants in supine position. Due to how this sequence was arbitrarily chosen and because it only concerns a single infant at a single age, we decided to not include it in the main manuscript, and leave it as an extra in the Supplementary Materials.

\begin{table}[!ht]%
    \addtolength{\tabcolsep}{-0.3em}
    \centering
    \begin{tabular}{c|c|c|c|c|c|c|c|c|}

    Features / Dataset & Supine & Extended Supine & Sequential Supine & Lap & SyRIP & MINI-RGBD \\
    \hline
    No. labeled images & 720 & 1440 & 900 & 359 & 500 & 12 000 \\
    Typical posture & Supine & Supine & Supine & Sitting & Mixed & Supine \\
    No. keypoints & 14 & 14 & 14 & 17 & 17 & 13 \\
    Real or Synth. & Real & Real & Real & Real & Real & Synthetic \\
    Images with adult & 0 & 0 & 0 & 359 & 1 & 0 \\
    \hline
    \end{tabular}
    \caption{Summary of the characteristics of each dataset used for evaluating the pose estimation methods, including extended and supplementary datasets not in the main manuscript.}
    \label{tab:datasets_summary}
\end{table}

\section*{Results}
\label{sec:results}

\subsection*{Settings for detection selection with redundant detections}
\label{subsubsec:methods_detselset}
In most applications, when there are redundant detections, a selection process is necessary to choose one of them for further processing. This can only be done using information provided by pose estimation methods along with keypoints, such as the rank order in which the detections are output or their scores and confidences. This is not a simple task, especially in multi-person scenes. We computed Euclidean distances between estimates and their ground truth, the first processing step at the base of the OKS and Neck-MidHip computations, under each of these three detection selection settings:
\begin{itemize}
 \item \textit{Det 1}: using the first ranked detection provided by the method. Redundant detections proposed by the method are ignored.
 \item \textit{Det 2}: the average Euclidean distance is computed for each detection of a given image. Then, the detection with the shortest distance is selected. This corresponds to the best detection that the method can offer, though it is not available without ground truth. 
 \item \textit{Det 3}: using the detection with the highest score, ignoring the rank order. For HRNet TD that did not provide scores, we used the score of the bounding-box provided by its detector as a substitute. For ViTPose, we used the median of its confidences.
\end{itemize}
These settings are used to 1) verify the general assumption that the detection with the highest score is always ranked first, and if not, 2) which of the two available selection metrics (rank or score) matches the best detection proposed by the methods and is a better choice.

Some minimal differences were observed when comparing the results for all relevant metrics between the three detection selection settings, namely, the first-rank detection, the optimal detection, and the highest-scored detection on our ``Supine'' dataset, as well as on the MINI-RGBD and the SyRIP datasets. The most extreme case was HRNet TD with a drop of 2 percentage points on the average Neck-MidHip errors for the SyRIP dataset when using the optimal detection, representing a relative difference of 16\% of the average error when using the first ranked detection (while ViTPose's errors dropped by 1.1 points, or 9\% of the first ranked detection error, keeping it as the method with the lowest average error).
DeepLabCut, MediaPipe, and OpenPose yielded identical results, as they only provided one detection in these cases. 
AlphaPose and ViTPose showed a few cases where the first-ranked detection did not have the highest score, though for ViTPose it might be due to how we estimated its score as the median of the confidences: a different internal detection ranking process might have led to different or no changes.
Both first-ranked and highest-score detections had differences with the optimal detection, meaning that there is no guaranteed way to select the best available detection for any of the methods when no ground-truth is available.
Generally, using the highest-scored detection resulted in the closest performance to the optimal detection for all methods. Hence, the whole manuscript focuses on showing only the results of the highest-scored detections.

\begin{figure}[!ht]
    \centering
    \includegraphics[scale=1, width=1\linewidth]{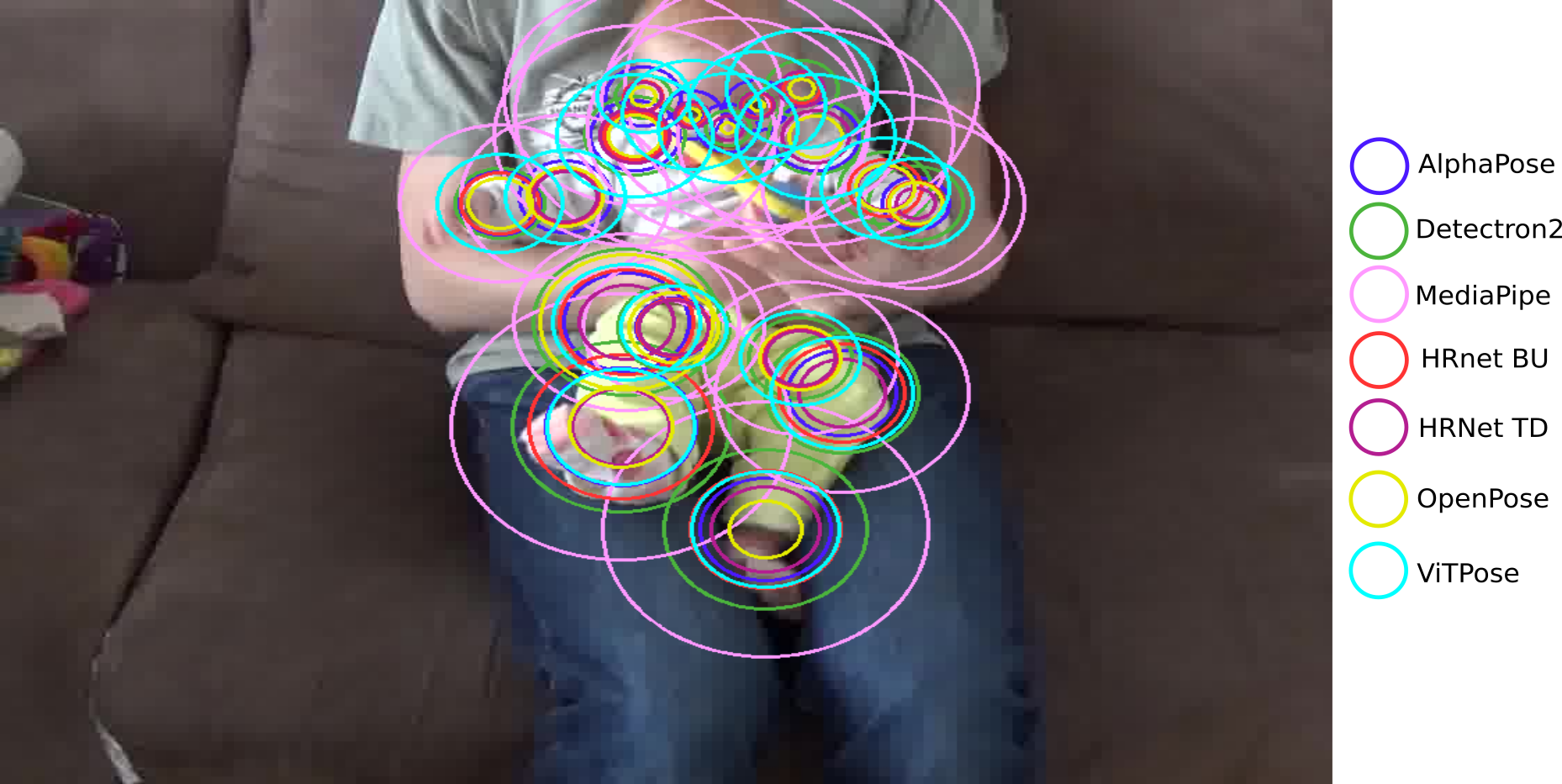}

    \caption{Average Neck-MidHip errors for each keypoint with available ground truth on our ``Lap'' dataset when using best matching detections. The centre of the circles is the ground-truth position for that keypoint. The radius of each circle shows the average amplitude of the errors, scaled to the Neck-MidHip segment. The colors separately represent each pose estimation method.}
    \label{fig:n-mh_circles_sup_lap}
\end{figure}

However, for the ``Lap'' dataset, where both the infant and a parent are present in the image, the differences are more drastic (see Table ST~\ref{tab:lap_matchdet} compared to the results in the main manuscript Tables 2, 4 and 7), most likely due to the bias of pose estimation methods towards adults: it would be expected that they would tend to have higher score or confidence values towards adult detections than infants, and we observed cases where only the adult was detected, which can be seen in particular from the large difference in the OKS and Neck-MidHip results (an example is given in Fig. 1, (f) in the main manuscript). Surprisingly, however, ViTPose is far from having the best results, being the second last method. This is most likely because it misses detecting the infant for a large potion of the images, similar to MediaPipe, as can be seen in Table 6 of the main manuscript, while the methods that had many redundant detections had an advantage due to being more likely to include at least one detection of the infant, improving their results in this best-matching selection scenario.
We can observe per-keypoint errors to be lower as well for all methods in this scenario, being closer to the errors we observed for the SyRIP dataset (see Fig. SF.~\ref{fig:n-mh_circles_sup_lap}, Tab. ST.~\ref{tab:n-mh_lap_kps} and~\ref{tab:n-mh_syrip_kps}).
In a multi-person scene, it would be possible to find alternatives for more optimal detection matching rather than use the first or highest ranked detections: for example, it would be possible to use a ``tracking" algorithm as a postprocess to separate each individual throughout the video using temporal and spatial coherence, or to use some naive method such as using the detection which has the lowest area, or the one with the smallest difference between its extreme X and Y axis values; though both of these naive solution have edge cases where they would fail, which can depend on the recording settings, in particular the camera angle and the position of the infant and parent with regard to this camera, or could fail whenever someone else (e.g. the experimenter) gets even partially (one arm) in view of the camera.

\begin{table}[!ht]%
    \addtolength{\tabcolsep}{-0.3em}
    \centering
    \begin{tabular}{l|c|c|c|c|c|c|c|c|}
    Metrics & AlphaPose & DeepLabCut & Detectron 2 & MediaPipe & HRNet BU & HRNet TD & OpenPose & ViTPose \\
    \hline
    OKS & 0.71$\pm$0.22 & N/A & 0.67$\pm$0.19 & 0.37$\pm$0.28 & 0.74$\pm$0.17 & \textbf{0.80$\pm$0.21} & 0.74$\pm$0.18 & 0.71$\pm$0.32 \\
    \hdashline
    N-MH & 16.8$\pm$4.9 & N/A & 20.4$\pm$10.2 & 50.7$\pm$10.4 & 14.8$\pm$8.4 & \textbf{12.1$\pm$4.5} & 12.6$\pm$8.2 & 25.5$\pm$3.5 \\
    \hdashline
    CPE & 0.82 & N/A & 0.77 & 0.44 & 0.85 & \textbf{0.88} & 0.37 & 0.75 \\
    \hdashline
    Sco.-OKS corr. & 0.23 & N/A & 0.17 & N/A & \textbf{0.78} & 0.49 & 0.57 & 0.37 \\
    \hline
    \end{tabular}
    \caption{Summary of the best-matching detection selection results for the distance and correlation metrics on our ``Lap'' dataset. For score-OKS Spearman Rank Coefficient Correlations, all p-values $\protect<$ 0.005. N/A values correspond to methods not evaluated on image inputs as they only accepted video input (see Sec. ``Comparison metrics'' first paragraph and Sec. ``Hardware and methods' code modifications'' in the main manuscript for more details).}
    \label{tab:lap_matchdet}
\end{table}

\subsection*{Results summary for the ``mixture of experts'' estimates}

The Neck-MidHip errors for the ``mixture of experts'' estimates---averaging the estimates from ViTPose, HRNet TD and HRNet BU---are shown in Sup. Tab.~\ref{tab:n-mh_hybrid}. Overall, it seems that the averaged estimates are worse than the best individual method, with the exception of the SyRIP dataset, where the average error is very slightly better than ViTPose's estimation alone (by 0.1\% of the Neck-MidHip length), but with a slightly higher standard deviation (0.7\% of the Neck-MidHIp length). In some cases (``Lap'' and Mini-RGBD datasets), the errors are larger for the ``mixture of experts'' estimates, but the standard deviation is slightly lower. For the ``Supine'' dataset, the ``mixture'' estimates have both higher average errors and higher standard deviation than ViTPose.

\begin{table}[!ht]%
    \centering
    \begin{tabular}{c|c|ccc|c|}
    Dataset & Input & HRNet BU & HRNet TD & ViTPose & Mixture\\
    \hline
    \textbf{Real} 
    Supine & Videos & $7.9\pm3.9$ & $6.7\pm3.6$ & \textbf{6.0}$\pm$\textbf{2.7} & 6.2$\pm$2.8\\ 
    \hdashline
    Lap & Images & \textbf{24.3$\pm$9.0} & 42.9$\pm$9.3 & 40.0$\pm$8.9 & 33.9$\pm$6.7\\
    \hdashline
    SyRIP & Images & 15.8$\pm$7.2 & 13.5$\pm$5.8 & 12.0$\pm$4.3 & \textbf{11.9$\pm$5.0}\\
    \hline
    \hline
    \textbf{Synth.}
    MINI-RGBD & Videos & 13.5$\pm$4.5 & 9.1$\pm$4.7 & \textbf{8.2}$\pm$\textbf{4.7} & 10.3$\pm$4.1\\
    \hline
    \end{tabular}
    \caption{Average errors across all keypoints as a percentage of the Neck-MidHip segment, with standard deviation, for the three best performing methods (ViTPose, HRNet TD, HRNet BU) and for the ``mixture of experts'' that averages the three methods' estimates.}
    \label{tab:n-mh_hybrid}
\end{table}

\subsection*{Results summary for the ``extended supine'' dataset processed by video input}
The summary of the results including all 1440 annotated images from the 16 recordings in supine position is described in the Tab. ST~\ref{tab:all_real}.

We observe similar results as in the main manuscript, with a tendency for slightly higher AP, AR, and OKS, and lower missing data, but also lower correlations between score and OKS values.

\begin{table}[!ht]%
    \addtolength{\tabcolsep}{-0.4em}
    \centering
    \begin{tabular}{l|c|c|c|c|c|c|c|c|}
    Metrics & AlphaPose & DeepLabCut & Detectron 2 & MediaPipe & HRNet BU & HRNet TD & OpenPose & ViTPose \\
    \hline
    OKS & 0.88$\pm$0.07 & 0.11$\pm$0.12 & 0.87$\pm$0.09 & 0.32$\pm$0.20 & 0.90$\pm$0.06 & 0.92$\pm$0.05 & 0.81$\pm$0.17 & \textbf{0.92$\pm$0.04} \\
    \hdashline
    AP & 66.5 & 0.8 & 25.3 & 0.6 & 77.8 & 59.1 & 59.9 & \textbf{86.7} \\
    AR & 76.9 & 0.3 & 79.1 & 4.0 & 85.9 & 88.5 & 66.8 & \textbf{90.0} \\
    \hdashline
    N-MH & $7.6\pm2.8$ & $103.6\pm20.5$ & $9.4\pm4.3$ & 44.6$\pm$15.5 & $7.1\pm3.4$ & $6.6\pm3.5$ & $10.7\pm3.4$ & \textbf{6.0}$\pm$\textbf{2.9} \\
    \hdashline
    Missing data & 3.3\% & \textbf{0\%} & \textbf{0\%} & 5.6\% & \textbf{0\%} & 0.9\% & 6.1\% & 0.1\% \\
    \hdashline
    CPE & 0.89 & 0.5 & 0.91 & 0.50 & 0.93 & 0.93 & 0.83 & \textbf{0.94} \\
    \hdashline
    Redun. det. & 4.6\% & \textbf{0\%} & 259.5\% & \textbf{0\%} & 5.1\% & 54.0\% & \textbf{0\%} & 19.7\% \\
    \hdashline
    Sco.-OKS corr. & 0.26 & 0.08 & 0.21 & N/A & \textbf{0.51} & 0.47 & 0.35 & 0.39 \\
    \hline
    \end{tabular}
    \caption{Summary of the highest-scored detection results for the full set of 1440 real infants annotated images processed by video input. For score-OKS Spearman Rank Coefficient Correlations, all p-values $\protect<$ 0.005.}
    \label{tab:all_real}
\end{table}

\newpage

\subsection*{Results summary for the ``sequential supine'' dataset}

The results summary for the 900 continuous sequence from a single recording session, AA\_17w, are shown in Supplementary Tab. ST.~\ref{tab:summary_seq}.

The AP, AR, and OKS values are higher than the averages found in the main manuscript or in the Supplementary Tab. ST~\ref{tab:all_real}, while the average Neck-MidHip errors are lower, except for DeepLabCut and MediaPipe.

Concerning the correlations, we observe that they are particularly low, especially for Detectron2, where they are negative. 

In retrospective when comparing to Supplementary Table ST.~\ref{tab:real_oks}, it seems that the recording from which this sequence was extracted was the one with the highest OKS for most methods. In a way, this would mean that the results here, in Supplementary Tab. ST.~\ref{tab:summary_seq} might represent the upper limit of performance that could be achieved by each method (except MediaPipe) in a best-case scenario.

\begin{table}[!ht]%
    \centering
    \addtolength{\tabcolsep}{-0.6em}
    \footnotesize
    \begin{tabular}{l|c|c|c|c|c|c|c|c|c|}
    Input & Metrics & AlphaPose & DeepLabCut & Detectron 2 & MediaPipe & HRNet BU & HRNet TD & OpenPose & ViTPose \\
    \hline
    Images & OKS & 0.90$\pm$0.04 & N/A & 0.93$\pm$0.04 & 0.40$\pm$0.10 & 0.94$\pm$0.03 & \textbf{0.95$\pm$0.03} & 0.93$\pm$0.05 & N/A \\
    & AP & 77.9 & N/A & 81.4 & 0.1 & \textbf{91.2} & 66.4 & 87.8 & N/A \\
    & AR & 81.7 & N/A & 90.0 & 2.4 & 94.3 & \textbf{95.2} & 90.6 & N/A \\
    &N-MH & 5.0$\pm$1.7\% & N/A & 4.7$\pm$2.4\% & 24.7$\pm$12.0\% & \textbf{4.0$\pm$2.0\%} & \textbf{3.9$\pm$2.2\%} & 4.5$\pm$1.9\% & N/A \\
    & CPE& 0.95 & N/A & 0.95 & 0.75 & \textbf{0.96} & \textbf{0.96} & 0.95 & N/A \\
    &Sco.-OKS cor. & 0.11 & N/A & -0.35 & N/A & 0.04 ($p=0.26$) & \textbf{0.30} & 0.06 ($p=0.06$) & N/A \\
    \hline
    \hline
    Video & OKS & 0.90$\pm$0.04 & 0.08$\pm$0.06 & 0.92$\pm$0.04 & 0.40$\pm$0.10 & 0.94$\pm$0.03 & 0.95$\pm$0.03 & 0.93$\pm$0.05 & \textbf{0.95$\pm$0.02} \\
    & AP & 76.9 & 0.0 & 34.2 & 0.3 & 90.8 & 63.9 & 87.3 & \textbf{91.8} \\
    & AR & 81.1 & 0.0 & 89.9 & 2.5 & 93.8 & \textbf{95.1} & 90.2 & 94.5 \\
    & N-MH & 5.1$\pm$1.8\% & 79.3$\pm$15.3\% & 4.9$\pm$2.5\% & 24.8$\pm$0.12\% & 4.0$\pm$2.0\% & 3.9$\pm$2.2\% & 4.6$\pm$1.9\% & \textbf{3.9$\pm$2.0\%} \\
    & CPE & 0.95 & 0.5 & 0.95 & 0.75 & \textbf{0.96} & \textbf{0.96} & 0.95 & \textbf{0.96} \\
    &Sco.-OKS cor. & 0.10 ($p<0.006$) & 0.11 & -0.32 & N/A & 0.01 ($p=0.70$) & \textbf{0.28} & 0.10 & 0.18 \\
    \end{tabular}
    \caption{Summary of the highest-scored detections results from each metric for the set of 900 consecutive annotated images from video AA\_17w. For score-OKS Spearman Rank Coefficient Correlations, all p-values $\protect<$ 0.005 when not stated.}
    \label{tab:summary_seq}
\end{table}

\subsection*{Object Keypoint Similarity (OKS)}

The average OKS values for individual videos in supine position can be seen in Tables ST~\ref{tab:real_oks} and ST~\ref{tab:synth_oks}. It can be used to identify the most challenging or easiest videos for each method.

As the synthetic infants from the MINI-RGBD dataset come with an estimate of the complexity of their sequence, we can observe that the performance between the easy group (IDs 1-4) and the medium group (5-9) seems close, except for synthetic infant 1 that some methods seem to struggle with, and synthetic infant 9 for which all methods show a drop of performance. Synthetic infants 7 and 8 are handled differently by the methods, some keeping their performance levels, while others display drops of performance. However, for all methods, we can observe a performance drop between the easy and medium video group and the difficult video group (IDs 10-12).

\begin{table}[!ht]%
    \addtolength{\tabcolsep}{-0.4em}
    \centering
    \begin{tabular}{c|c|c|c|c|c|c|c|c|c|}
    \textbf{Real} & Video ID & AlphaPose & DeepLabCut & Detectron 2 & MediaPipe & HRNet BU & HRNet TD & OpenPose & ViTPose\\
    \hline
    Images & AA\_8w & 0.86 & N/A & 0.78 & 0.50 & 0.81 & \textbf{0.89} & 0.79 & N/A \\
    & AA\_17w & 0.90 & N/A & 0.92 & 0.39 & \textbf{0.94} & \textbf{0.94} & 0.92 & N/A \\
          & TH\_8w\_st1 & 0.89 & N/A & 0.88 & 0.43 & 0.90 & \textbf{0.91} & 0.88 & N/A \\
          & TH\_8w\_st2 & 0.90 & N/A & 0.89 & 0.39 & 0.91 & \textbf{0.93} & 0.89 & N/A \\
          & TH\_8w\_st3 & 0.88 & N/A & 0.89 & 0.32 & 0.90 & \textbf{0.91} & 0.89 & N/A \\
          & TH\_15w & 0.77 & N/A & 0.86 & 0.17 & 0.90 & \textbf{0.92} & 0.80 & N/A \\
          & TH\_19w & 0.87 & N/A & 0.86 & 0.19 & 0.90 & \textbf{0.93} & 0.88 & N/A \\
          & TH\_25w & 0.89 & N/A & 0.87 & 0.50 & \textbf{0.93} & \textbf{0.93} & 0.88 & N/A \\
          & \textbf{Mean} & 0.87 & N/A & 0.87 & 0.39 & 0.90 & \textbf{0.92} & 0.87 & N/A \\
    \hline
    \hline
    Videos & AA\_8w & 0.86 & 0.31 & 0.77 & 0.52 & 0.81 & 0.89 & 0.79 & \textbf{0.94}\\
    & AA\_11w & 0.86 & 0.19 & 0.85 & 0.19 & 0.90 & \textbf{0.91} & 0.85 & \textbf{0.91} \\
          & AA\_13w & 0.90 & 0.48 & 0.91 & 0.19 & 0.91 & \textbf{0.94} & 0.81 & \textbf{0.94} \\
          & AA\_17w & 0.90 & 0.09 & 0.92 & 0.37 & 0.94 & 0.94 & 0.92 & \textbf{0.95} \\
          & AA\_19w & 0.89 & 0.02 & 0.90 & 0.15 & \textbf{0.93} & \textbf{0.93} & 0.91 & \textbf{0.93} \\
          & TH\_8w\_st1 & 0.88 & 0.15 & 0.87 & 0.42 & 0.90 & \textbf{0.91} & 0.87 & \textbf{0.91} \\
          & TH\_8w\_st2 & 0.90 & 0.07 & 0.88 & 0.39 & 0.91 & \textbf{0.93} & 0.90 & \textbf{0.93} \\
          & TH\_8w\_st3 & 0.88 & 0.10 & 0.89 & 0.36 & 0.90 & 0.90 & 0.89 & \textbf{0.92} \\
          & TH\_10w\_st1 & 0.90 & 0.15 & 0.89 & 0.47 & 0.91 & \textbf{0.93} & 0.89 & \textbf{0.93} \\
          & TH\_10w\_st2 & 0.84 & 0.08 & 0.87 & 0.33 & \textbf{0.90} & 0.89 & 0.80 & \textbf{0.90} \\
          & TH\_12w & 0.89 & 0.20 & 0.88 & 0.40 & 0.92 & 0.90 & 0.84 & \textbf{0.93} \\
          & TH\_15w & 0.75 & 0.04 & 0.86 & 0.19 & 0.90 & \textbf{0.92} & 0.58 & \textbf{0.92} \\
          & TH\_17w & 0.87 & 0.04 & 0.86 & 0.07 & \textbf{0.90} & 0.88 & 0.83 & 0.89 \\
          & TH\_19w & 0.88 & 0.05 & 0.86 & 0.11 & 0.90 & 0.92 & 0.52 & \textbf{0.93} \\
          & TH\_23w & 0.87 & 0.06 & 0.82 & 0.24 & 0.91 & 0.91 & 0.68 & \textbf{0.93} \\
          & TH\_25w & 0.88 & 0.17 & 0.87 & 0.49 & 0.93 & 0.93 & 0.76 & \textbf{0.94} \\
          & \textbf{Mean} & 0.88 & 0.11 & 0.87 & 0.32 & 0.90 & \textbf{0.92} & 0.81 & \textbf{0.92}\\
    \hline
    \end{tabular}
    \caption{Average OKS values over all manually-annotated images for each method and input type on real infants.}
    \label{tab:real_oks}
\end{table}

\begin{table}[!ht]%
    \addtolength{\tabcolsep}{-0.4em}
    \centering
    \begin{tabular}{c|c|c|c|c|c|c|c|c|c|}
    \textbf{Synth.} & Video ID & AlphaPose & DeepLabCut & Detectron 2 & MediaPipe & HRNet BU & HRNet TD & OpenPose & ViTPose\\
    \hline
    Images & Syn. 1 & 0.83 & N/A & 0.84 & 0.34 & 0.61 & \textbf{0.90} & 0.84 & N/A\\
        & Syn. 2 & 0.87 & N/A & 0.83 & N/A & 0.90 & \textbf{0.91} & 0.86 & N/A\\
        & Syn. 3 & 0.89 & N/A & 0.90 & 0.52 & 0.90 & \textbf{0.91} & 0.89 & N/A\\
        & Syn. 4 & 0.90 & N/A & 0.90 & 0.62 & \textbf{0.90} & \textbf{0.90} & 0.84 & N/A\\
        & Syn. 5 & 0.89 & N/A & 0.87 & 0.59 & \textbf{0.90} & \textbf{0.90} & 0.85 & N/A\\
        & Syn. 6 & 0.86 & N/A & 0.88 & 0.57 & \textbf{0.91} & 0.90 & 0.88 & N/A\\
        & Syn. 7 & 0.89 & N/A & 0.89 & 0.59 & \textbf{0.91} & \textbf{0.91} & 0.88 & N/A\\
        & Syn. 8 & 0.85 & N/A & 0.85 & 0.45 & 0.87 & \textbf{0.89} & 0.84 & N/A\\
        & Syn. 9 & 0.73 & N/A & 0.82 & 0.52 & 0.86 & \textbf{0.90} & 0.72 & N/A\\
        & Syn. 10 & 0.79 & N/A & 0.79 & 0.39 & 0.76 & \textbf{0.81} & 0.75 & N/A\\
        & Syn. 11 & 0.84 & N/A & 0.87 & 0.55 & 0.87 & \textbf{0.88} & 0.86 & N/A\\
        & Syn. 12 & 0.67 & N/A & 0.66 & 0.35 & 0.75 & \textbf{0.78} & 0.75 & N/A\\
        & \textbf{Mean} & 0.84 & N/A & 0.84 & 0.50 & 0.83 & \textbf{0.88} & 0.83 & N/A \\
    \hline
    \hline
    Videos & Syn. 1 & 0.76 & 0.37 & 0.85 & 0.22 & 0.59 & 0.87 & 0.85 & \textbf{0.89}\\
        & Syn. 2 & 0.86 & 0.46 & 0.82 & N/A & 0.89 & \textbf{0.90} & 0.84 & 0.89\\
        & Syn. 3 & 0.88 & 0.49 & 0.88 & 0.51 & 0.88 & \textbf{0.89} & 0.87 & \textbf{0.89} \\
        & Syn. 4 & 0.86 & 0.40 & 0.86 & 0.61 & 0.82 & 0.84 & 0.80 & \textbf{0.88}\\
        & Syn. 5 & 0.88 & 0.59 & 0.85 & 0.57 & 0.87 & \textbf{0.90} & 0.85 & \textbf{0.90}\\
        & Syn. 6 & 0.86 & 0.51 & 0.87 & 0.57 & \textbf{0.90} & \textbf{0.90} & 0.86 & 0.89\\
        & Syn. 7 & 0.87 & 0.55 & 0.87 & 0.58 & \textbf{0.89} & \textbf{0.89} & 0.86 & 0.88\\
        & Syn. 8 & 0.83 & 0.36 & 0.83 & 0.44 & 0.85 & \textbf{0.87} & 0.82 & 0.85\\
        & Syn. 9 & 0.48 & 0.38 & 0.74 & 0.42 & 0.78 & 0.87 & 0.70 & \textbf{0.89}\\
        & Syn. 10 & 0.78 & 0.35 & 0.75 & 0.38 & 0.74 & 0.78 & 0.72 & \textbf{0.83}\\
        & Syn. 11 & 0.83 & 0.50 & 0.85 & 0.54 & \textbf{0.86} & \textbf{0.86} & 0.84 & 0.84\\
        & Syn. 12 & 0.56 & 0.19 & 0.59 & 0.35 & 0.69 & 0.75 & 0.69 & \textbf{0.79}\\
        & \textbf{Mean} & 0.81 & 0.43 & 0.81 & 0.47 & 0.81 & 0.86 & 0.81 & \textbf{0.87} \\
    \hline
    \end{tabular}
    \caption{Average OKS values over all manually-annotated images for each method and input type on synthetic infants.}
    \label{tab:synth_oks}
\end{table}

\newpage

\subsection*{Neck-MidHip error}
\label{subsec:n-mhip_sup}

\begin{figure}[!ht]
    \centering
    \includegraphics[scale=1, width=1\linewidth]{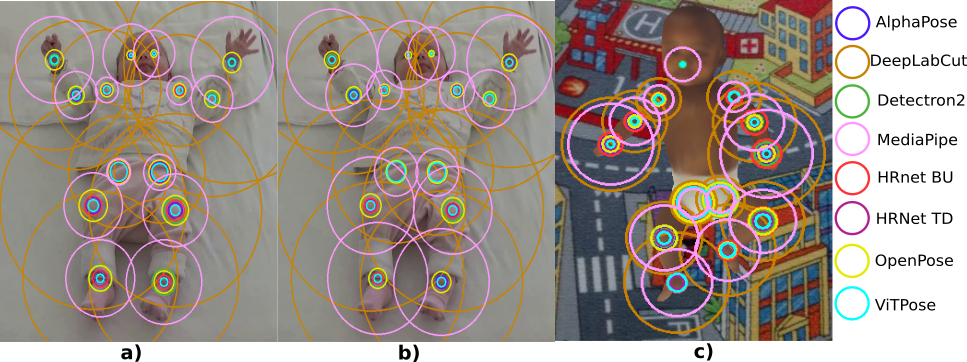}

    \caption{Average Neck-MidHip errors for each keypoint with available ground truth on our ``Supine'' and the MINI-RGBD datasets. The centre of the circles is the ground-truth position for that keypoint. The radius of each circle shows the average amplitude of the errors, scaled to the Neck-MidHip segment. The colors separately represent each pose estimation method. (a) Real infants, video input (``Supine'' 720 annotations); (b) Real infants, video input (``Extended supine'': 1440 annotations); (c) Synthetic infants, video input}
    \label{fig:n-mh_circles_sup}
\end{figure}

Figure SF~\ref{fig:n-mh_circles_sup} shows the average Neck-MidHip errors on supine datasets, including DeepLabCut which was not shown in the main manuscript due to visual clarity concerns, for each individual keypoint with available ground truth.

All detailed values for individual keypoints on each dataset are also shown in Tables ST~\ref{tab:n-mh_supine_kps},~\ref{tab:n-mh_lap_kps},~\ref{tab:n-mh_syrip_kps}, and~\ref{tab:n-mh_minirgbd_kps}.

Across all methods, the face keypoints seem to be the most accurate, followed by the upper body keypoints. The legs are more challenging, especially the knees, with the highest errors occurring on the hips.

While the average Neck-MidHip errors for the human annotations are a little higher than for some of the pose estimation methods on the ``Supine'' dataset (Tab. ST.~\ref{tab:n-mh_supine_kps}), the detail per keypoints shows that humans have better estimations and less variability on legs compared to the arms but suffer from a larger lack of accuracy on the hips and on the shoulders, which brings the average errors up.

\begin{table}[!ht]%
    \addtolength{\tabcolsep}{-0.5em}
    \centering
    \begin{tabular}{c|c|c|c|c|c|c|c|c|c||c|}
    \textbf{R. Supine} & kp & AlphaPose & DeepLabCut & Detectron 2 & MediaPipe & HRNet BU & HRNet TD & OpenPose & ViTPose & Human\\
    \hline
    Images & Nose & N/A & N/A & N/A & N/A & N/A & N/A & N/A & N/A & {N/A} \\
    & LEye & 3.4 & N/A & 2.4 & 30.0 & \textbf{1.9} & \textbf{1.9} & 2.6 & N/A & {2.7} \\
    & REye & 3.2 & N/A & 2.1 & 39.1 & \textbf{1.7} & \textbf{1.7} & 3.2 & N/A & {2.3} \\
    & LEar & N/A & N/A & N/A & N/A & N/A & N/A & N/A & N/A & {N/A} \\
    & REAr & N/A & N/A & N/A & N/A & N/A & N/A & N/A & N/A & {N/A} \\
    & LShoulder & 6.6 & N/A & 7.4 & 19.7 & 7.3 & \textbf{5.7} & 7.6 & N/A & {12.2} \\
    & RShoulder & 6.5 & N/A & 6.4 & 17.8 & 6.7 & \textbf{5.8} & 7.0 & N/A & {10.0} \\
    & LElbow & 8.2 & N/A & 7.9 & 30.8 & 6.4 & \textbf{6.}3 & 6.9 & N/A & {5.9} \\
    & RElbow & 7.5 & N/A & 9.0 & 31.8 & 6.3 & \textbf{5.3} & 7.7 & N/A & {5.6} \\
    & LWrist & 6.5 & N/A & 6.0 & 58.1 & \textbf{5.0} & \textbf{5.0} & 5.8 & N/A & {5.1} \\
    & RWrist & 7.6 & N/A & 7.2 & 51.7 & 5.5 & \textbf{5.2} & 6.6 & N/A & {4.6} \\
    & LHip & \textbf{12.1} & N/A & 15.3 & 26.6 & 14.8 & 13.9 & 14.5 & N/A & {17.2} \\
    & RHip & 14.4 & N/A & 14.3 & 21.4 & \textbf{12.8} & 14.6 & 15.2 & N/A & {14.6} \\
    & LKnee & 10.7 & N/A & 15.6 & 49.7 & 12.4 & \textbf{9.0} & 14.8 & N/A & {6.8} \\
    & RKnee & 11.7 & N/A & 18.7 & 44.7 & 11.0 & \textbf{7.7} & 13.9 & N/A & {6.5} \\
    & LAnkle & 9.1 & N/A & 14.9 & 56.7 & 8.0 & \textbf{6.2} & 8.5 & N/A & {6.7} \\
    & RAnkle & 9.3 & N/A & 15.7 & 53.2 & 8.8 & \textbf{6.1} & 9.3 & N/A & {6.1} \\
    \hline
    Videos & Nose & N/A & N/A & N/A & N/A & N/A & N/A & N/A & N/A &  \\
    & LEye & 3.5 & N/A & 2.4 & 28.5 & \textbf{1.9} & \textbf{1.9} & 4.9 & 2.1 &  \\
    & REye & 3.3 & 2.2 & N/A & 3.8 & \textbf{1.7} & \textbf{1.7} & 5.8 & 2.1 &  \\
    & LEar & N/A & N/A & N/A & N/A & N/A & N/A & N/A & N/A &  \\
    & REAr & N/A & N/A & N/A & N/A & N/A & N/A & N/A & N/A &  \\
    & LShoulder & 6.6 & 68.6 & 7.7 & 17.7 & 7.1 & 5.7 & 8.5 & \textbf{4.8} &  \\
    & RShoulder & 6.2 & 73.5 & 6.4 & 16.0 & 6.6 & 5.8 & 8.4 & \textbf{5.3} &  \\
    & LElbow & 8.2 & 119.8 & 8.0 & 28.8 & 6.4 & 6.0 & 10.7 & \textbf{5.6} &  \\
    & RElbow & 7.4 & 91.1 & 9.4 & 29.5 & 6.0 & 5.5 & 11.4 & \textbf{5.3} &  \\
    & LWrist & 7.0 & 147.6 & 6.2 & 56.5 & 5.1 & \textbf{4.6} & 12.5 & 4.9 &  \\
    & RWrist & 7.8 & 110.3 & 7.2 & 53.2 & 5.4 & \textbf{5.2} & 12.3 & 5.3 &  \\
    & LHip & 11.7 & 62.4 & 15.3 & 22.1 & 14.6 & 13.9 & 15.1 & \textbf{11.2} &  \\
    & RHip & 14.1 & 68.6 & 14.0 & 16.7 & 12.9 & 14.6 & 15.4 & \textbf{12.1} &  \\
    & LKnee & 10.2 & 118.3 & 15.6 & 44.9 & 12.4 & 8.9 & 18.7 & \textbf{6.9} &  \\
    & RKnee & 10.9 & 19.5 & 127.3 & 40.4 & 11.8 & 7.9 & 19.4 & \textbf{7.2} &  \\
    & LAnkle & 8.7 & 102.2 & 16.5 & 51.7 & 8.5 & 6.2 & 13.8 & \textbf{5.2} &  \\
    & RAnkle & 9.9 & 106.0 & 16.4 & 47.2 & 10.7 & 6.2 & 15.8 & \textbf{5.3} &  \\
    \hline
    \end{tabular}
    \caption{Average errors for each individual keypoint location as a percentage of the mean Neck-MidHip segment length computed for per individual recording, for each method on our ``Supine'' dataset. The evaluation of the manually-annotated keypoints were computed over the 20\% of the dataset that were double coded to compute reliability.}
    \label{tab:n-mh_supine_kps}
\end{table}

\begin{table}[!ht]%
    \addtolength{\tabcolsep}{-0.3em}
    \centering
    \begin{tabular}{c|c|c|c|c|c|c|c|c}
    \textbf{Lap} & AlphaPose & DeepLabCut & Detectron 2 & MediaPipe & HRNet BU & HRNet TD & OpenPose & ViTPose\\
    \hline
    Nose & 39.4 & N/A & 34.7 & 60.6 & \textbf{16.4} & 46.4 & 19.0 & 46.5 \\
    LEye & 41.3 & N/A & 35.4 & 66.7 & \textbf{17.8} & 49.8 & 20.2 & 47.6 \\
    REye & 40.8 & N/A & 36.0 & 73.4 & \textbf{18.0} & 49.4 & 20.7 & 47.9 \\
    LEar & 48.4 & N/A & 48.3 & 74.5 & \textbf{20.6} & 59.4 & \textbf{20.6} & 57.4 \\
    REAr & 27.6 & N/A & 19.6 & 59.9 & 18.4 & 22.6 & \textbf{18.1} & 24.2 \\
    LShoulder & 43.7 & N/A & 47.8 & 57.0 & \textbf{23.6} & 54.2 & 30.7 & 52.0 \\
    RShoulder & 44.7 & N/A & 47.0 & 57.5 & \textbf{21.7} & 56.5 & 30.9 & 52.7 \\
    LElbow & 35.0 & N/A & 38.7 & 48.9 & \textbf{22.2} & 42.4 & 24.6 & 41.0 \\
    RElbow & 34.4 & N/A & 37.2 & 42.4 & \textbf{20.2} & 35.3 & 28.4 & 34.4 \\
    LWrist & 30.3 & N/A & 31.4 & 43.9 & \textbf{19.0} & 32.3 & 20.3 & 33.6 \\
    RWrist & 33.4 & N/A & 33.2 & 40.2 & \textbf{21.7} & 33.6 & 25.1 & 33.4 \\
    LHip & 30.2 & N/A & 40.8 & 39.3 & \textbf{20.7} & 36.7 & 25.5 & 32.6 \\
    RHip & 28.8 & N/A & 35.3 & 36.5 & \textbf{17.5} & 34.7 & 22.5 & 29.1 \\
    LKnee & 40.5 & N/A & 48.2 & 52.9 & \textbf{34.5} & 42.5 & 40.1 & 35.2 \\
    RKnee & 39.5 & N/A & 56.2 & 46.3 & \textbf{30.5} & 41.5 & 36.7 & 35.2 \\
    LAnkle & 42.8 & N/A & 53.7 & 67.0 & 41.3 & 44.5 & \textbf{23.0} & 39.3 \\
    RAnkle & 44.8 & N/A & 63.8 & 71.0 & 49.6 & 46.8 & \textbf{27.3} & 37.9 \\
    \hline
    \end{tabular}
    \caption{Average errors for each individual keypoint location as a percentage of the Neck-MidHip segment computed per-frame for each method on our ``Lap'' dataset.}
    \label{tab:n-mh_lap_kps}
\end{table}

\begin{table}[!ht]%
    \addtolength{\tabcolsep}{-0.3em}
    \centering
    \begin{tabular}{c|c|c|c|c|c|c|c|c}
    \textbf{SyRIP} & AlphaPose & DeepLabCut & Detectron 2 & MediaPipe & HRNet BU & HRNet TD & OpenPose & ViTPose\\
    \hline
    Nose & 7.8 & N/A & 6.2 & 229.5 & 6.2 & 6.9 & \textbf{5.2} & 6.4 \\
    LEye & 8.3 & N/A & 6.8 & 230.7 & 7.0 & 6.7 & \textbf{4.9} & 6.7 \\
    REye & 7.4 & N/A & 6.9 & 230.0 & 6.8 & 6.1 & \textbf{4.8} & 6.2 \\
    LEar & 12.4 & N/A & 9.3 & 228.1 & 12.2 & 10.0 & \textbf{6.3} & 10.6 \\
    REar & 10.7 & N/A & 8.6 & 225.3 & 11.6 & 9.7 & \textbf{7.6} & 9.7 \\
    LShoulder & 11.5 & N/A & 11.3 & 224.4 & \textbf{9.8} & 10.0 & 9.9 & 9.9 \\
    RShoulder & 11.7 & N/A & 10.9 & 217.1 & 10.4 & 9.8 & 10.3 & \textbf{9.5} \\
    LElbow & 13.5 & N/A & 16.8 & 242.9 & 14.4 & 11.3 & 11.2 & \textbf{10.2} \\
    RElbow & 13.7 & N/A & 17.0 & 227.3 & 13.9 & 11.1 & 13.5 & \textbf{10.0} \\
    LWrist & 15.7 & N/A & 21.8 & 257.0 & 12.7 & 11.1 & 11.2 & \textbf{10.4} \\
    RWrist & 14.9 & N/A & 19.5 & 238.1 & 15.0 & 11.6 & 13.0 & \textbf{9.8} \\
    LHip & 22.1 & N/A & 26.1 & 244.7 & 21.7 & 19.3 & 26.1 & \textbf{18.0} \\
    RHip & 21.4 & N/A & 24.2 & 240.2 & 21.7 & 19.3 & 27.5 & \textbf{17.6} \\
    LKnee & 26.6 & N/A & 39.1 & 259.2 & 28.4 & 22.1 & 27.8 & \textbf{17.3} \\
    RKnee & 21.3 & N/A & 31.4 & 253.2 & 24.3 & 17.9 & 24.9 & \textbf{14.9} \\
    LAnkle & 27.8 & N/A & 39.7 & 297.2 & 29.6 & 26.0 & 28.1 & \textbf{20.4} \\
    RAnkle & 21.5 & N/A & 32.3 & 295.3 & 22.2 & 20.0 & 21.4 & \textbf{17.3} \\
    \hline
    \end{tabular}
    \caption{Average errors for each individual keypoint location as a percentage of the Neck-MidHip segment computed per-frame for each method on the SyRIP dataset.}
    \label{tab:n-mh_syrip_kps}
\end{table}

\begin{table}[!ht]%
    \addtolength{\tabcolsep}{-0.5em}
    \centering
    \begin{tabular}{c|c|c|c|c|c|c|c|c|c|}
    \textbf{Mini-RGBD} & kp & AlphaPose & DeepLabCut & Detectron 2 & MediaPipe & HRNet BU & HRNet TD & OpenPose & ViTPose\\
    \hline
    Images & Nose & \textbf{2.6} & N/A & 3.2 & 20.1 & 4.6 & \textbf{2.6} & 2.9 & N/A \\
    & LEye & N/A & N/A & N/A & N/A & N/A & N/A & N/A & N/A \\
    & REye & N/A & N/A & N/A & N/A & N/A & N/A & N/A & N/A \\
    & LEar & N/A & N/A & N/A & N/A & N/A & N/A & N/A & N/A \\
    & REAr & N/A & N/A & N/A & N/A & N/A & N/A & N/A & N/A \\
    & LShoulder & 7.4 & N/A & 6.4 & 15.2 & 9.0 & \textbf{5.8} & 7.3 & N/A \\
    & RShoulder & 6.5 & N/A & 6.2 & 14.3 & 8.4 & \textbf{5.6} & 7.8 & N/A \\
    & LElbow & 7.0 & N/A & 8.1 & 28.6 & 10.6 & \textbf{4.6} & 7.4 & N/A \\
    & RElbow & 6.0 & N/A & 6.1 & 24.2 & 9.7 & \textbf{4.5} & 7.9 & N/A \\
    & LWrist & 8.5 & N/A & 8.6 & 47.2 & 12.3 & \textbf{3.8} & 8.1 & N/A \\
    & RWrist & 7.8 & N/A & 5.8 & 41.8 & 12.1 & \textbf{4.2} & 8.7 & N/A \\
    & LHip & 18.1 & N/A & 20.7 & \textbf{15.4} & 18.5 & 18.3 & 20.4 & N/A \\
    & RHip & 19.0 & N/A & 20.1 & \textbf{16.2} & 19.5 & 19.1 & 22.9 & N/A \\
    & LKnee & 13.9 & N/A & 12.2 & 34.5 & 14.3 & \textbf{9.3} & 14.8 & N/A \\
    & RKnee & 11.3 & N/A & 10.1 & 36.0 & 11.8 & \textbf{7.7} & 12.7 & N/A \\
    & LAnkle & 9.1 & N/A & 8.4 & 34.9 & 10.1 & \textbf{5.7} & 8.2 & N/A \\
    & RAnkle & 10.5 & N/A & 9.1 & 38.1 & 11.3 & \textbf{7.2} & 9.5 & N/A \\
    \hline
    Videos & Nose & 3.1 & N/A & 3.2 & 20.4 & 3.0 & \textbf{2.6} & 3.1 & 3.1 \\
    & LEye & N/A & N/A & N/A & N/A & N/A & N/A & N/A & N/A \\
    & REye & N/A & N/A & N/A & N/A & N/A & N/A & N/A & N/A \\
    & LEar & N/A & N/A & N/A & N/A & N/A & N/A & N/A & N/A \\
    & REAr & N/A & N/A & N/A & N/A & N/A & N/A & N/A & N/A \\
    & LShoulder & 8.5 & 30.2 & 6.1 & 17.8 & 9.7 & 6.7 & 6.9 & \textbf{6.6} \\
    & RShoulder & 7.8 & 25.8 & 6.4 & 17.1 & 9.3 & 6.6 & 7.5 & \textbf{5.5} \\
    & LElbow & 9.6 & 46.7 & 10.4 & 33.5 & 14.6 & 6.4 & 8.6 & \textbf{4.7} \\
    & RElbow & 8.4 & 37.1 & 8.0 & 29.1 & 11.7 & 6.1 & 8.0 & \textbf{4.4} \\
    & LWrist & 13.0 & 66.6 & 14.2 & 53.1 & 17.5 & 7.4 & 10.1 & \textbf{5.1} \\
    & RWrist & 11.6 & 55.7 & 10.2 & 48.2 & 14.3 & 7.1 & 9.4 & \textbf{4.8} \\
    & LHip & 18.4 & 27.3 & 21.2 & \textbf{16.0} & 19.3 & 18.8 & 20.4 & 17.1 \\
    & RHip & 19.2 & 26.8 & 20.2 & \textbf{16.9} & 19.9 & 19.4 & 23.0 & 18.9 \\
    & LKnee & 15.7 & 47.7 & 14.0 & 37.7 & 17.5 & 10.6 & 16.1 & \textbf{9.0} \\
    & RKnee & 13.1 & 43.3 & 11.4 & 39.0 & 14.6 & 9.7 & 14.6 & \textbf{7.9} \\
    & LAnkle & 11.3 & 54.5 & 9.8 & 36.8 & 12.1 & \textbf{7.3} & 9.4 & 8.7 \\
    & RAnkle & 11.8 & 60.4 & 11.4 & 39.8 & 12.0 & \textbf{8.9} & 10.5 & 10.4 \\
    \hline
    \end{tabular}
    \caption{Average errors for each individual keypoint location as a percentage of the mean Neck-MidHip segment length computed for per individual recording, for each method on the MINI-RGBD dataset.}
    \label{tab:n-mh_minirgbd_kps}
\end{table}

\clearpage

\subsection*{Redundant detections}
The complete table with the percentage of redundant detections is shown in Tab. ST.~\ref{tab:real_rdet} and ST.~\ref{tab:synth_rdet} for real and synthetic infants in supine position, respectively.

For real infants, it is difficult to estimate the difficulty of each sequence exactly. We observe that video input has a tendency to produce more redundant detections.

For synthetic infants, we observe that Detectron2 and HRNet TD seem to produce more redundant detections when the sequence is difficult (IDs 10-12), while HRNet BU seems to produce more redundant detections when the sequence is easy (IDs 1-4). We observe a tendency to produce more redundant detections with video input than for image input. The percentage could not be computed for MediaPipe on Synthetic infant ID 2 because it did not provide a single detection for any image on this particular sequence, making the redundant detection percentage not computable.

\begin{table}[!ht]%
    \addtolength{\tabcolsep}{-0.3em}
    \centering
    \begin{tabular}{c|c|c|c|c|c|c|c|c|c|}
    \textbf{Real} & Video ID & AlphaPose & DeepLabCut & Detectron 2 & MediaPipe & HRNet BU & HRNet TD & OpenPose & ViTPose\\
    \hline
    Images 
    &AA\_8w       & \textbf{0} & N/A & 11.5 & \textbf{0} & 1.4 & 1.1 & \textbf{0} & N/A \\
    &AA\_17w     & 0.6 & N/A & 8.0 & \textbf{0} & 1.1 & 40.5 & \textbf{0} & N/A \\
    &TH\_8w\_st1 & \textbf{0} & N/A & 9.9 & \textbf{0} & 0.5 & 3.3 & \textbf{0} & N/A \\
    &TH\_8w\_st2 & \textbf{0} & N/A & 34.6 & \textbf{0} & 5.3 & 22.1 & \textbf{0} & N/A \\
    &TH\_8w\_st3 & 1.4 & N/A & 34.3 & \textbf{0} & 2.6 & 12.1 & \textbf{0} & N/A \\
    &TH\_15w    & 14.4 & N/A & 86.3 & \textbf{0} & 11.8 & 103.5 & \textbf{0} & N/A \\
    &TH\_19w    & 0.4 & N/A & 28.9 & \textbf{0} & 7.8 & 53.2 & \textbf{0} & N/A \\
    &TH\_25w    & 0.4 & N/A & 7.1 & \textbf{0} & 5.1 & 54.6 & \textbf{0} & N/A \\

    \hline
    Videos 
    &AA\_8w         & \textbf{0} & \textbf{0} & 92.0 & \textbf{0} & 9 & 1.0 & \textbf{0} & 1.4 \\
    &AA\_11w       & 0.3 & \textbf{0} & 199.8 & \textbf{0} & 0.3 & 44.9 & \textbf{0} & 0.1 \\
    &AA\_13w       & 4.7 & \textbf{0} & 290.9 & \textbf{0} & 4.9 & 66.7 & \textbf{0} & 4.0 \\
    &AA\_17w       & 1.2 & \textbf{0} & 170.1 & \textbf{0} & 1.4 & 47.7 & \textbf{0} & 6.4 \\
    &AA\_19w      & 6.7 & \textbf{0} & 308.6 & \textbf{0} & 10.0 & 26.4 & \textbf{0} & 4.0 \\
    &TH\_8w\_st1  & 0.1 & \textbf{0} & 197.0 & \textbf{0} & 0.9 & 1.9 & \textbf{0} & 2.9\\
    &TH\_8w\_st2   & \textbf{0} & \textbf{0} & 188.7 & \textbf{0} & 6.5 & 20.6 & \textbf{0} & 38.6\\
    &TH\_8w\_st3   & 0.9 & \textbf{0} & 249.2 & \textbf{0} & 2.9 & 12.7 & \textbf{0} & 26.6 \\
    &TH\_10w\_st1 & 0.1 & \textbf{0} & 197.0 & \textbf{0} & 0.1 & 7.7 & \textbf{0} & 22.5 \\
    &TH\_10w\_st2 & 1.2 & \textbf{0} & 388.0 & \textbf{0} & 3.3 & 130.0 & \textbf{0} & 82.3 \\
    &TH\_12w      & 0.1 & \textbf{0} & 164.4 & \textbf{0} & 0.3 & 5.4 & \textbf{0} & 5.4 \\
    &TH\_15w      & 15.2 & \textbf{0} & 522.8 & \textbf{0} & 13.9 & 98.8 & \textbf{0} & 28.8 \\
    &TH\_17w      & 2.4 & \textbf{0} & 198.9 & \textbf{0} & 9.4 & 101.2 & \textbf{0} & 40.3 \\
    &TH\_19w      & 0.7 & \textbf{0} & 257.6 & \textbf{0} & 8.2 & 51.5 & \textbf{0} & 2.7 \\
    &TH\_23w       & 51.9 & \textbf{0} & 198.0 & \textbf{0} & 2.3 & 91.0 & \textbf{0} & 43.9 \\
    &TH\_25w      & 0.5 & \textbf{0} & 211.3 & \textbf{0} & 4.9 & 43.1 & \textbf{0} & 36.6 \\
    \hline
    \end{tabular}
    \caption{Percentage of redundant detections for each method and input type on real infants videos.}
    \label{tab:real_rdet}
\end{table}

\begin{table}[!ht]%
    \addtolength{\tabcolsep}{-0.2em}
    \centering
    \begin{tabular}{c|c|c|c|c|c|c|c|c|c}
    \textbf{Synth.} & Video ID & AlphaPose & DeepLabCut & Detectron 2 & MediaPipe & HRNet BU & HRNet TD & OpenPose & ViTPose\\
    \hline
    Images & Syn. 1 & 0.3 & N/A & 5.5 & \textbf{0} & 56.8 & 0.6 & \textbf{0} & N/A \\
    & Syn. 2 & 0.5 & N/A & 1.3 & N/A & \textbf{0} & 0.6 & \textbf{0} & N/A \\
    & Syn. 3 & \textbf{0} & N/A & \textbf{0} & \textbf{0} & \textbf{0} & 0.2 & \textbf{0} & N/A \\
    & Syn. 4 & \textbf{0} & N/A & 0.1 & \textbf{0} & 20.7 & 3.6 & \textbf{0} & N/A \\
    & Syn. 5 & \textbf{0} & N/A & \textbf{0} & \textbf{0} & 3.6 & 1.6 & \textbf{0} & N/A \\
    & Syn. 6 & \textbf{0} & N/A & \textbf{0} & \textbf{0} & \textbf{0} & 7.1 & \textbf{0} & N/A \\
    & Syn. 7 & \textbf{0} & N/A & 2.0 & \textbf{0} & 0.1 & 0.3 & \textbf{0} & N/A \\
    & Syn. 8 & \textbf{0} & N/A & 0.5 & \textbf{0} & 1.9 & 0.5 & \textbf{0} & N/A \\
    & Syn. 9 & \textbf{0} & N/A & 29.0 & \textbf{0} & 6.2 & 95.4 & \textbf{0} & N/A \\
    & Syn. 10 & 0.1 & N/A & 240.2 & \textbf{0} & 1.9 & 101.6 & \textbf{0} & N/A \\
    & Syn. 11 & \textbf{0} & N/A & 102.2 & \textbf{0} & 1.0 & 0.4 & \textbf{0} & N/A \\
    & Syn. 12 & \textbf{0} & N/A & 31.7 & \textbf{0} & 7.6 & 106.8 & \textbf{0} & N/A \\
    \hline
    \hline
    Videos & Syn. 1 & 0.5 & \textbf{0} & 20.5 & \textbf{0} & 61.8 & 1.5 & \textbf{0} & 2.6\\
    & Syn. 2 & \textbf{0} & \textbf{0} & 38.1 & N/A & \textbf{0} & 2.7 & \textbf{0} & \textbf{0} \\
    & Syn. 3 & \textbf{0} & \textbf{0} & \textbf{0} & \textbf{0} & 0.1 & 0.1 & \textbf{0} & \textbf{0} \\
    & Syn. 4 & 1 & \textbf{0} & 92.2 & \textbf{0} & 31.3 & 15.1 & \textbf{0} & 2.4 \\
    & Syn. 5 & \textbf{0} & \textbf{0} & 13.6 & \textbf{0} & 9.1 & 3.7 & \textbf{0} & 3.6\\
    & Syn. 6 & \textbf{0} & \textbf{0} & 42.2 & \textbf{0} & 0.9 & 24.0 & \textbf{0} & 9.2\\
    & Syn. 7 & \textbf{0} & \textbf{0} & 98.7 & \textbf{0} & 1.3 & 0.9 & \textbf{0} & 0.1\\
    & Syn. 8 & \textbf{0} & \textbf{0} & 70.7 & \textbf{0} & 2.9 & 2.2 & \textbf{0} & \textbf{0}\\
    & Syn. 9 & 0.1 & \textbf{0} & 97.2 & \textbf{0} & 23.7 & 93.1 & \textbf{0} & 50.4\\
    & Syn. 10 & 0.4 & \textbf{0} & 378.1 & \textbf{0} & 3.3 & 30.6 & \textbf{0} & 2.5\\
    & Syn. 11 & \textbf{0} & \textbf{0} & 137.1 & \textbf{0} & 1.0 & 0.8 & \textbf{0} & 74.5\\
    & Syn. 12 & 0.1 & \textbf{0} & 121.6 & \textbf{0} & 5.8 & 110.5 & \textbf{0} & 30.5\\
    \hline
    \end{tabular}
    \caption{Percentage of redundant detections for each method and input type on synthetic infants.}
    \label{tab:synth_rdet}
\end{table}

\subsection*{Correlations between scores and OKS}

\begin{figure}[!htb]
\centering
\begin{subfigure}{.25\textwidth}
  \centering \includegraphics[scale=1, width=1\linewidth]{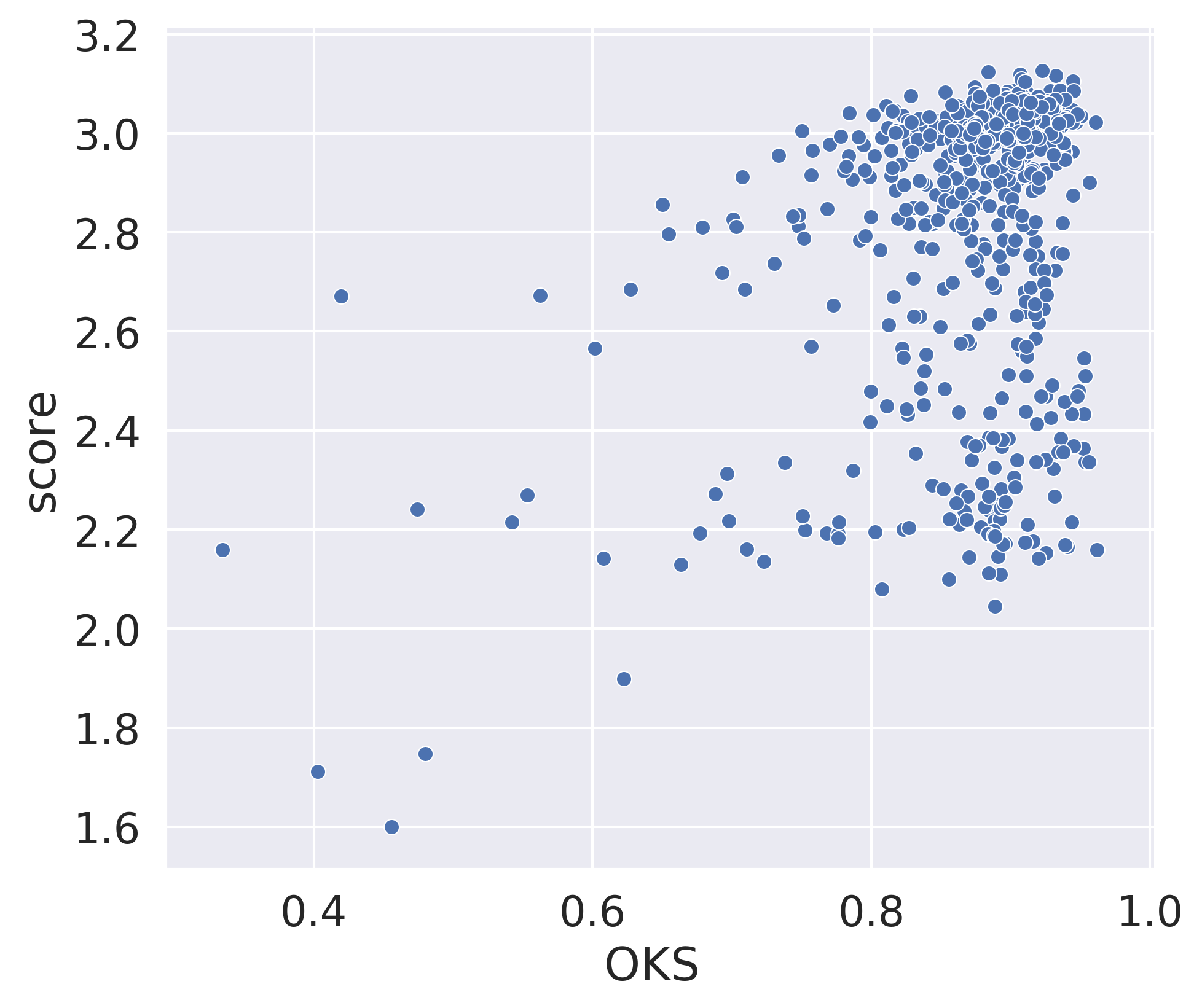}
  \small{(a) Alphapose -- img}
\end{subfigure}%
\begin{subfigure}{.25\textwidth}
  \centering \includegraphics[scale=1, width=1\linewidth]{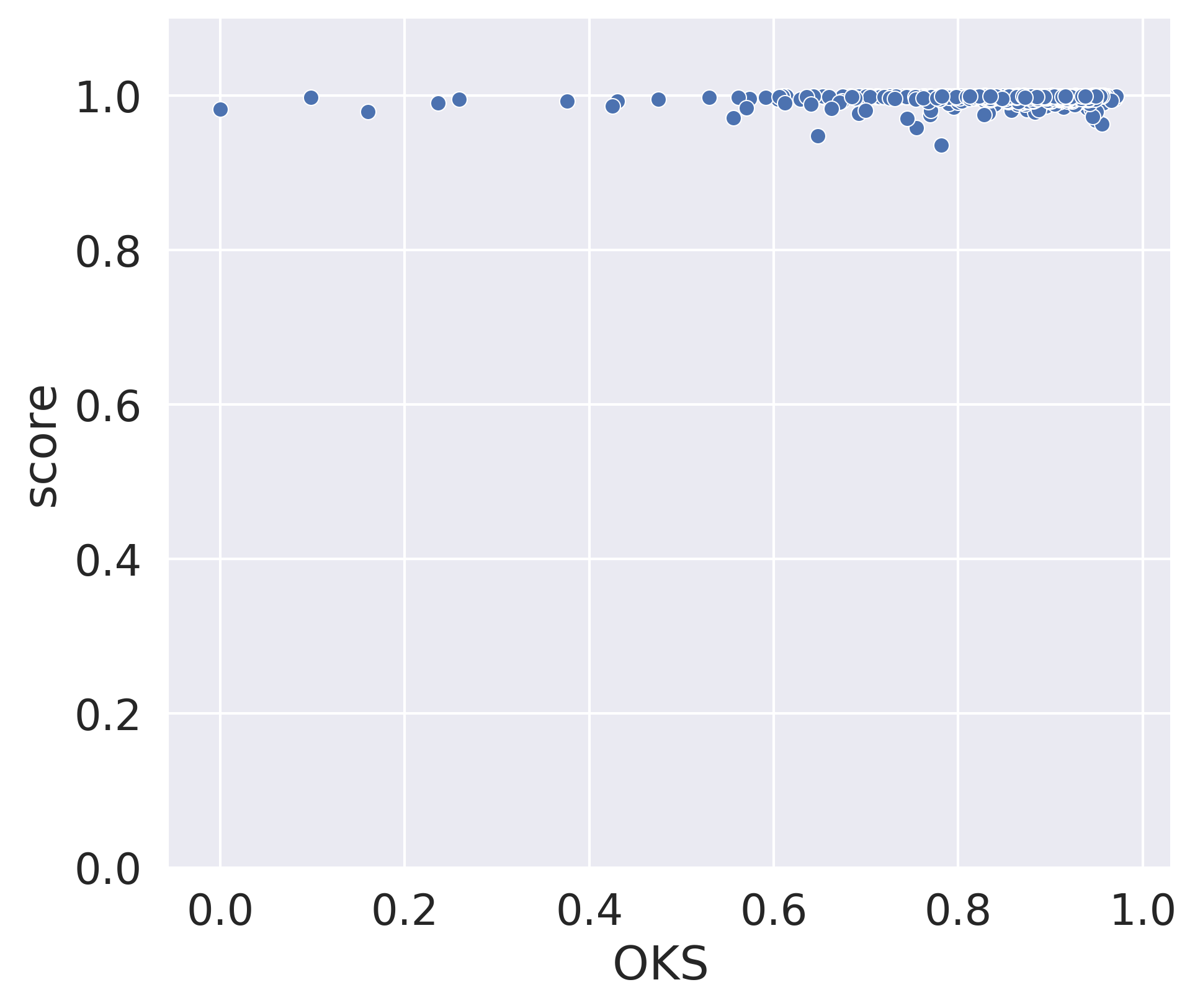}
  \small{(b) Detectron2 -- img}
\end{subfigure}%
\begin{subfigure}{.25\textwidth}
  \centering \includegraphics[scale=1, width=1\linewidth]{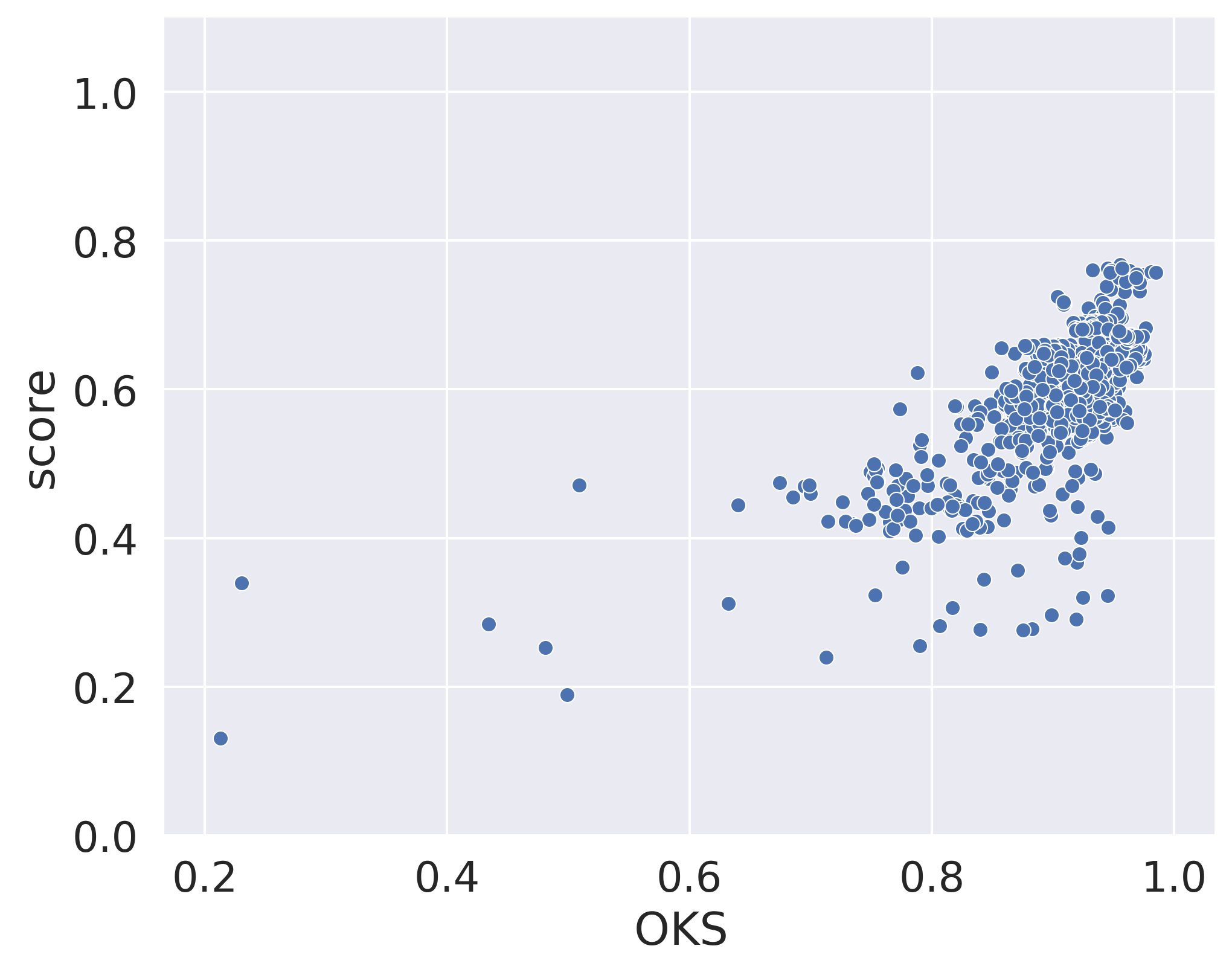}
  \small{(c) HRNet BU -- img}
\end{subfigure}%
\begin{subfigure}{.25\textwidth}
  \centering \includegraphics[scale=1, width=1\linewidth]{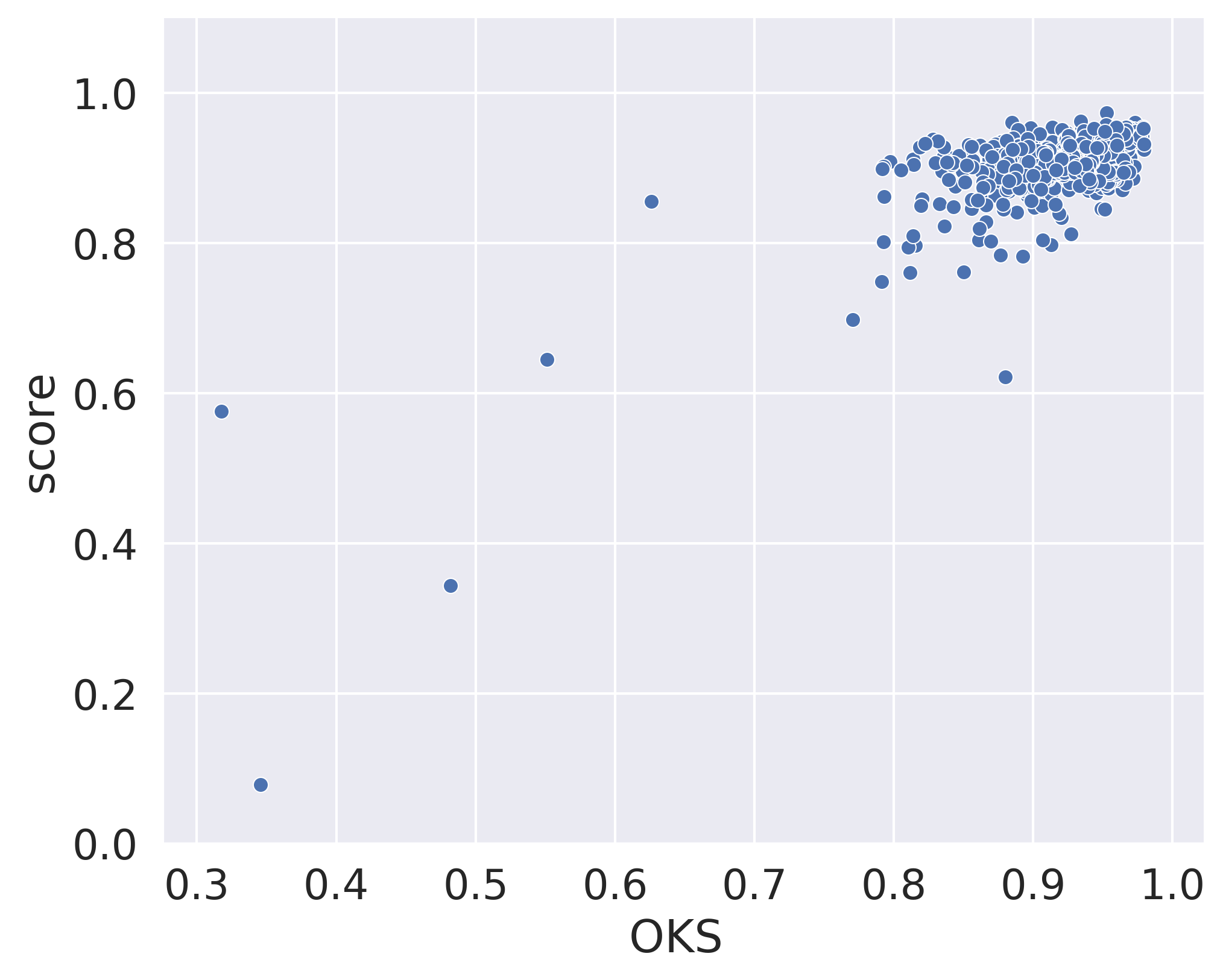}
  \small{(d) HRNet TD -- img}
\end{subfigure}%

\begin{subfigure}{.25\textwidth}
  \centering \includegraphics[scale=1, width=1\linewidth]{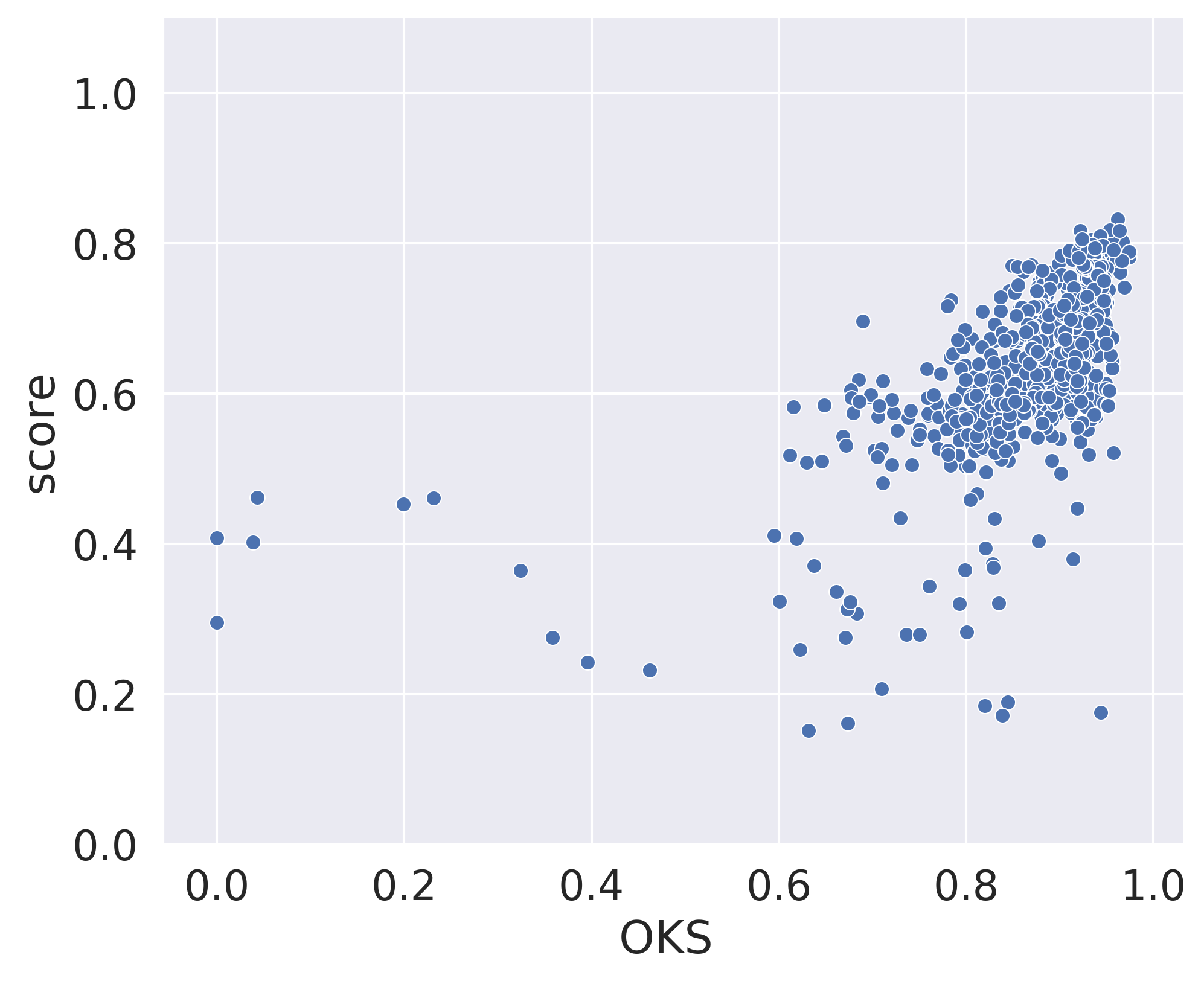}
  \small{(e) OpenPose -- img}
\end{subfigure}%
\begin{subfigure}{.25\textwidth}
  \centering \includegraphics[scale=1, width=1\linewidth]{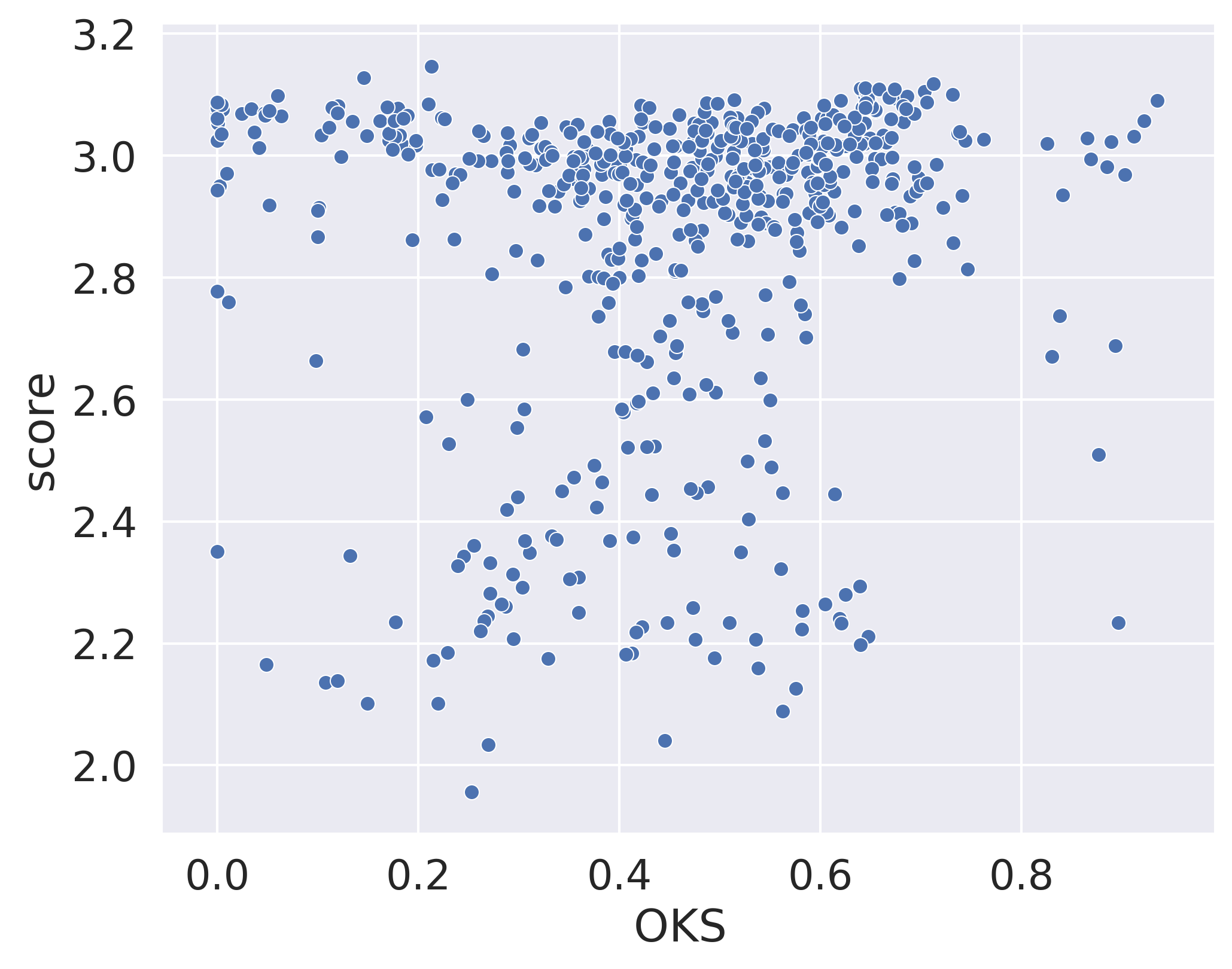}
  \small{(f) AlphaPose -- vid}
\end{subfigure}%
\begin{subfigure}{.25\textwidth}
  \centering \includegraphics[scale=1, width=1\linewidth]{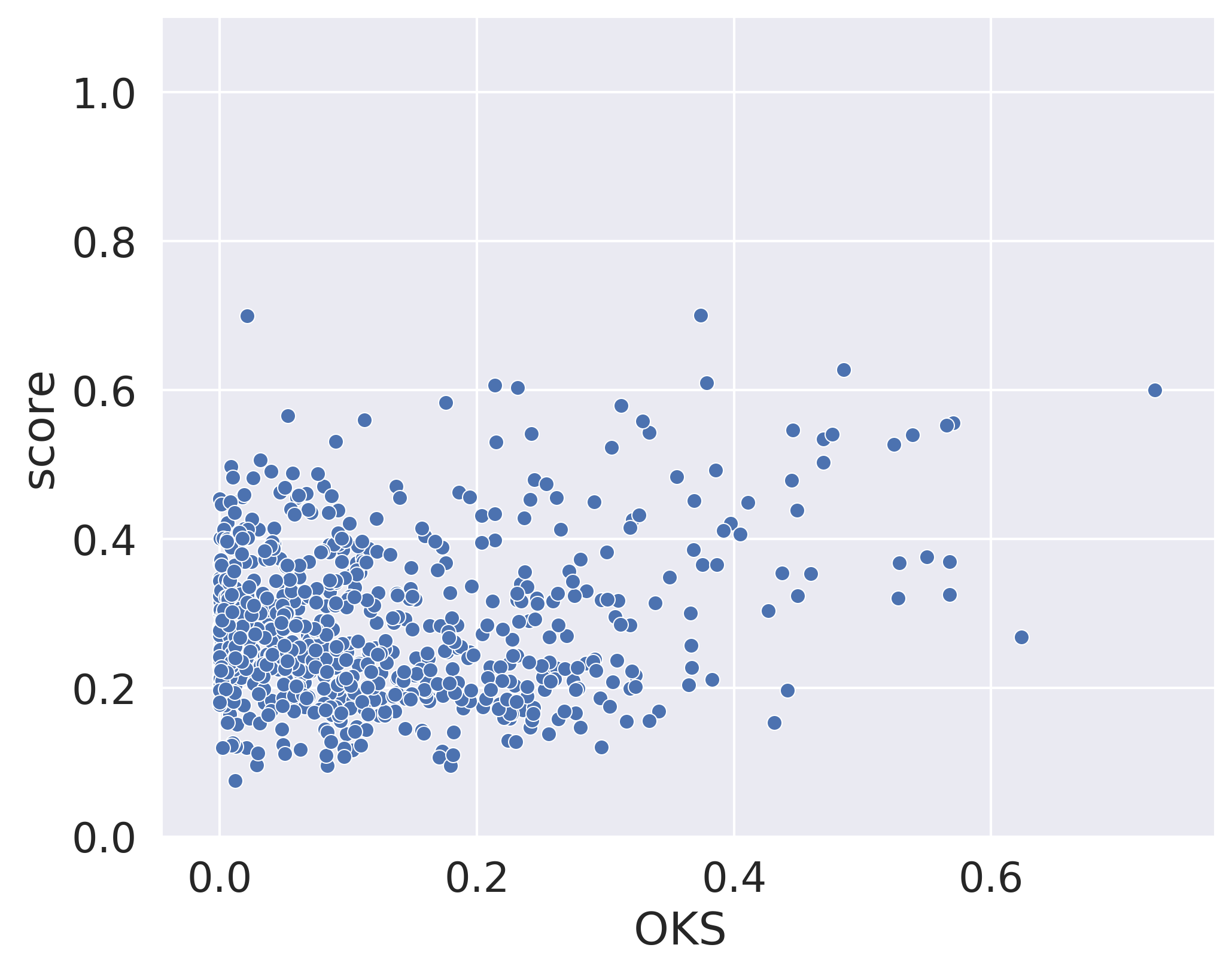}
  \small{(g) DeepLabCut -- vid}
\end{subfigure}%
\begin{subfigure}{.25\textwidth}
  \centering \includegraphics[scale=1, width=1\linewidth]{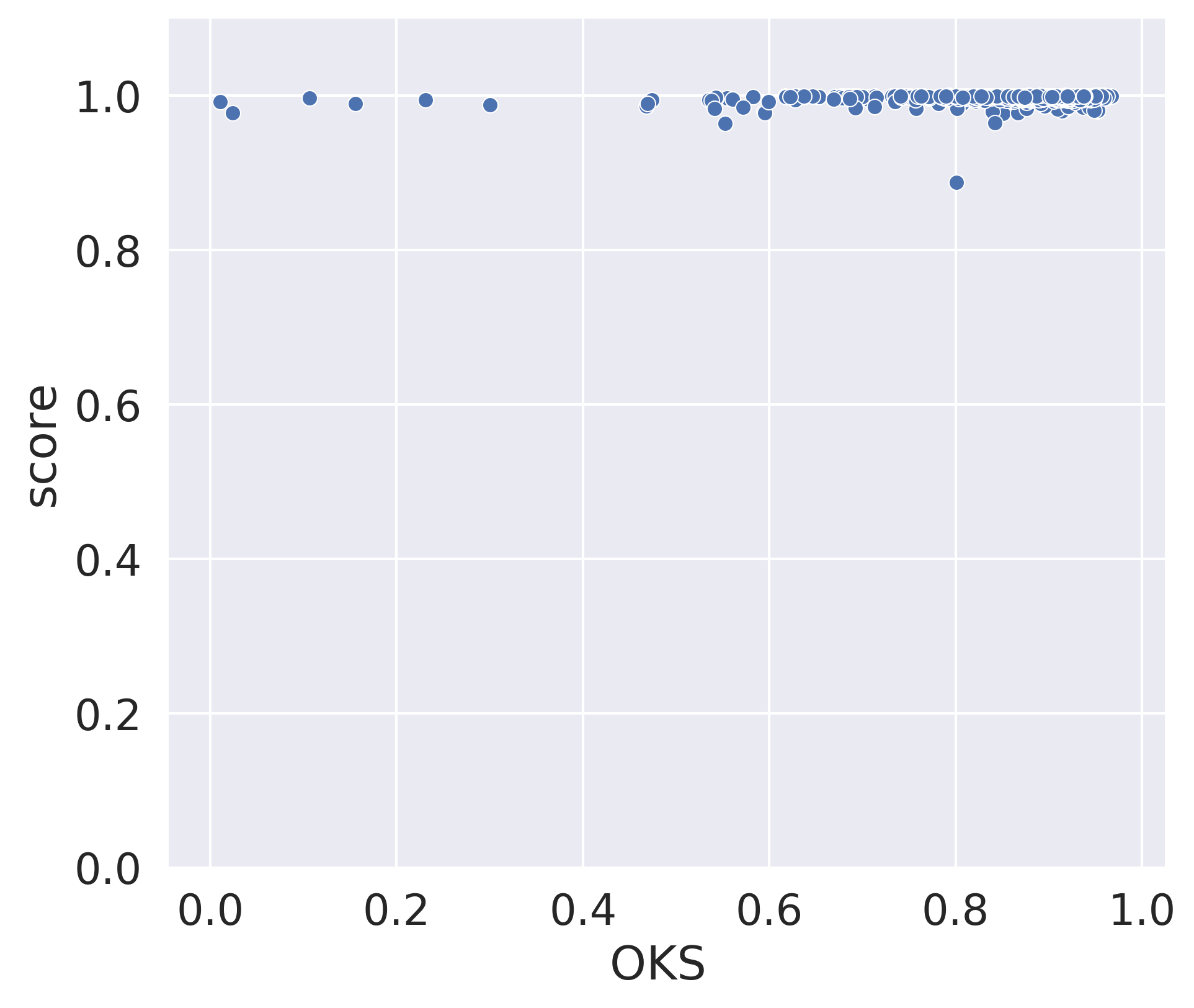}
  \small{(h) Detectron2 -- vid}
\end{subfigure}%

\begin{subfigure}{.25\textwidth}
  \centering \includegraphics[scale=1, width=1\linewidth]{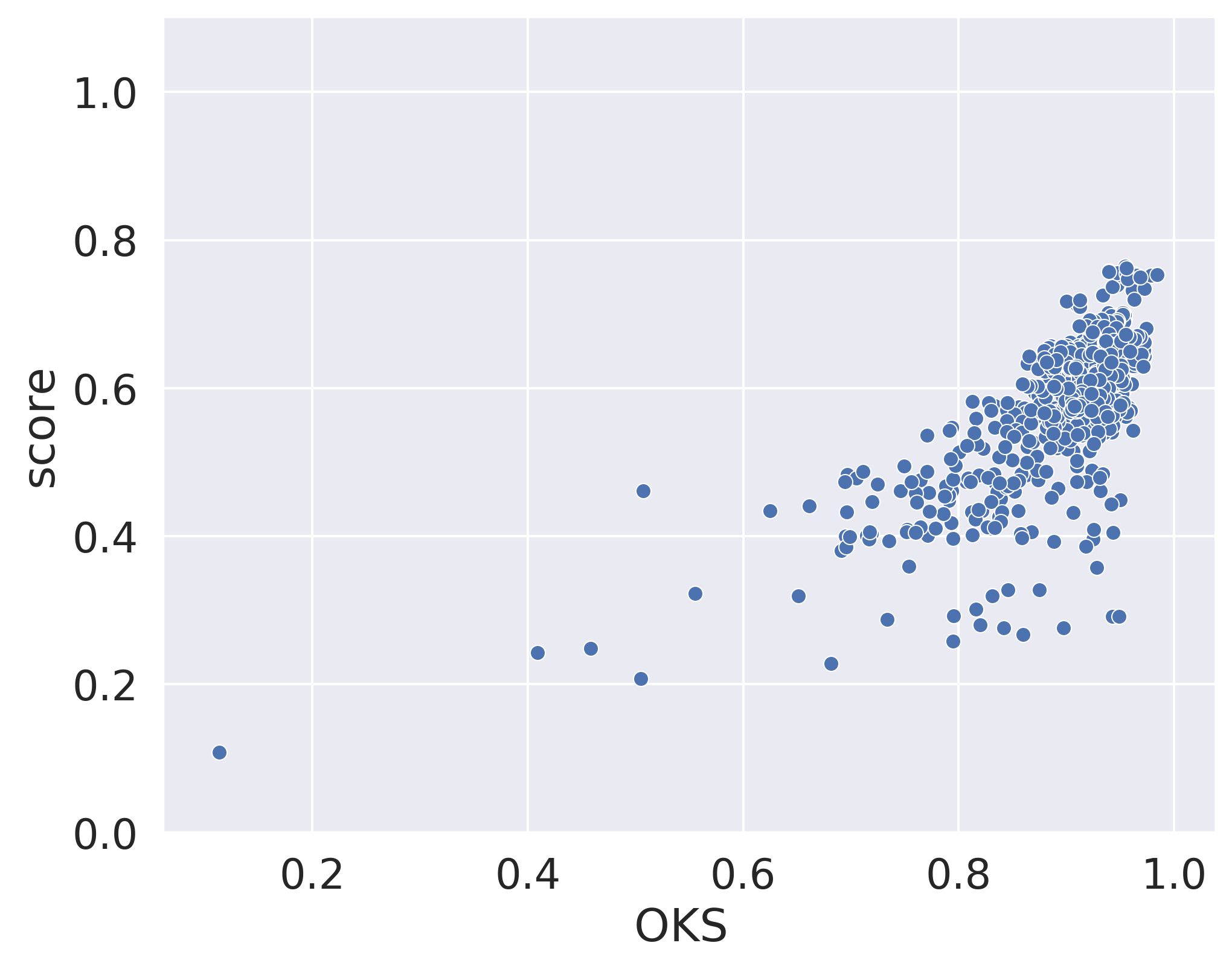}
  \small{(i) HRNet BU -- vid}
\end{subfigure}%
\begin{subfigure}{.25\textwidth}
  \centering \includegraphics[scale=1, width=1\linewidth]{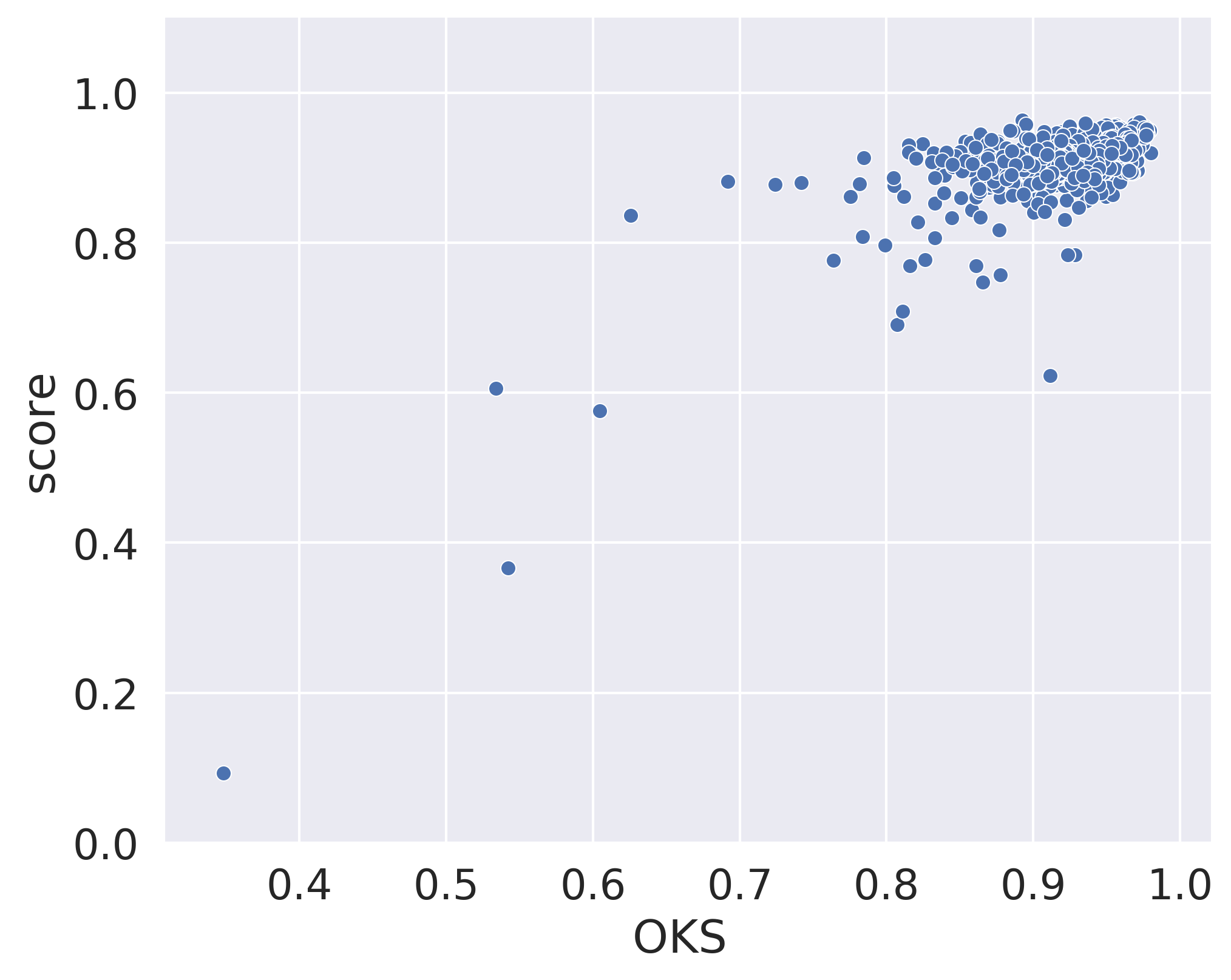}
  \small{(j) HRNet TD -- vid}
\end{subfigure}%
\begin{subfigure}{.25\textwidth}
  \centering \includegraphics[scale=1, width=1\linewidth]{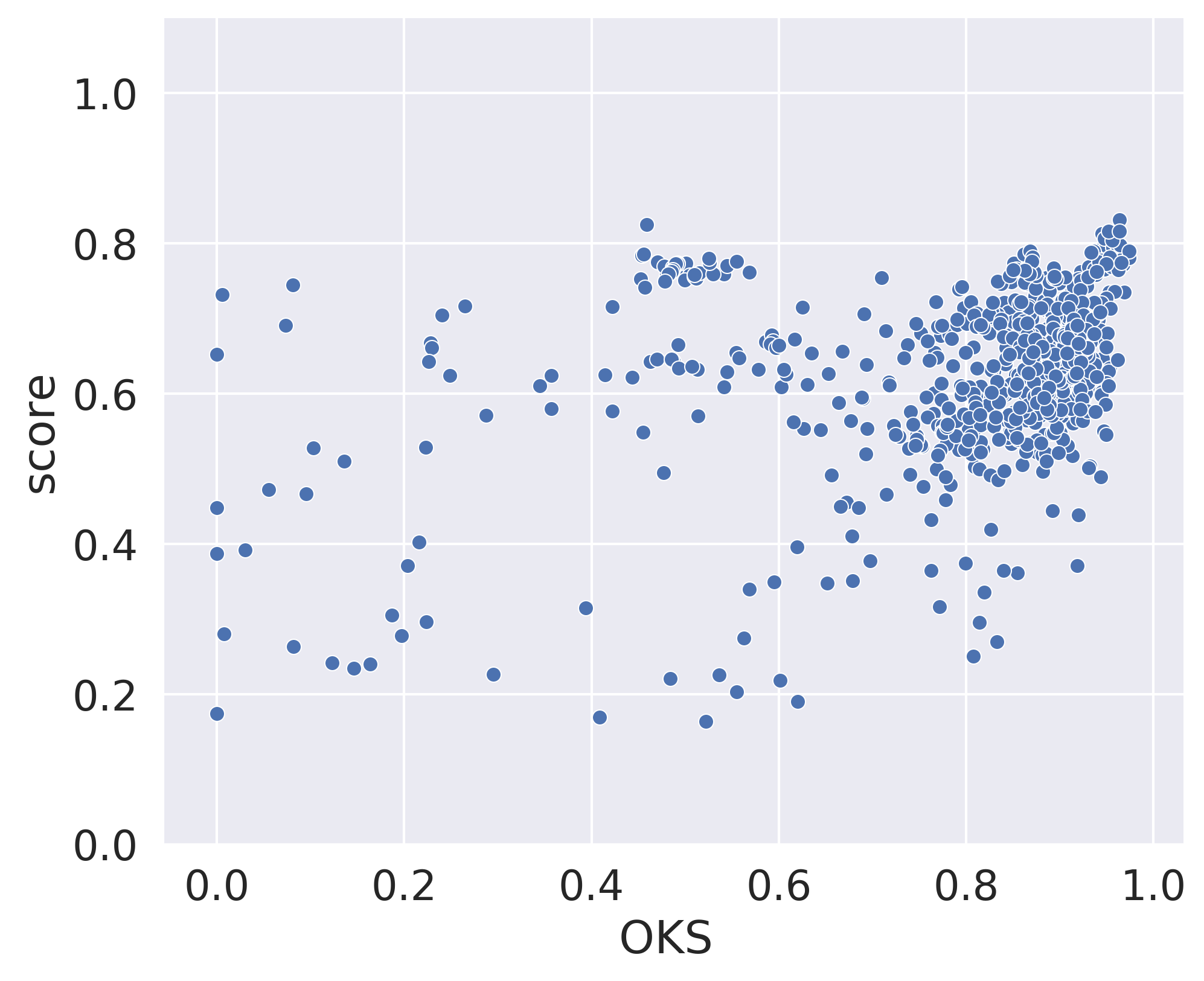}
  \small{(k) OpenPose -- vid}
\end{subfigure}%
\begin{subfigure}{.25\textwidth}
  \centering \includegraphics[scale=1, width=1\linewidth]{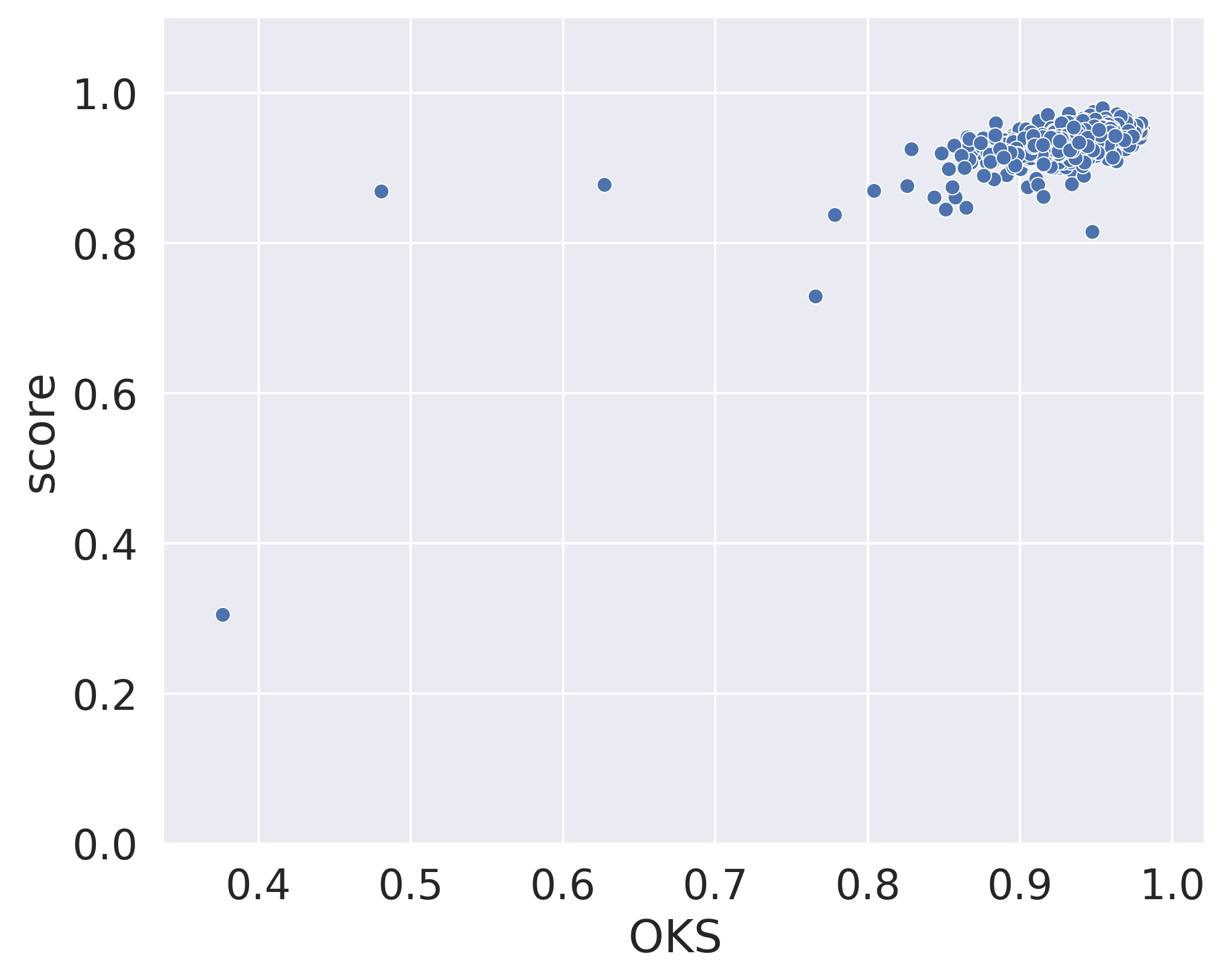}
  \small{(l) ViTPose -- vid}
\end{subfigure}%

\caption{Scatterplots of score and OKS values for each methods on our ``Supine'' dataset, (a-e) with image input; (f-l) with video input.}
\label{fig:scatter_sco_oks_suprea}
\end{figure}

\begin{figure}[!htb]
\centering
\begin{subfigure}{.33\textwidth}
  \centering \includegraphics[scale=1, width=1\linewidth]{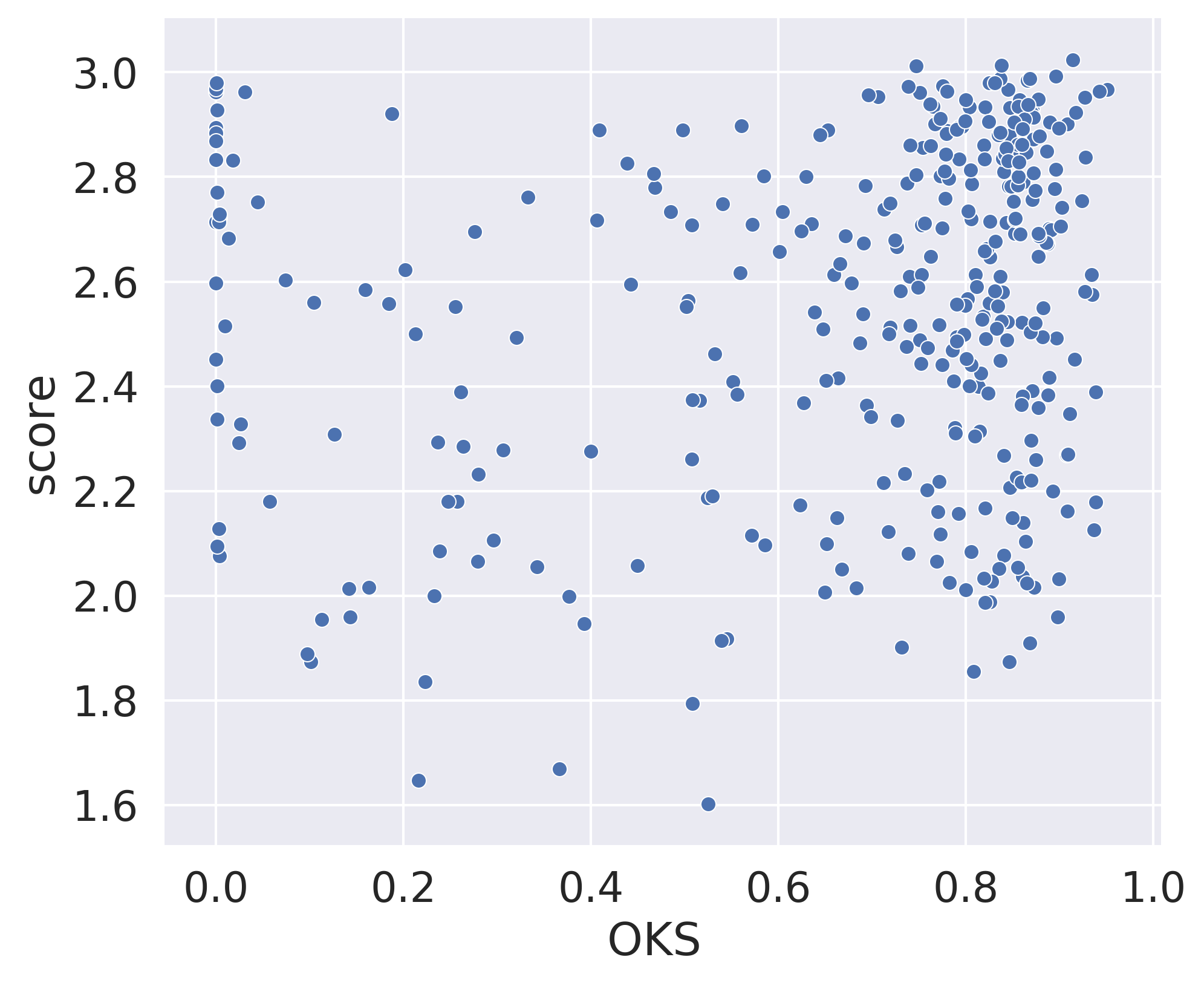}
  \small{(a) AlphaPose}
\end{subfigure}%
\begin{subfigure}{.33\textwidth}
  \centering \includegraphics[scale=1, width=1\linewidth]{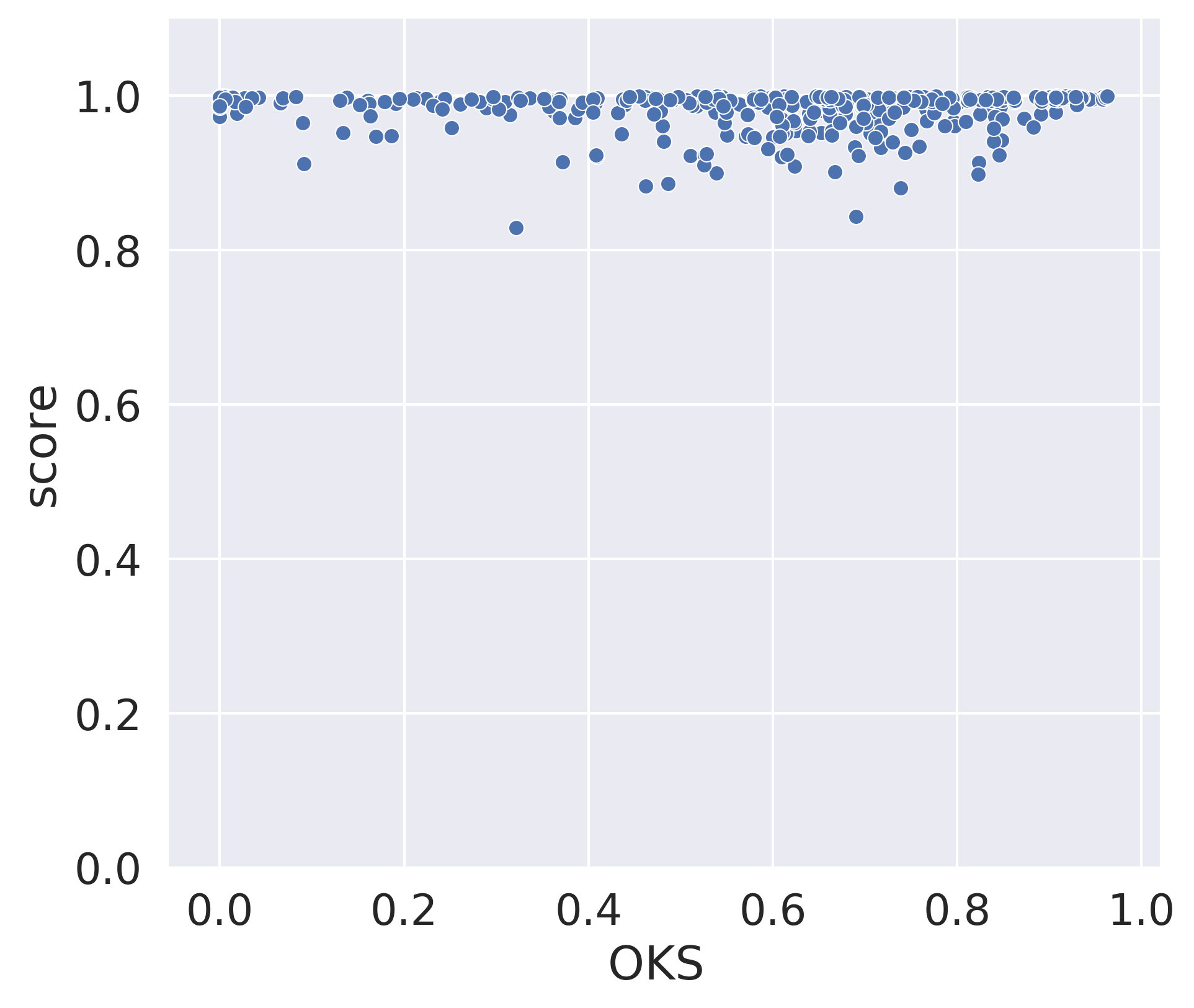}
  \small{(b) Detectron2}
\end{subfigure}%
\begin{subfigure}{.33\textwidth}
  \centering \includegraphics[scale=1, width=1\linewidth]{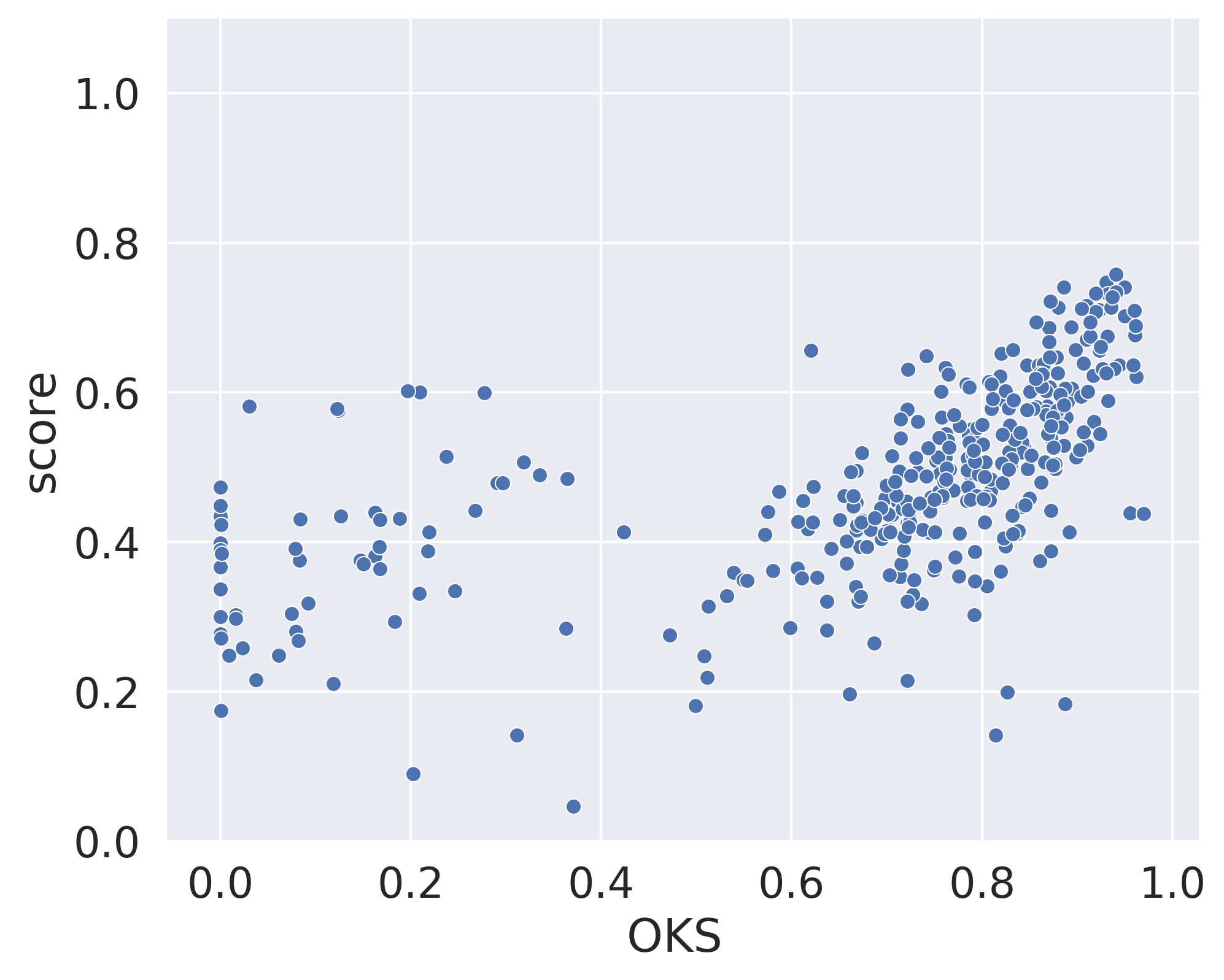}
  \small{(c) HRNet BU}
\end{subfigure}%

\begin{subfigure}{.33\textwidth}
  \centering \includegraphics[scale=1, width=1\linewidth]{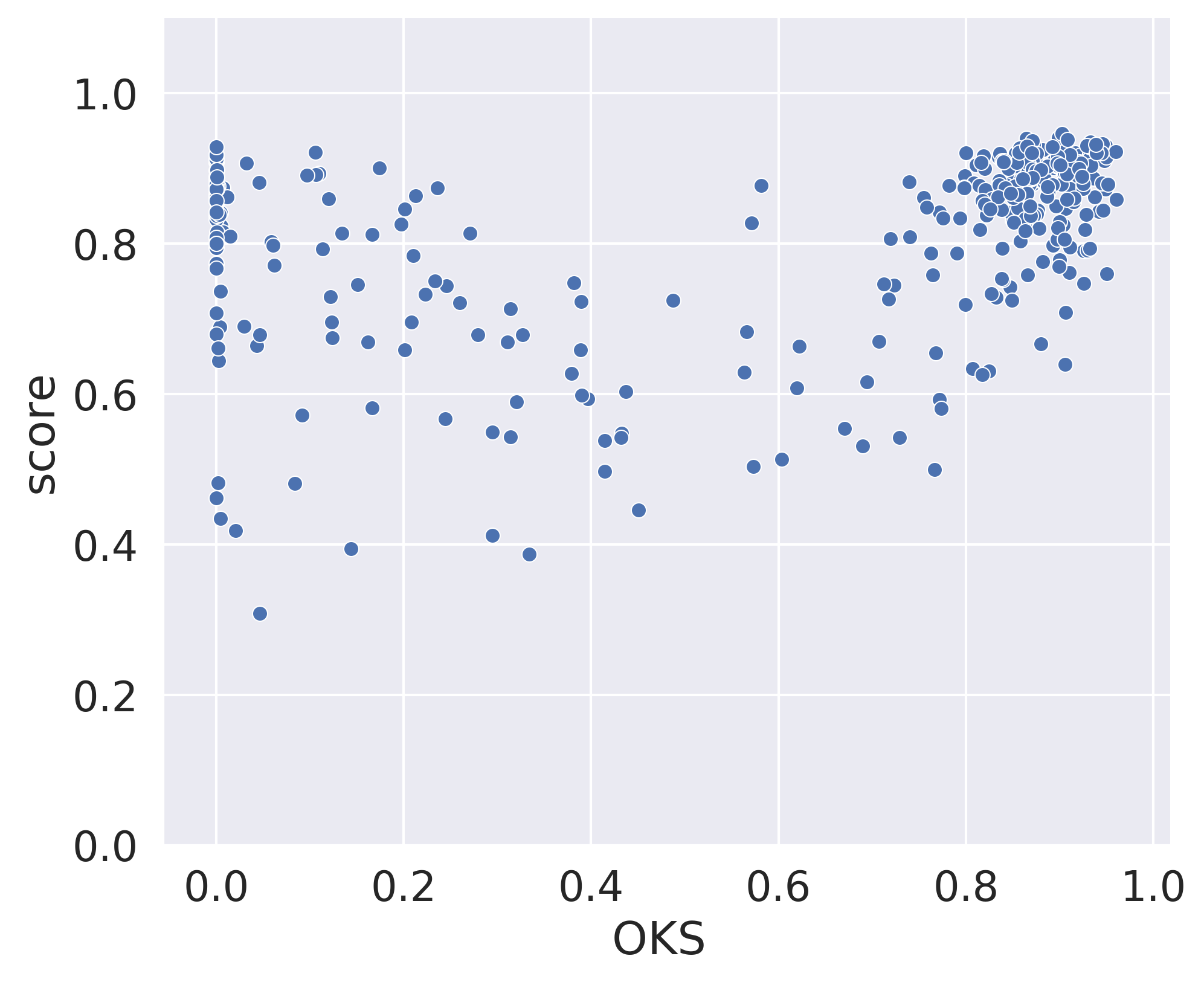}
  \small{(d) HRNet TD}
\end{subfigure}%
\begin{subfigure}{.33\textwidth}
  \centering \includegraphics[scale=1, width=1\linewidth]{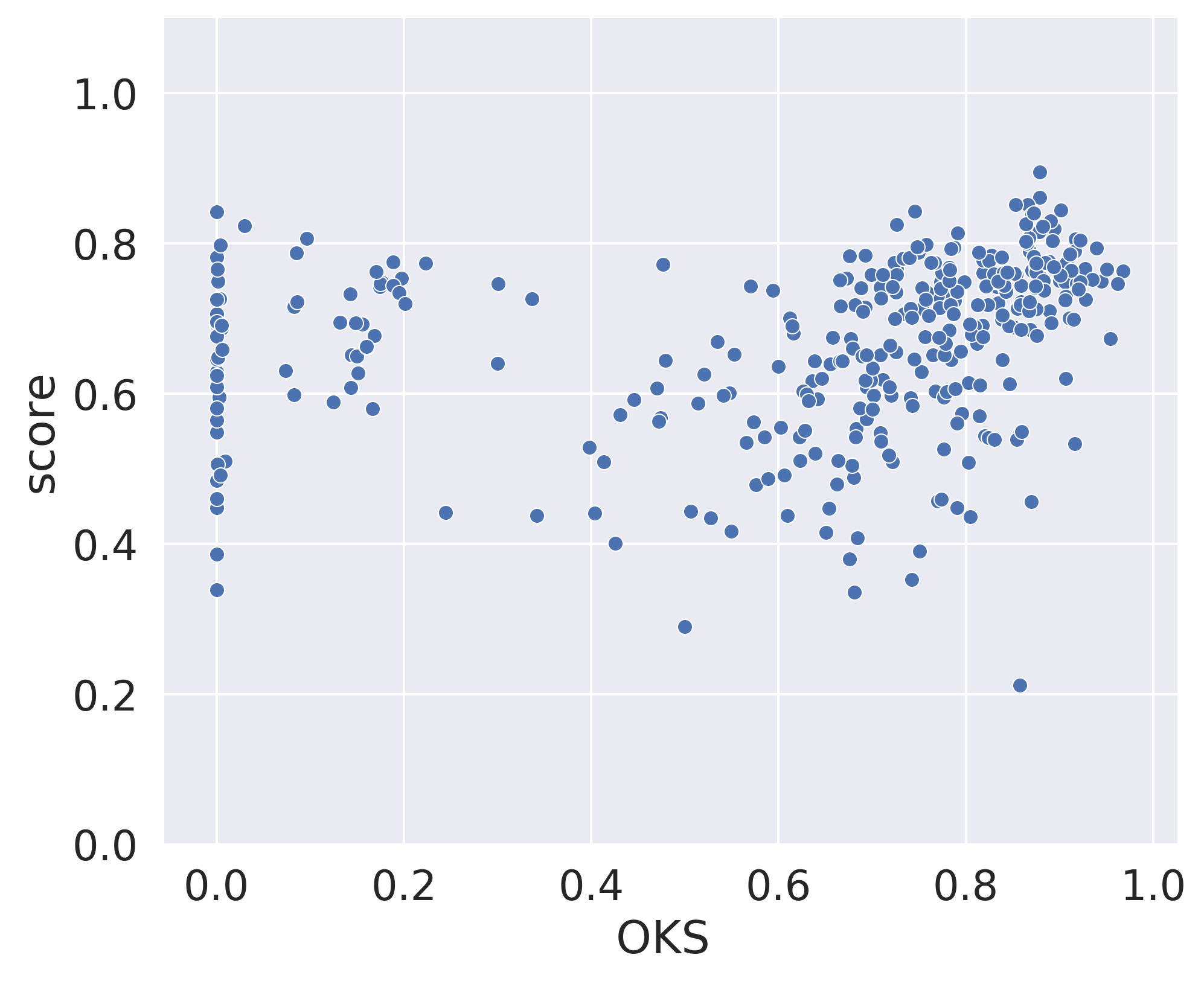}
  \small{(e) OpenPose}
\end{subfigure}%
\begin{subfigure}{.33\textwidth}
  \centering \includegraphics[scale=1, width=1\linewidth]{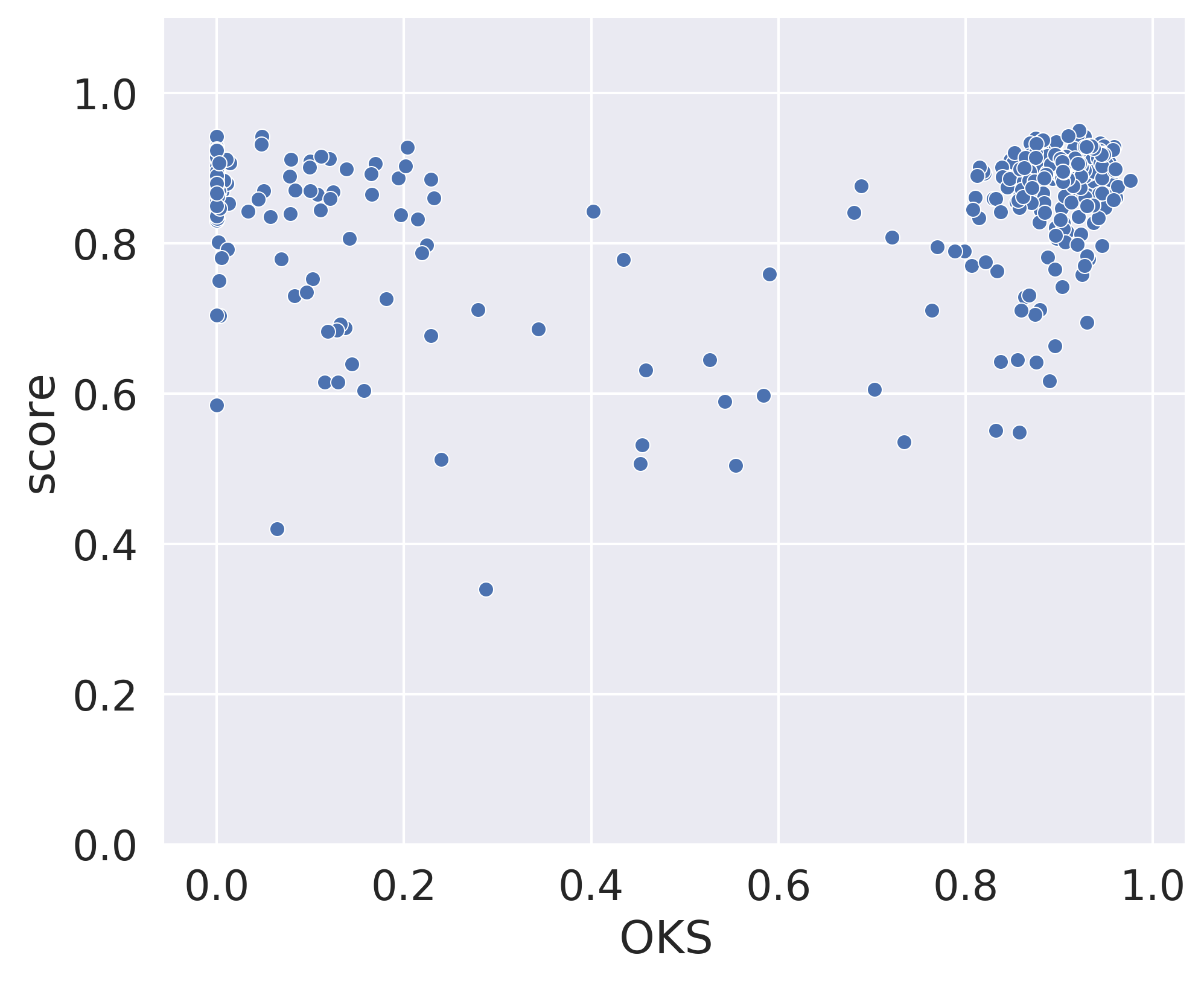}
  \small{(f) ViTPose}
\end{subfigure}%

\caption{Scatterplots of score and OKS values for each methods on our ``Lap'' dataset.}
\label{fig:scatter_sco_oks_laprea}
\end{figure}

\begin{figure}[!htb]
\centering
\begin{subfigure}{.33\textwidth}
  \centering \includegraphics[scale=1, width=1\linewidth]{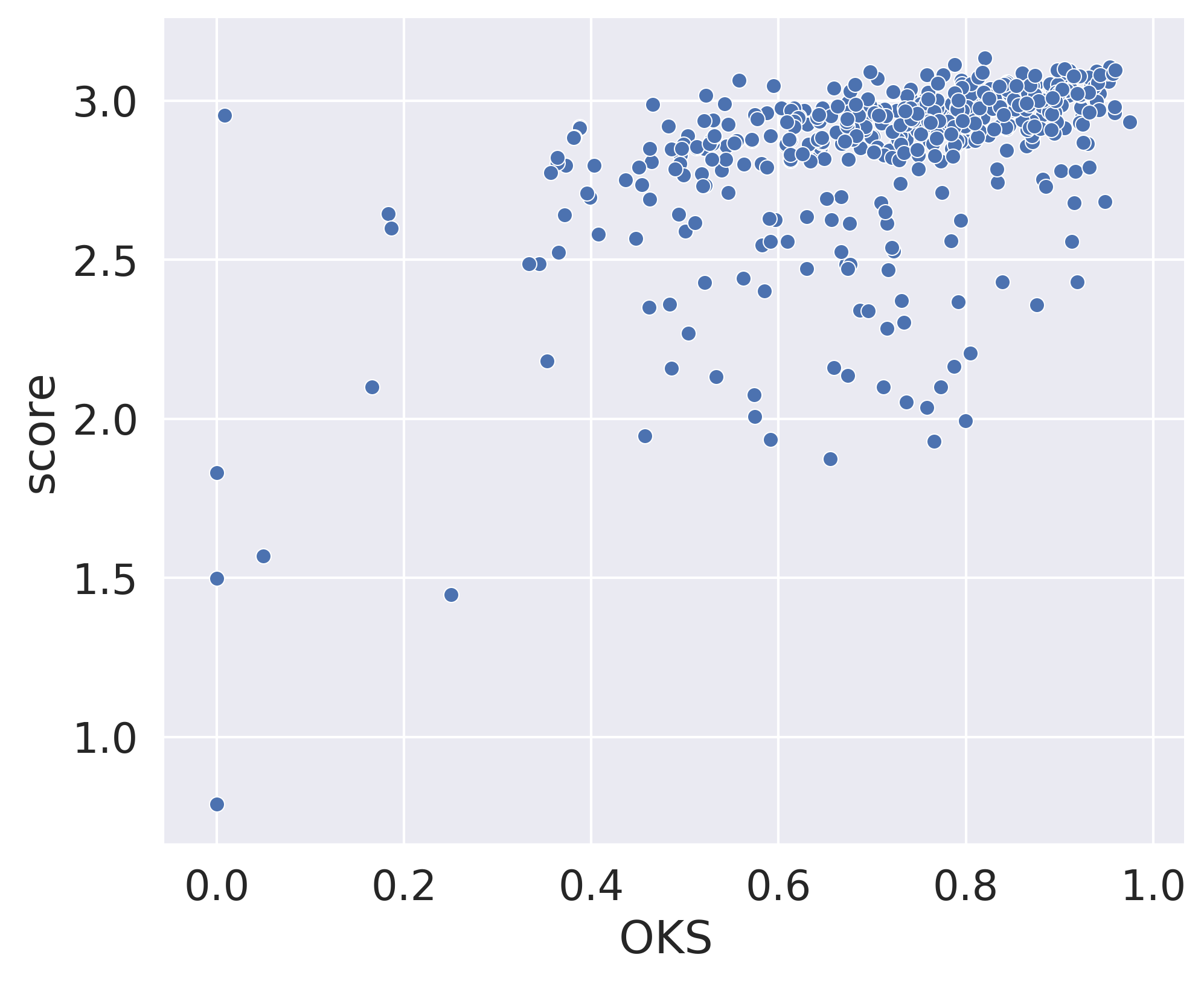}
  \small{(a) AlphaPose}
\end{subfigure}%
\begin{subfigure}{.33\textwidth}
  \centering \includegraphics[scale=1, width=1\linewidth]{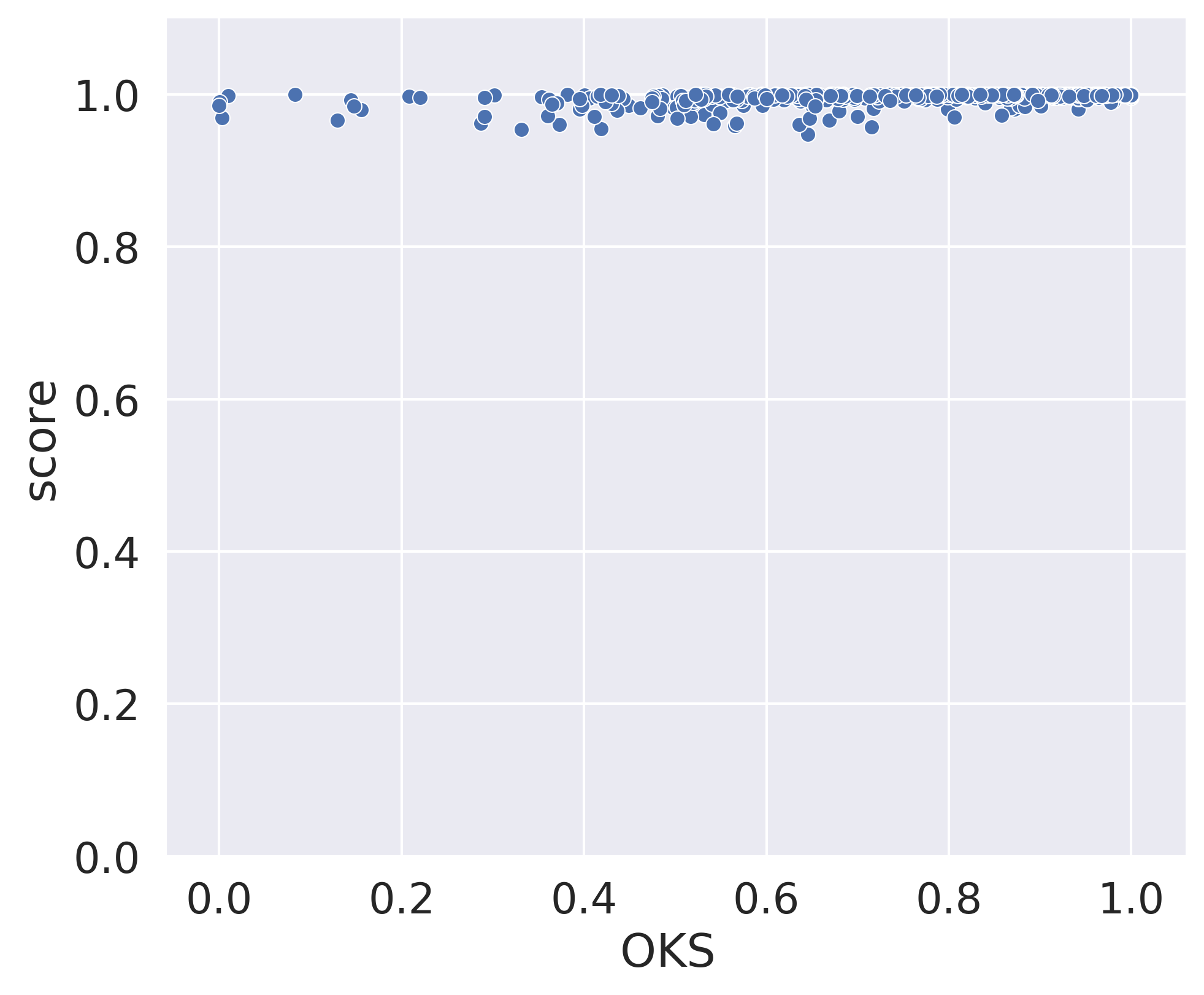}
  \small{(b) Detectron2}
\end{subfigure}%
\begin{subfigure}{.33\textwidth}
  \centering \includegraphics[scale=1, width=1\linewidth]{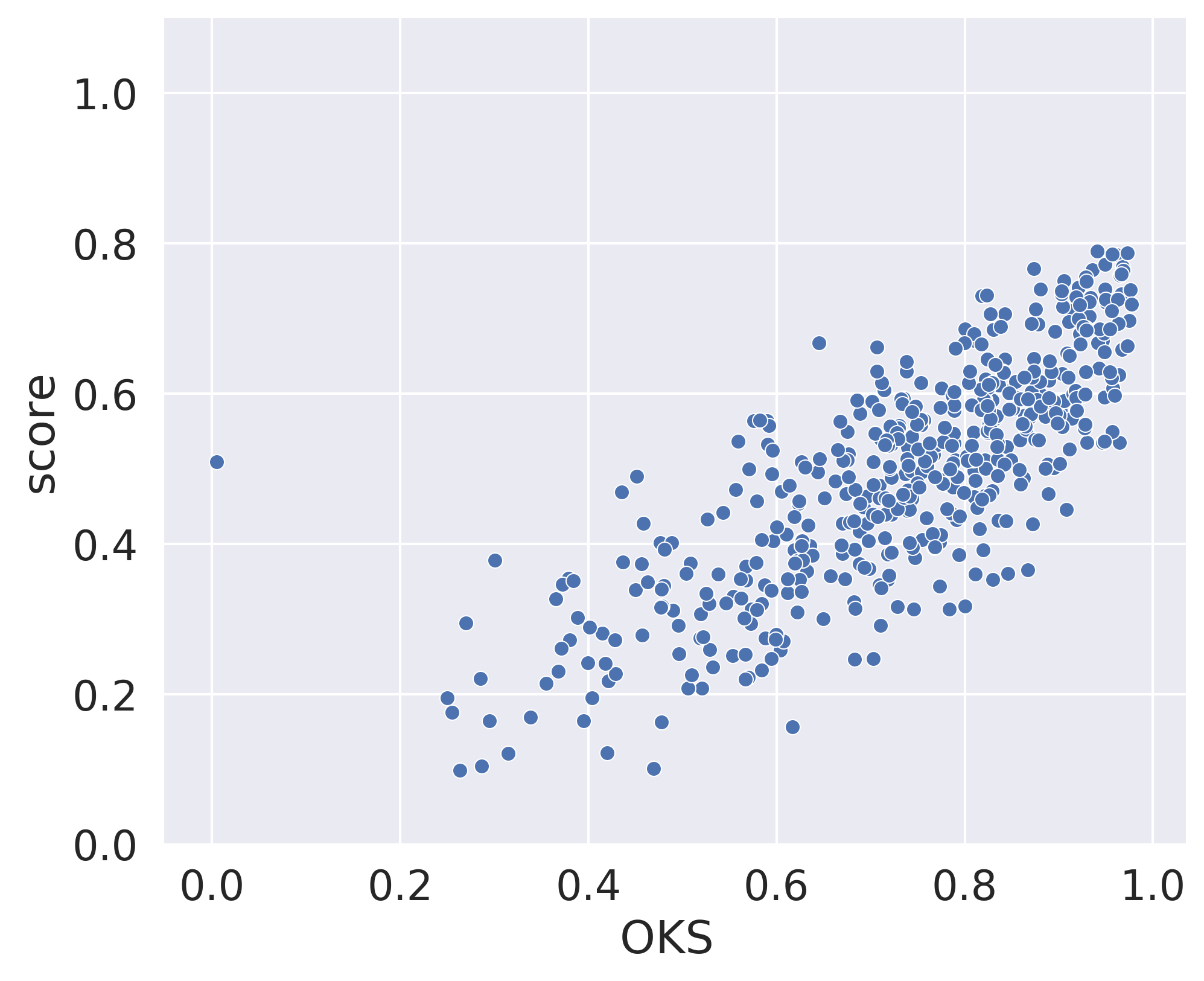}
  \small{(c) HRNet BU}
\end{subfigure}%

\begin{subfigure}{.33\textwidth}
  \centering \includegraphics[scale=1, width=1\linewidth]{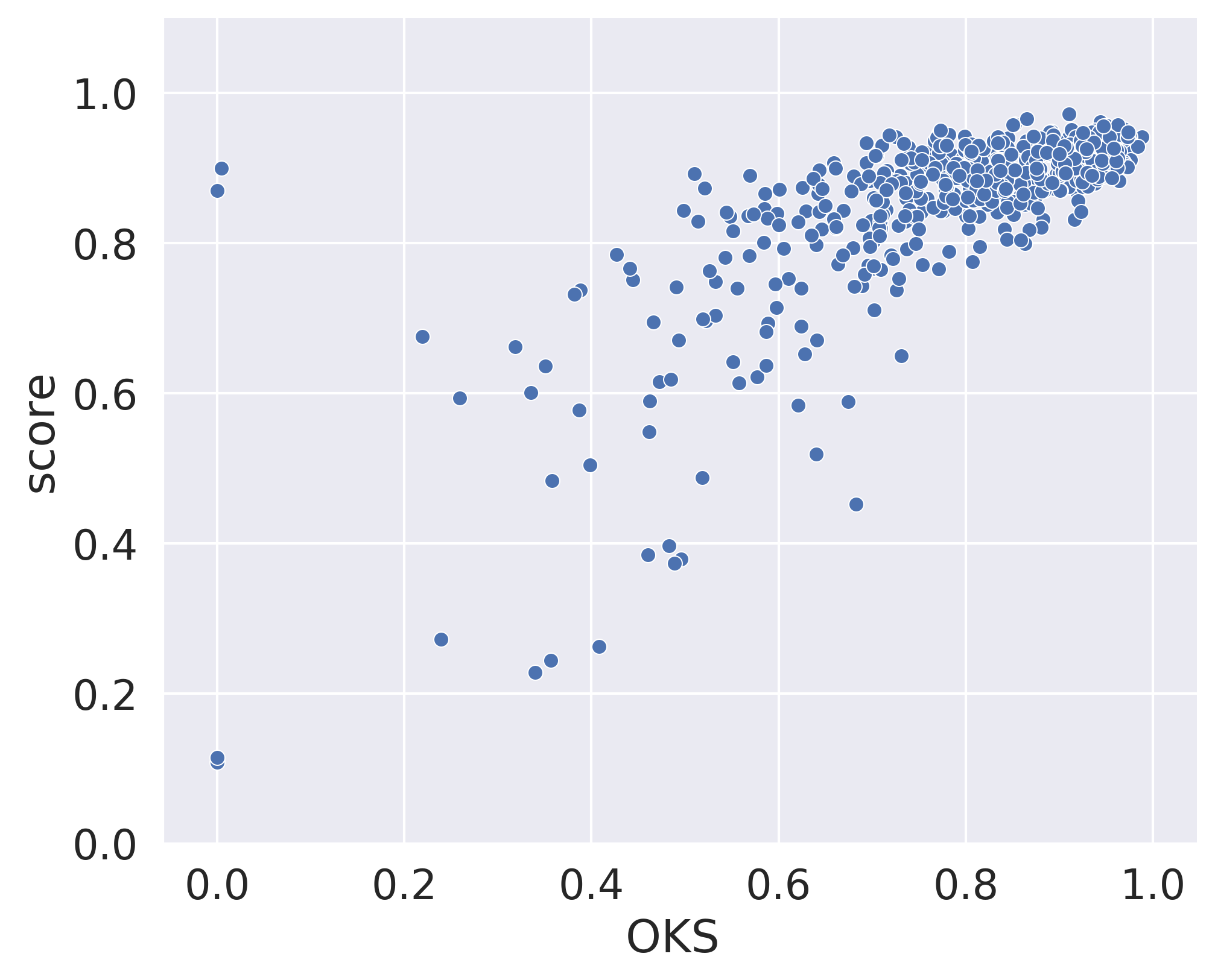}
  \small{(d) HRNet TD}
\end{subfigure}%
\begin{subfigure}{.33\textwidth}
  \centering \includegraphics[scale=1, width=1\linewidth]{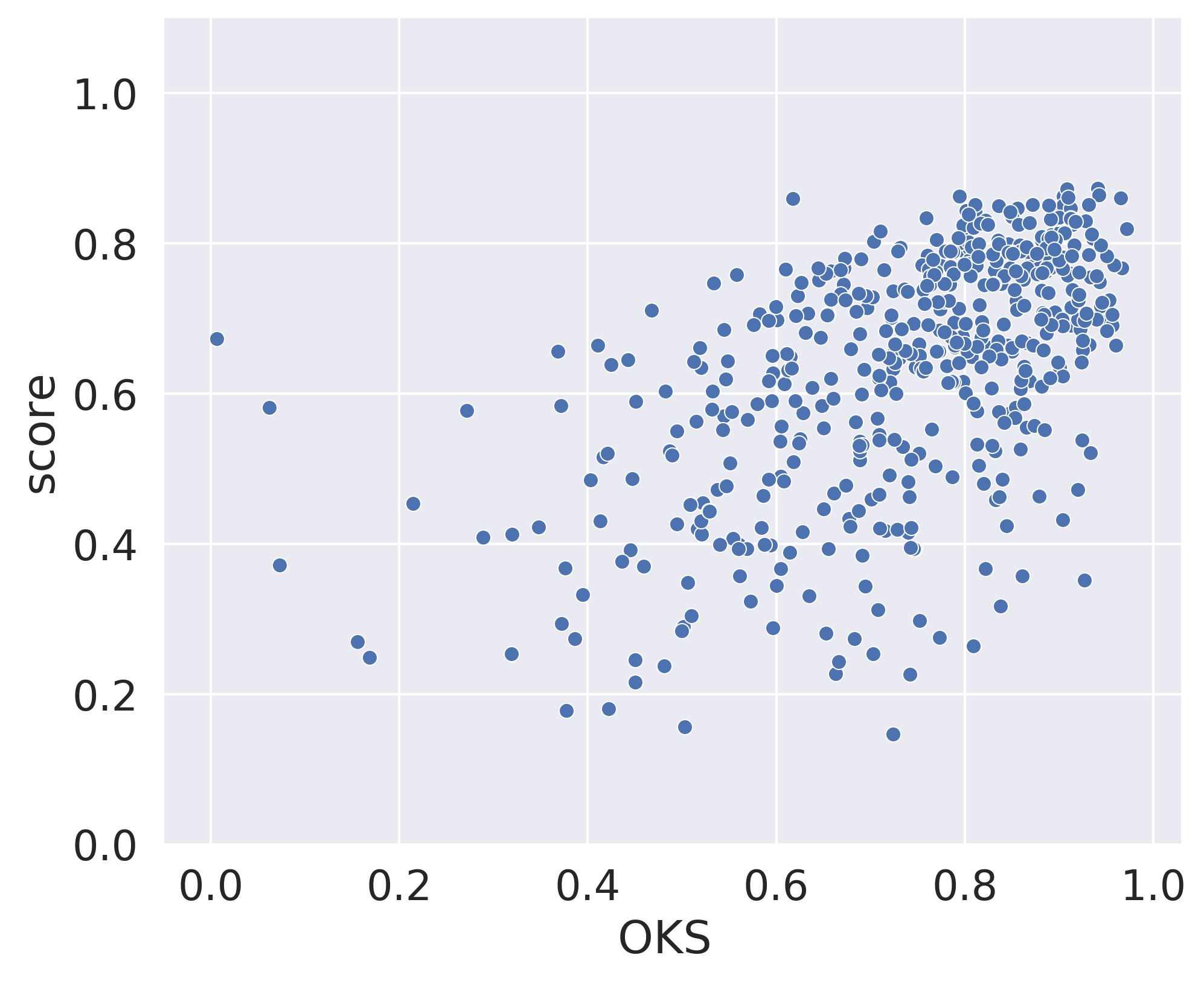}
  \small{(e) OpenPose}
\end{subfigure}%
\begin{subfigure}{.33\textwidth}
  \centering \includegraphics[scale=1, width=1\linewidth]{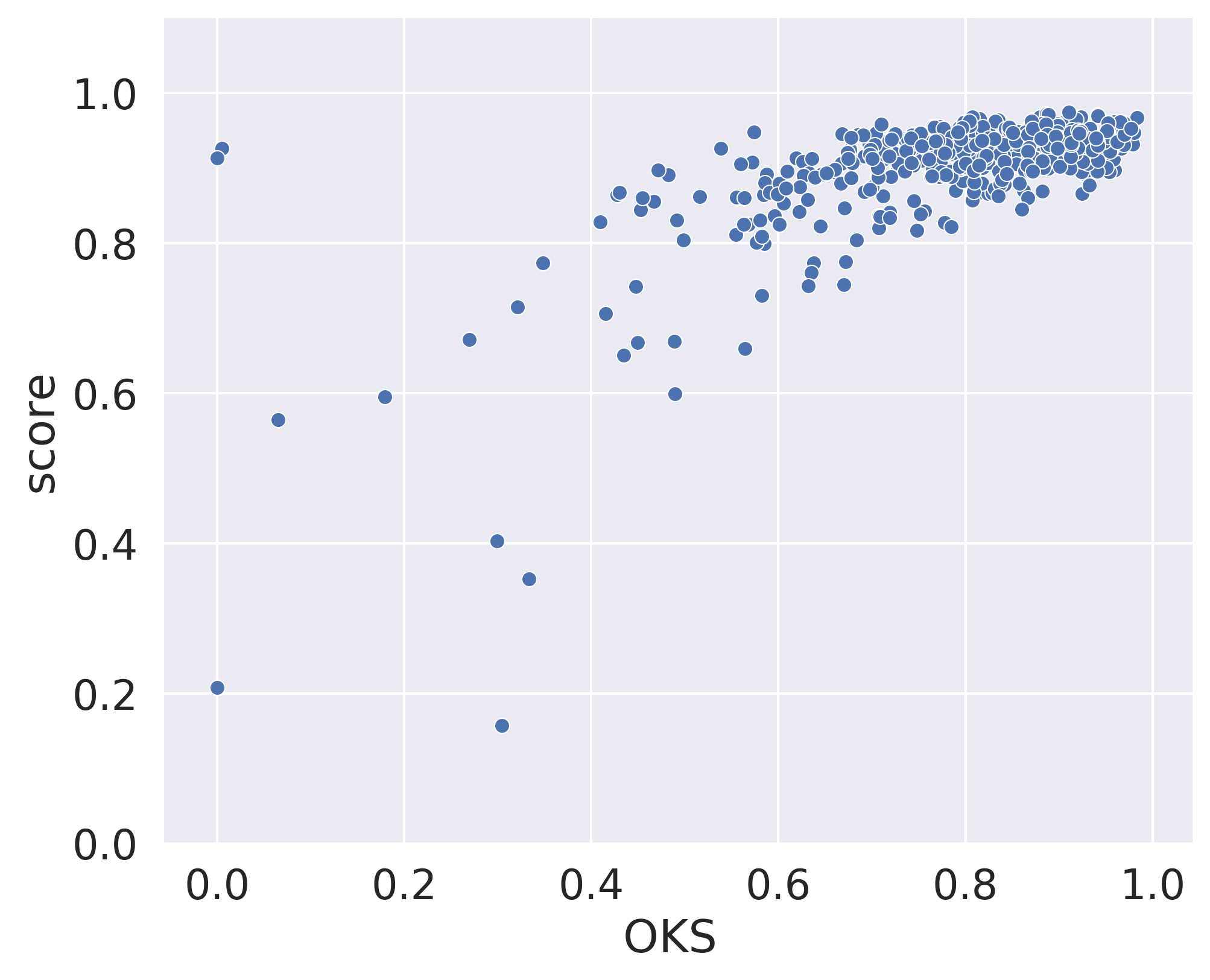}
  \small{(f) ViTPose}
\end{subfigure}%

\caption{Scatterplots of score and OKS values for each methods on the SyRIP dataset.}
\label{fig:scatter_sco_oks_syrip}
\end{figure}

\begin{figure}[!htb]
\centering
\begin{subfigure}{.25\textwidth}
  \centering \includegraphics[scale=1, width=1\linewidth]{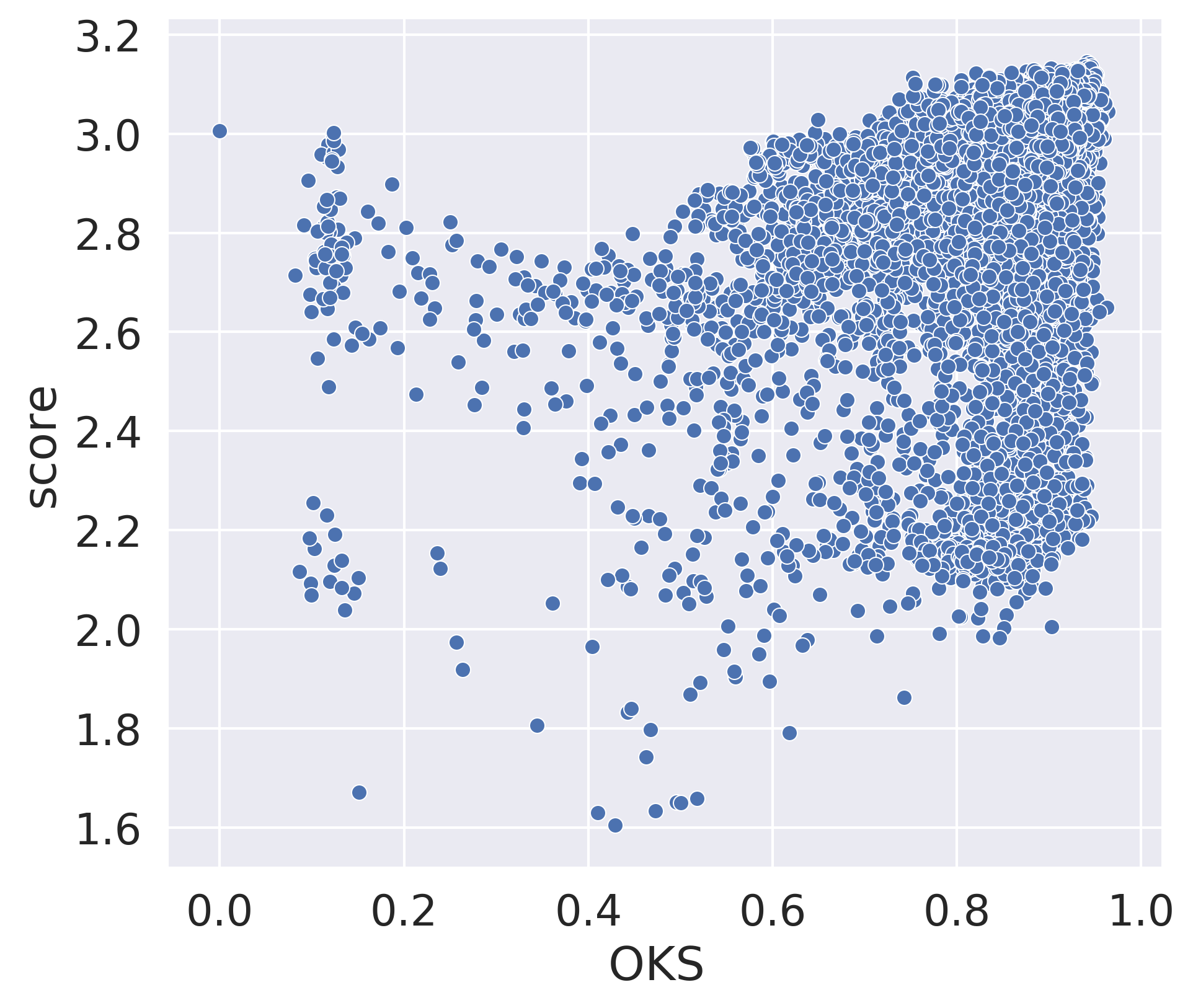}
  \small{(a) AlphaPose -- img}
\end{subfigure}%
\begin{subfigure}{.25\textwidth}
  \centering \includegraphics[scale=1, width=1\linewidth]{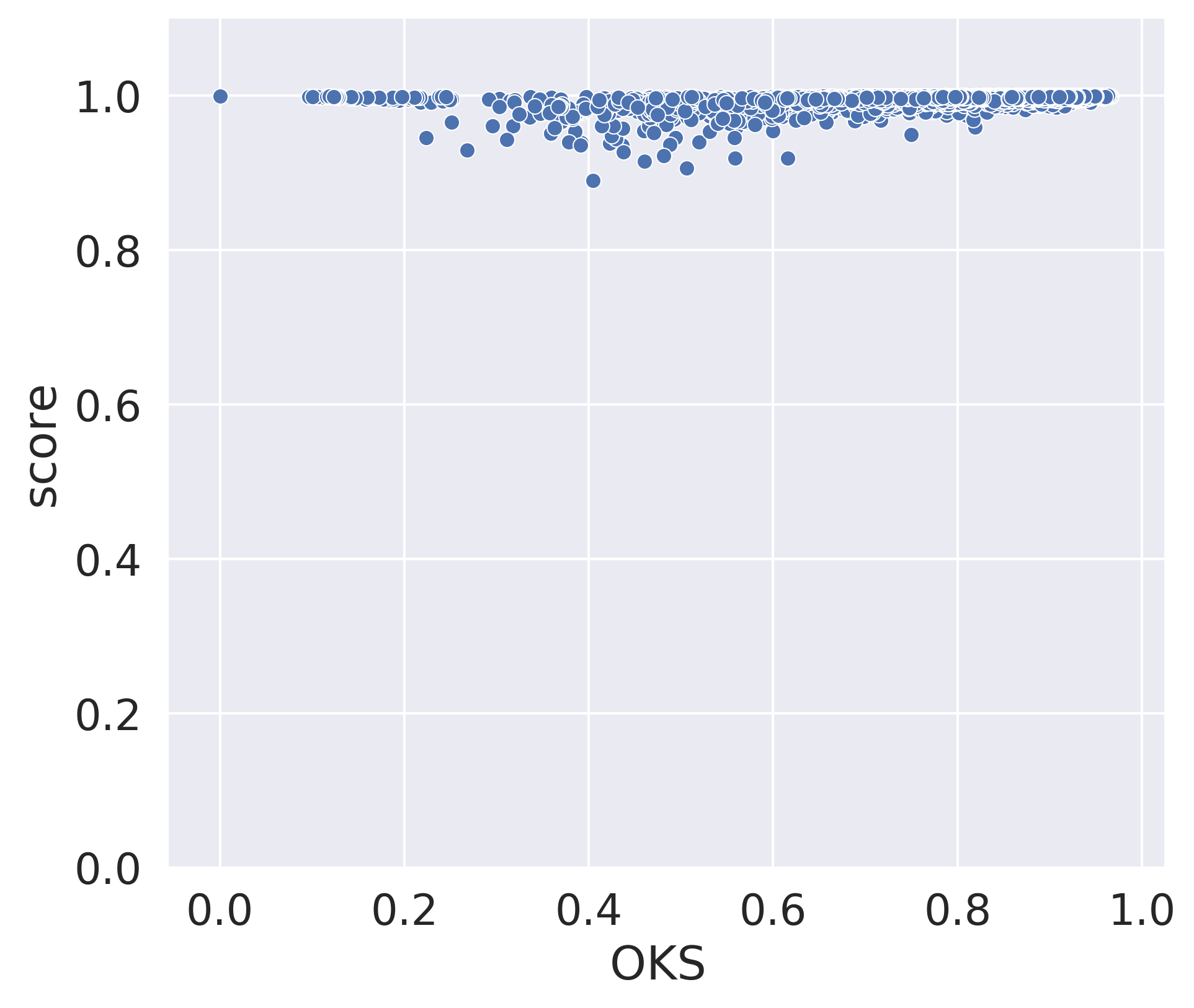}
  \small{(b) Detectron2 -- img}
\end{subfigure}%
\begin{subfigure}{.25\textwidth}
  \centering \includegraphics[scale=1, width=1\linewidth]{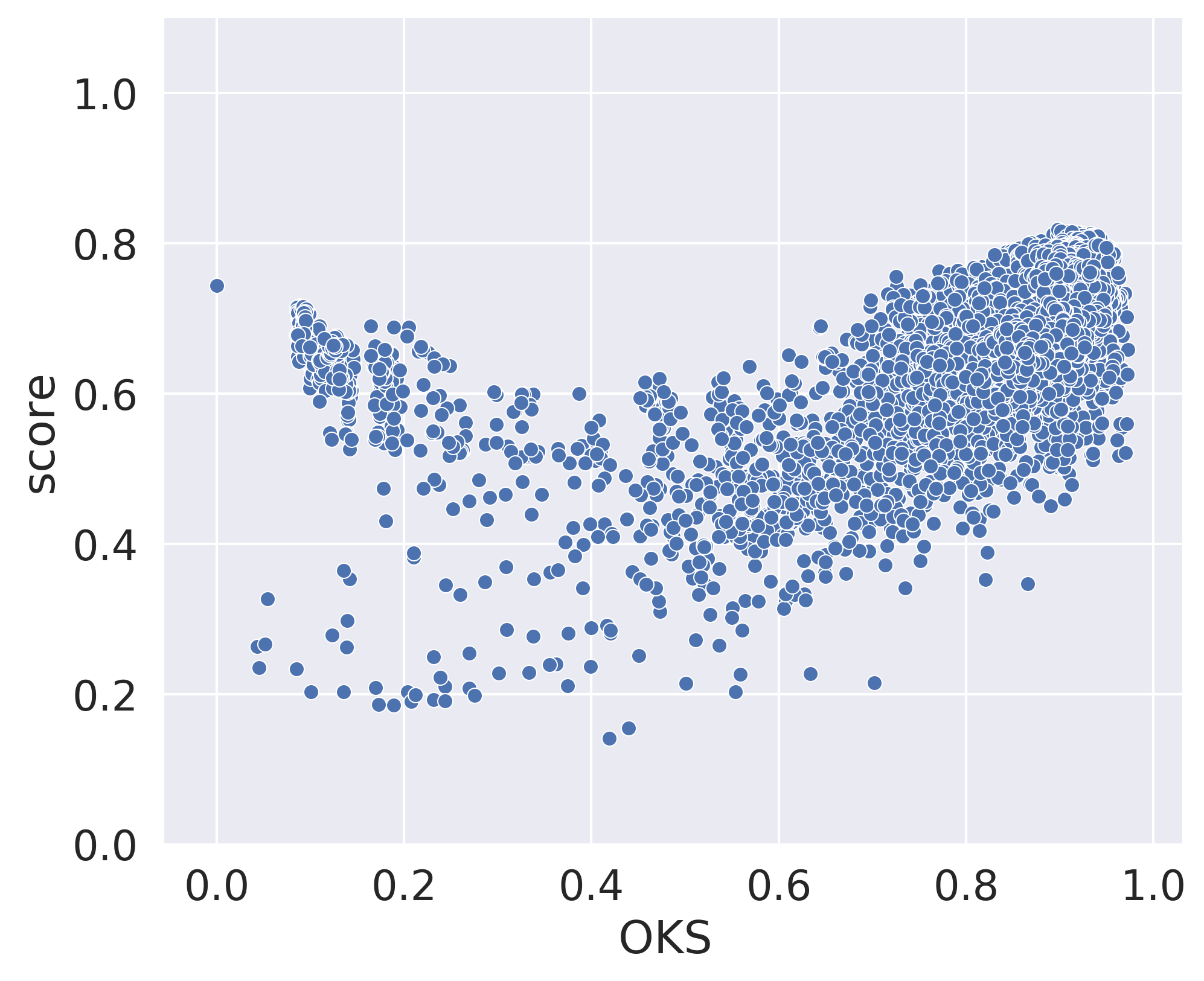}
  \small{(c) HRNet BU -- img}
\end{subfigure}%
\begin{subfigure}{.25\textwidth}
  \centering \includegraphics[scale=1, width=1\linewidth]{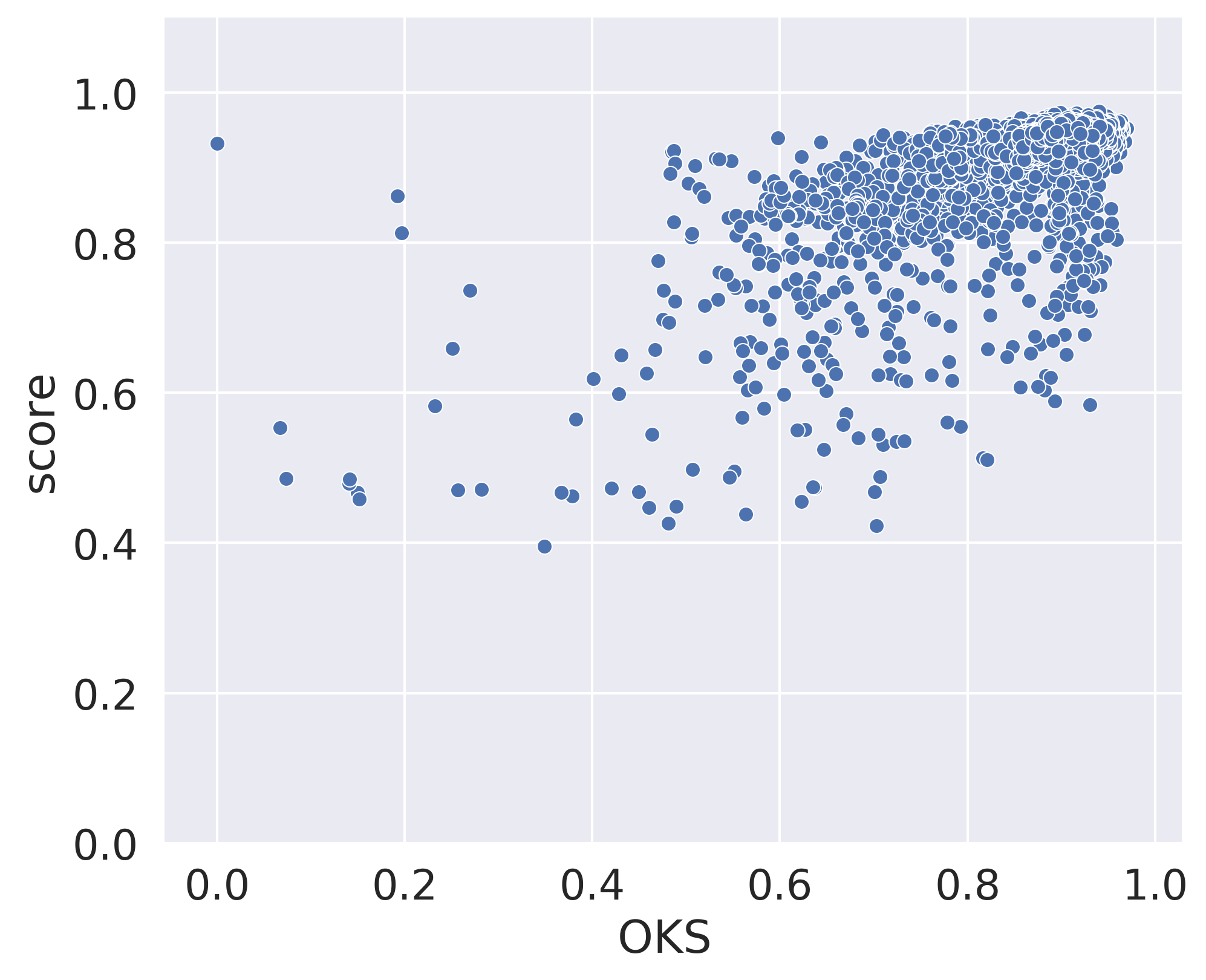}
  \small{(d) HRNet TD -- img}
\end{subfigure}%

\begin{subfigure}{.25\textwidth}
  \centering \includegraphics[scale=1, width=1\linewidth]{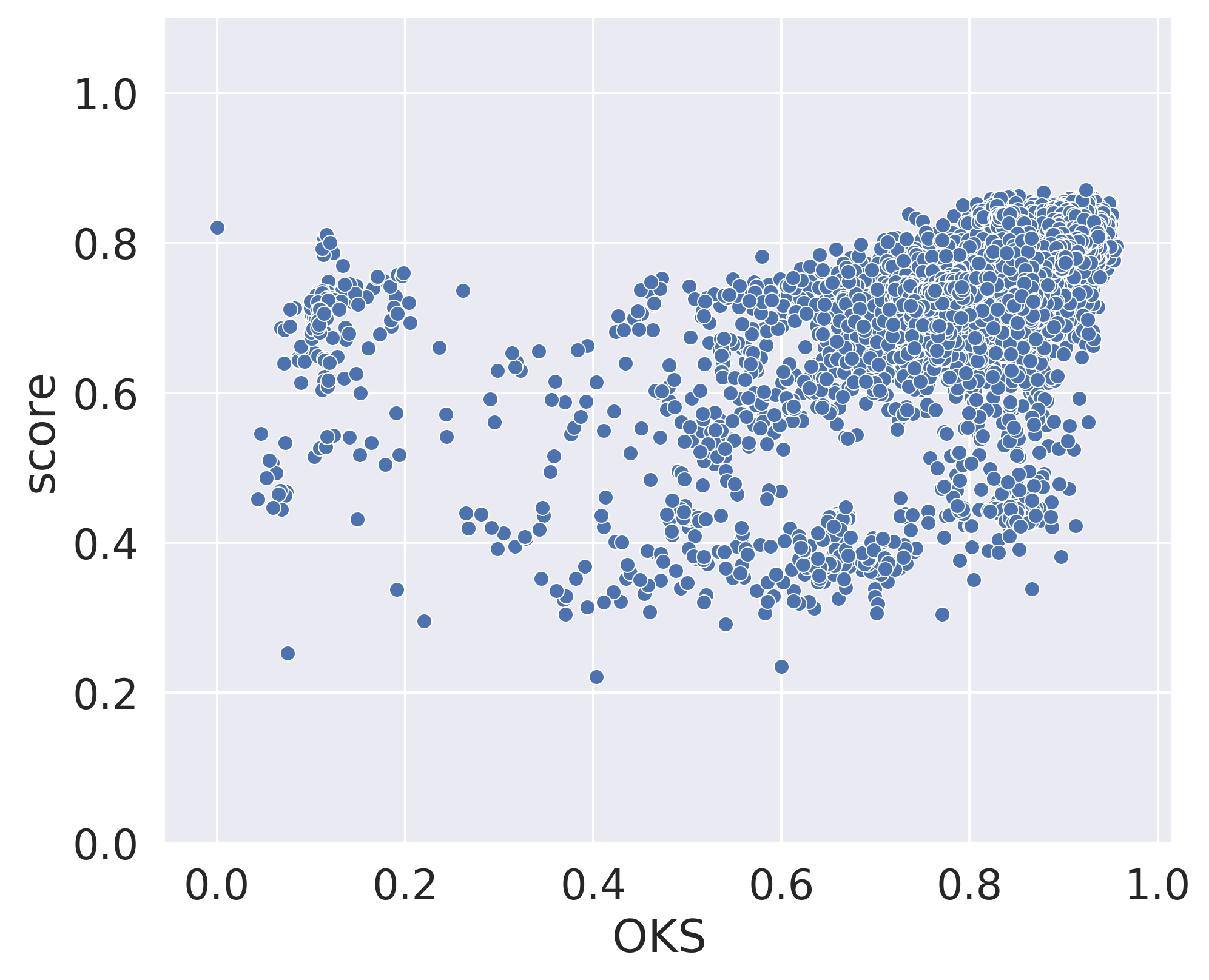}
  \small{(e) OpenPose -- img}
\end{subfigure}%
\begin{subfigure}{.25\textwidth}
  \centering \includegraphics[scale=1, width=1\linewidth]{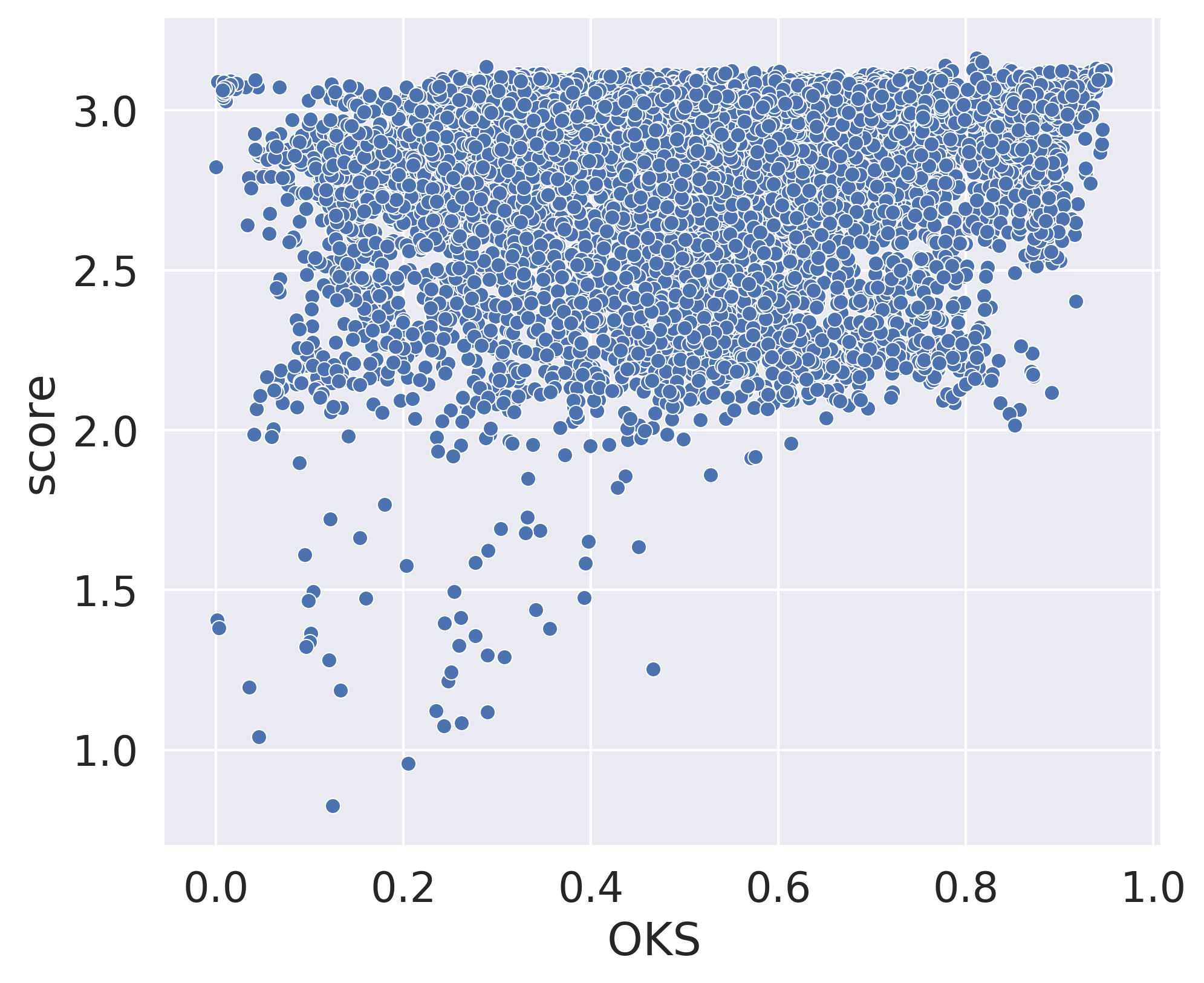}
  \small{(f) AlphaPose -- vid}
\end{subfigure}%
\begin{subfigure}{.25\textwidth}
  \centering \includegraphics[scale=1, width=1\linewidth]{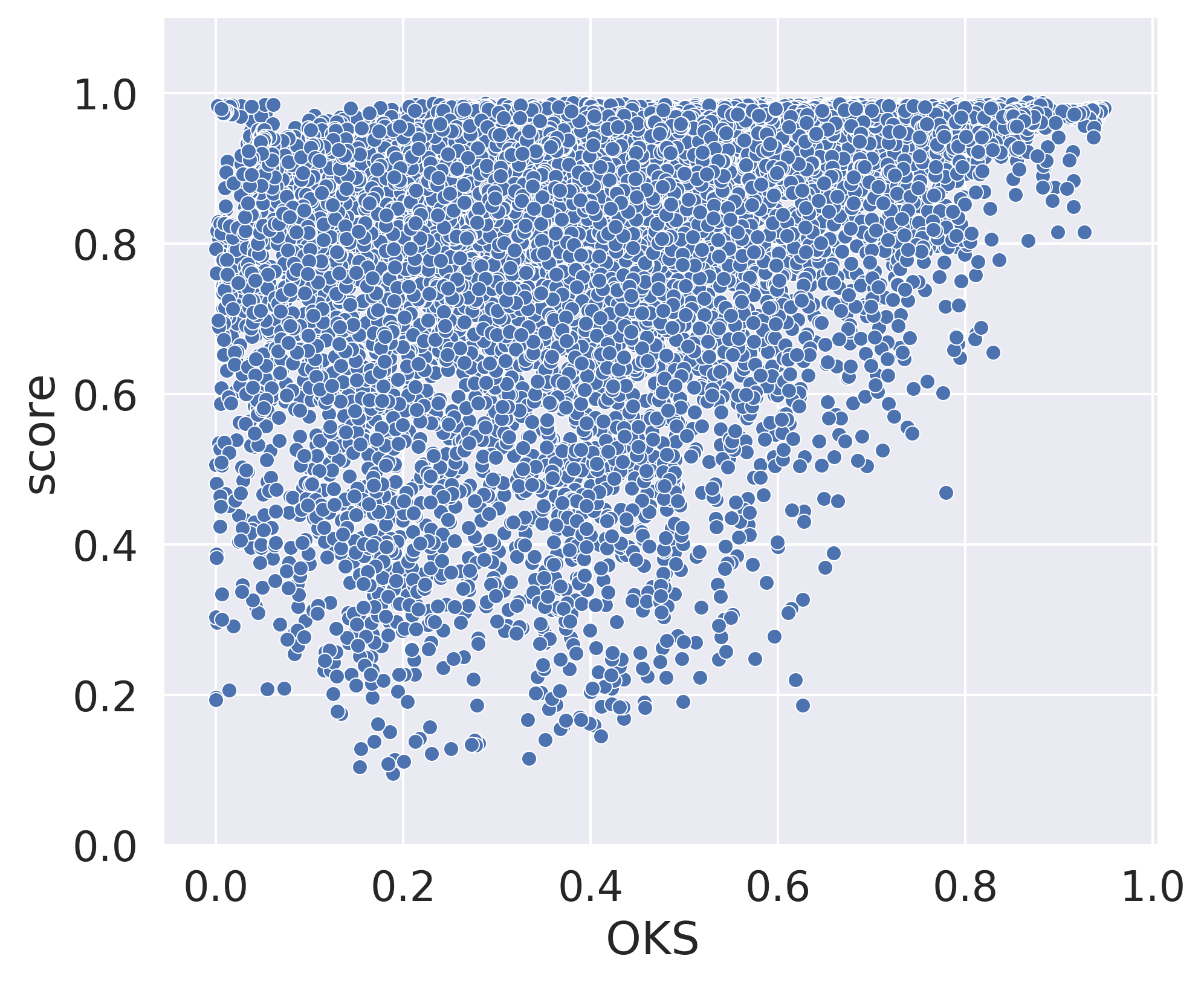}
  \small{(g) DeepLabCut -- vid}
\end{subfigure}%
\begin{subfigure}{.25\textwidth}
  \centering \includegraphics[scale=1, width=1\linewidth]{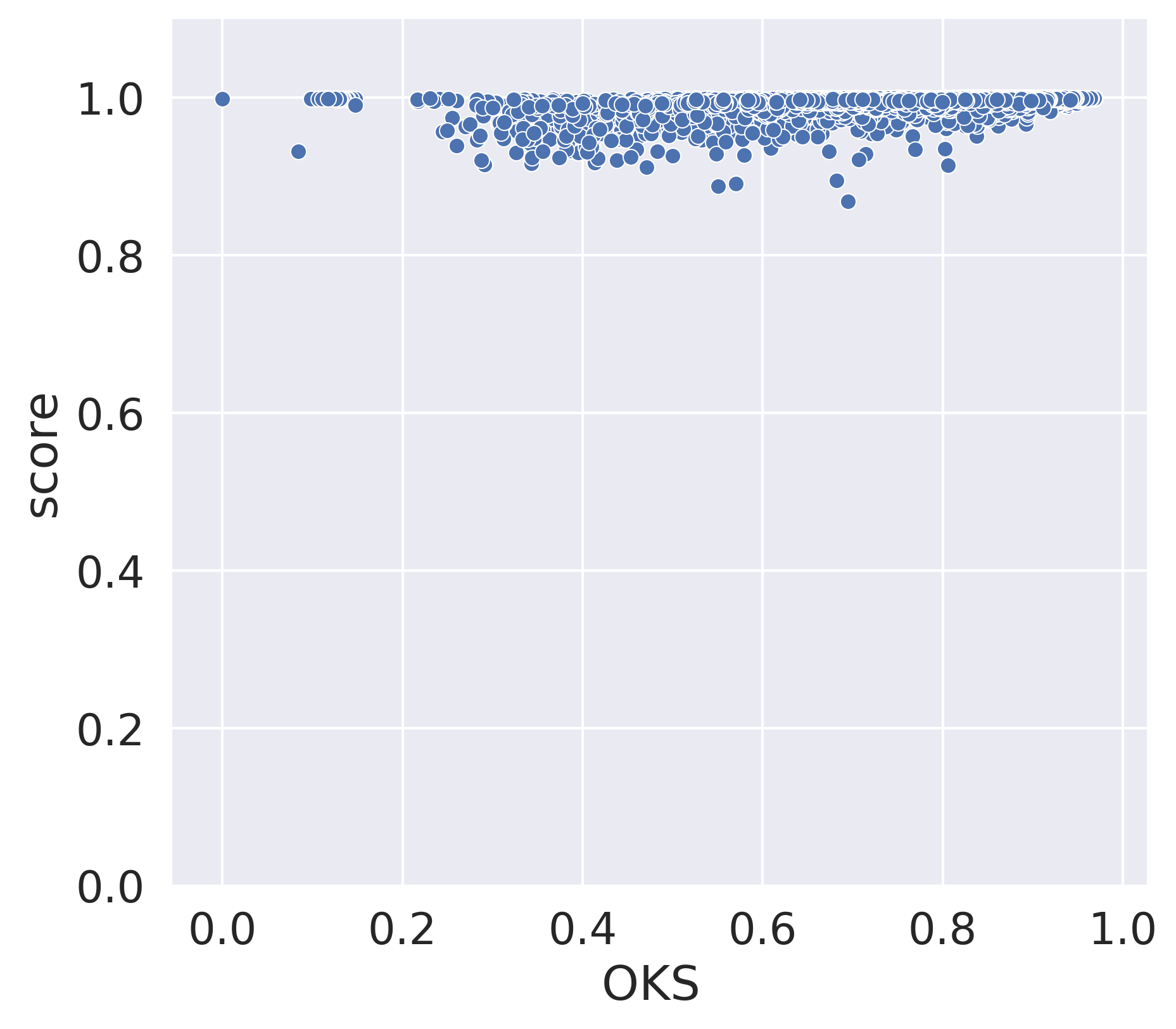}
  \small{(h) Detectron2 -- vid}
\end{subfigure}%

\begin{subfigure}{.25\textwidth}
  \centering \includegraphics[scale=1, width=1\linewidth]{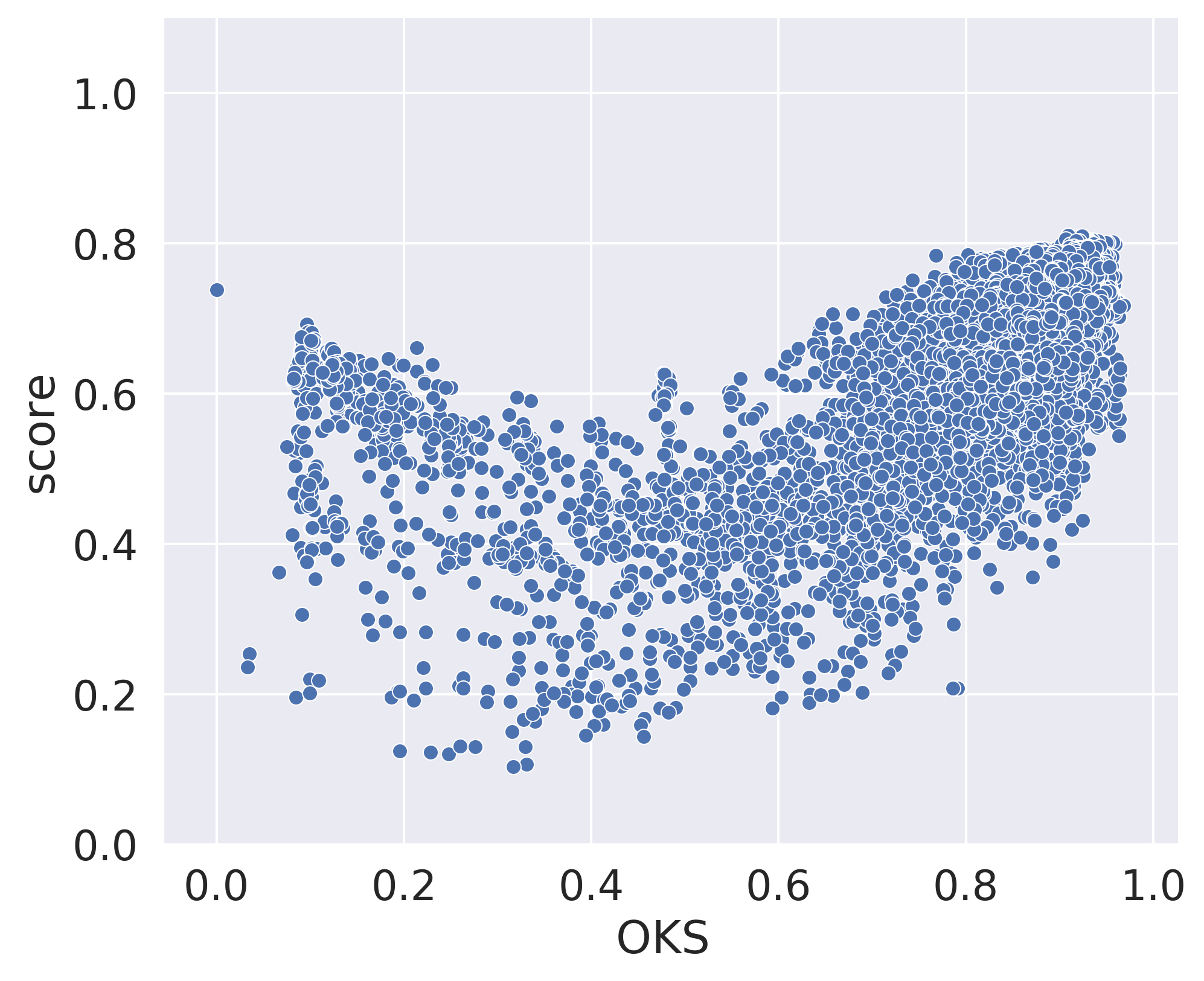}
  \small{(i) HRNet BU -- vid}
\end{subfigure}%
\begin{subfigure}{.25\textwidth}
  \centering \includegraphics[scale=1, width=1\linewidth]{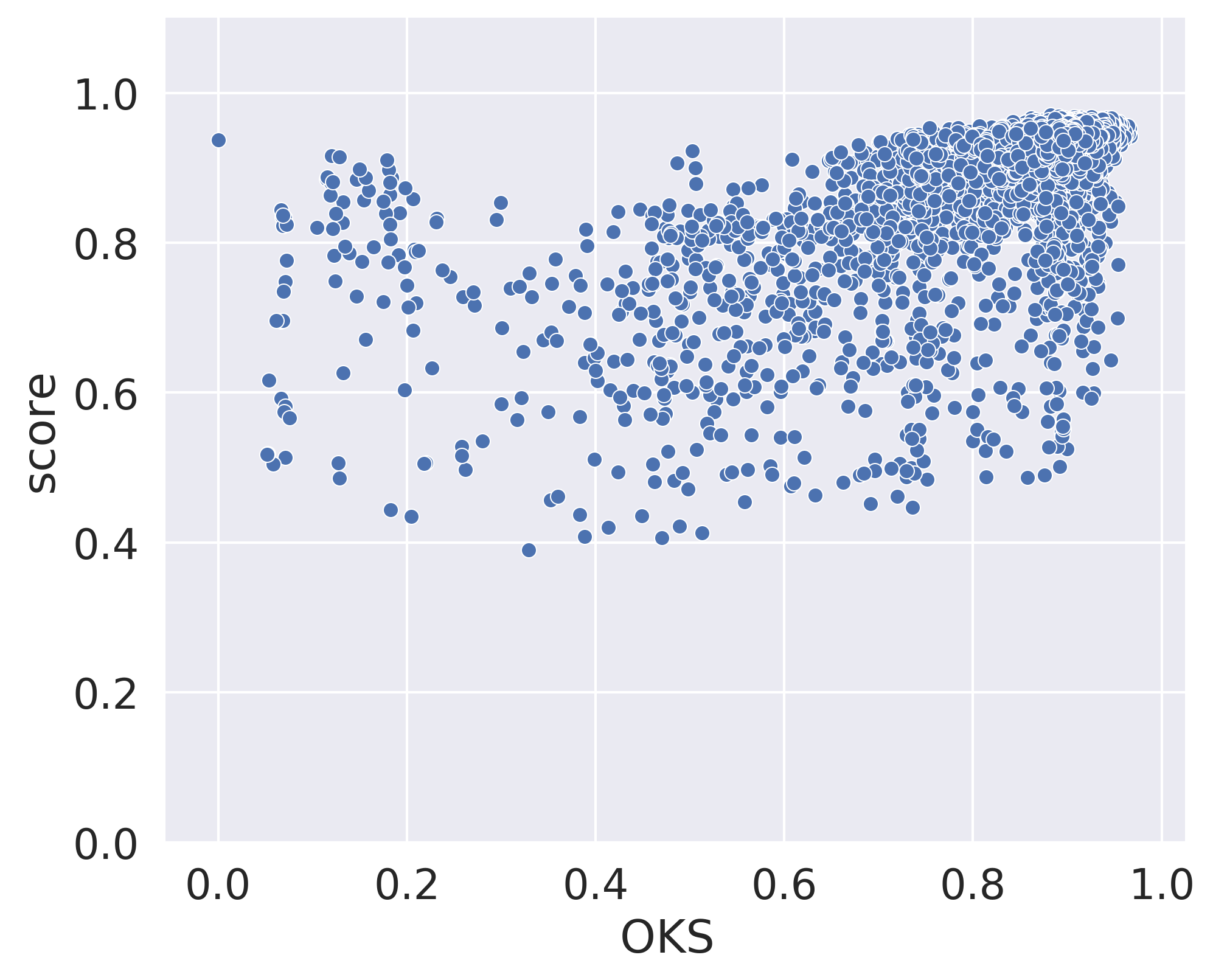}
  \small{(j) HRNet TD -- vid}
\end{subfigure}%
\begin{subfigure}{.25\textwidth}
  \centering \includegraphics[scale=1, width=1\linewidth]{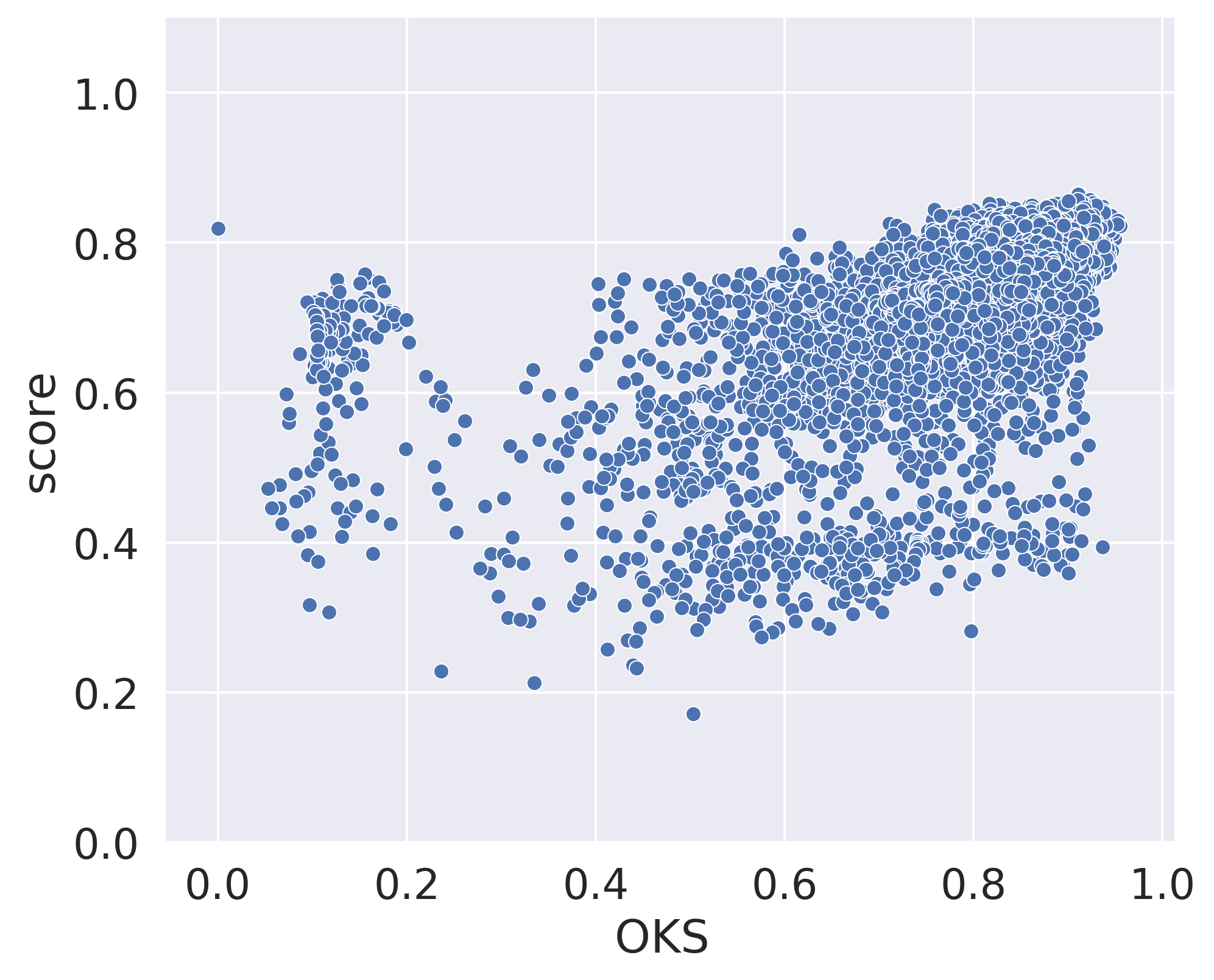}
  \small{(k) OpenPose -- vid}
\end{subfigure}%
\begin{subfigure}{.25\textwidth}
  \centering \includegraphics[scale=1, width=1\linewidth]{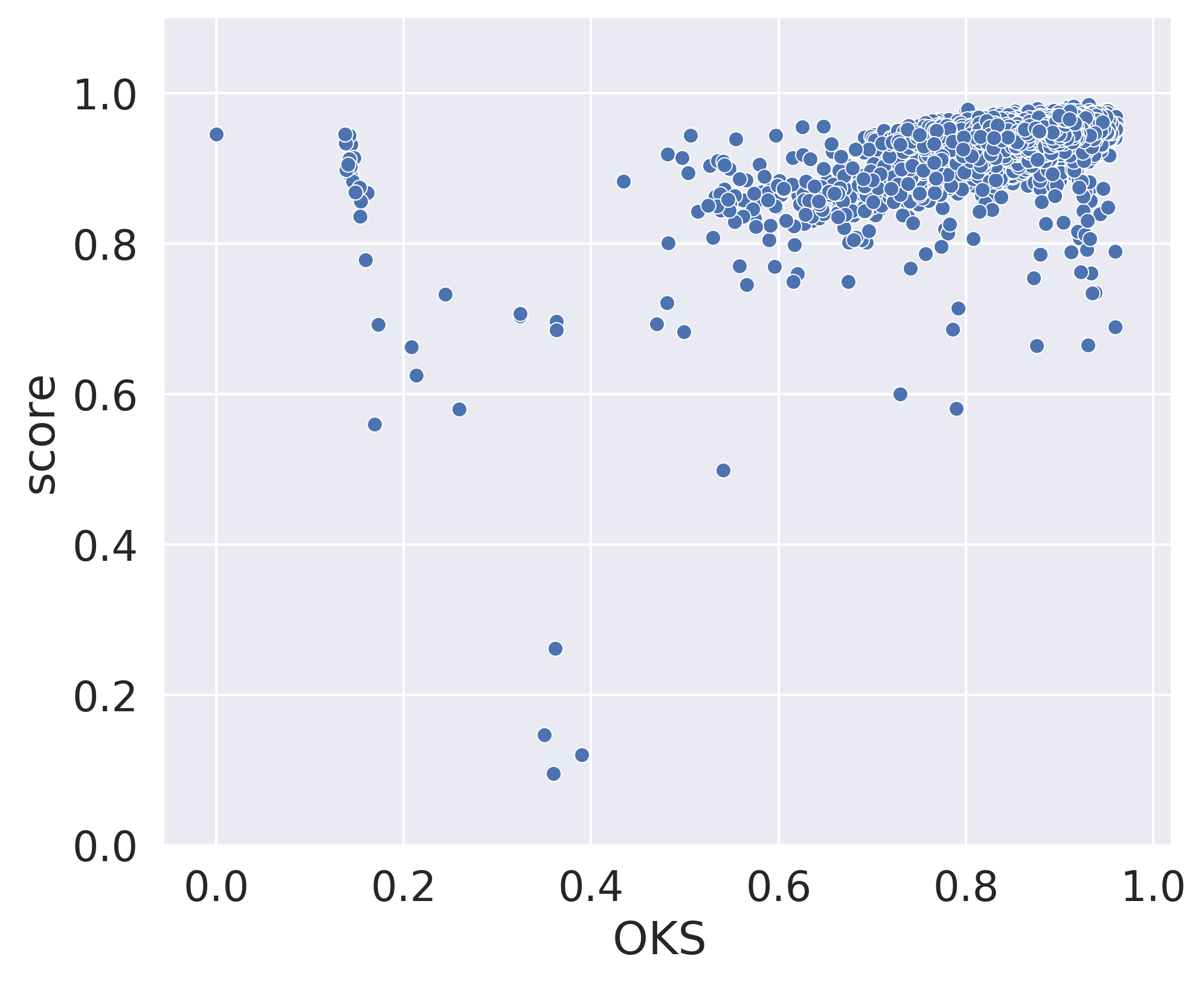}
  \small{(l) ViTPose -- vid}
\end{subfigure}%

\caption{Scatterplots of score and OKS values for each methods on the MINi-RGBD dataset, (a-e) with image input; (f-l) with video input.}
\label{fig:scatter_sco_oks_synth}
\end{figure}

The scatterplots of scores and OKS values are shown in Fig. SF.~\ref{fig:scatter_sco_oks_suprea}, SF.~\ref{fig:scatter_sco_oks_laprea} for real infants in supine and lap positions, SF.~\ref{fig:scatter_sco_oks_syrip} for the SyRIP dataset, and lastly,  SF.~\ref{fig:scatter_sco_oks_synth} for the MINI-RGBD synthetic infants.

We observe that no matter the dataset or input modes, Detectron2 always provides high scores making this value have little informative value for this method. DeepLabCut does not have many high-OKS estimates, and its low-OKS estimates can have scores that go over the full range of possible values, including high score values. AlphaPose and OpenPose show scattered results, with a number of low-OKS estimates getting high scores, and high-OKS estimates getting lower scores.
HRNet BU, TD, and ViTPose seem to have a more meaningful relation between OKS and score value, though it might be biased due to having mainly good estimates (high-OKS), and very few low-OKS estimates. Indeed, in the most difficult dataset, the ``Lap'', the results are more scattered, though it is possibly due to the methods estimating the parent instead of the infant: if there is only one detection by the method, we compared it to the ground truth, which in the case of a parent detection, would turn into low-OKS values for these estimates, but the method might be correct about giving high scores and might have gotten a high OKS score if the detection had been compared to a ground truth of the parent instead of the parent's. The methods do keep seemingly more meaningful relationship in the SyRIP dataset, which is harder than the ``Supine'' dataset while not at risk of frequently detecting an adult instead of an infant.

\clearpage

\section*{Supplementary results summary}
\label{sec:results_sum}

In a best-case scenario (see Tab. ST~\ref{tab:summary_seq}), pose estimation methods can reach high levels of performance on infants in supine position, with low variability, with ViTPose reaching an Average Precision up to 91.8, and average errors around 3.9$\pm$2.0\% of the Neck-MidHip segment (which corresponds roughly to the infant's torso length).

With the detailed OKS values for each video (Tables ST.~\ref{tab:real_oks} and~\ref{tab:synth_oks}), we could get further insight with regards to the complexity estimation made by Hesse et al. on their MINI-RGBD dataset based on the actual performance of the methods' estimates. Such a table can also help to identify for each specific method which kind of sequences they seem to struggle with to identify the possible causes of keypoint misplacement, so that more of such examples can be included in future methods' training or fine-tuning. For example, MediaPipe did not manage to identify a single image among the sequence for synthetic infant 2, despite it being deemed "easy", possibly due to its unique background.

From Table 7 of the main manuscripts and Tables of Supplementary Materials ST~\ref{tab:all_real} and ST~\ref{tab:summary_seq}, it seems that the better the estimates and the easier the videos, the lower the correlation between scores and OKS values. This could be explained by a lower range of high values from OKS, while the scores might be less reliable and might not reflect a similar reduction in their variability and range of values.

\end{document}